\documentclass{article}

\usepackage{multirow}
\usepackage{adjustbox}

\usepackage[final,nonatbib]{neurips_2022}

\usepackage[utf8]{inputenc} %
\usepackage[T1]{fontenc}    %
\usepackage[colorlinks,linkcolor=red,citecolor=blue,urlcolor=magenta,bookmarks=true]{hyperref}       %
\usepackage{url}            %
\usepackage{booktabs}       %
\usepackage{amsfonts}       %
\usepackage{nicefrac}       %
\usepackage{microtype}      %
\usepackage{xcolor}         %
\usepackage{subcaption}
\usepackage{comment}
\usepackage{floatrow}
\newfloatcommand{capbtabbox}{table}[][\FBwidth]
\usepackage{array}
\usepackage{multirow}
\usepackage{wrapfig}
\usepackage{bm}
\usepackage{amsmath}
\usepackage{color}
\usepackage{diagbox}
\usepackage{amssymb}%
\usepackage{pifont}%
\newcommand{\cmark}{\ding{51}}%
\newcommand{\xmark}{\ding{55}}%

\DeclareMathOperator*{\argmin}{argmin}
\usepackage{adjustbox}

\definecolor{asparagus}{rgb}{0.53, 0.66, 0.42}

\newcommand{\ourmodel}{\textsc{GET3D}}

\definecolor{dark_green}{rgb}{0.0, 0.7, 0.0}

\title{GET3D: A Generative Model of High Quality 3D Textured Shapes Learned from Images}

\author{%
Jun Gao$^{1,2,3}$  \quad\qquad Tianchang Shen$^{1,2,3}$ \quad\qquad Zian Wang$^{1,2,3}$  \quad\qquad Wenzheng Chen$^{1,2,3}$  \vspace{8pt}\\
 \textbf{Kangxue Yin$^{1}$} \quad\qquad \textbf{Daiqing Li$^{1}$} \quad\qquad \textbf{Or Litany$^{1}$}  \quad\qquad \textbf{Zan Gojcic$^{1}$} \quad\qquad \textbf{Sanja Fidler$^{1,2,3}$} \vspace{8pt}\\
\small{NVIDIA\textsuperscript{1} \quad University of Toronto\textsuperscript{2} \quad Vector Institute\textsuperscript{3} } \vspace{3pt}\\
\texttt{\scriptsize \{jung, frshen, zianw, wenzchen, kangxuey, daiqingl, olitany, zgojcic, sfidler\}@nvidia.com}
}

\begin{document}

\maketitle
\begin{abstract}
As several industries are moving towards modeling massive 3D virtual worlds, the need for content creation tools that can scale in terms of the quantity, quality, and diversity of 3D content is becoming evident. 
In our work, we aim to train performant 3D generative models that synthesize textured meshes that can be directly
consumed by 3D rendering engines, thus immediately usable in downstream applications. Prior works on 3D generative modeling either lack geometric details, are limited in the mesh topology they can produce, typically do not support textures, or utilize neural renderers in the synthesis process, which makes their use in common 3D software non-trivial. In this work, we introduce {\ourmodel}, a \textbf{G}enerative model that directly generates \textbf{E}xplicit \textbf{T}extured \textbf{3D} meshes with complex topology, rich geometric details, and high fidelity textures. 
We bridge recent success in the differentiable surface modeling, differentiable rendering as well as 2D Generative Adversarial Networks to train our model from 2D image collections.
{\ourmodel} is able to generate high-quality 3D textured meshes, ranging from cars, chairs, animals, motorbikes and human characters to buildings, achieving significant improvements over previous methods. Our project page: \url{https://nv-tlabs.github.io/GET3D}
\end{abstract}

\section{Introduction}
Diverse, high-quality 3D content is becoming increasingly important for several industries, including gaming, robotics, architecture, and social platforms. However, manual creation of 3D assets is very time-consuming and requires specific technical knowledge as well as artistic modeling skills. One of the main challenges is thus scale -- while one can find 3D models on 3D marketplaces such as Turbosquid~\cite{Turbosquid} or Sketchfab~\cite{sketchfab}, creating many 3D models to, say, populate a game or a movie with a crowd of characters that all look different still takes a significant amount of artist time. 

To facilitate the content creation process and make it accessible to a variety of (novice) users, generative 3D networks that can produce high-quality and diverse 3D assets have recently become an active area of research~\cite{achlioptas2018learning,imnet, occnet, mo2019structurenet, pavllo2021textured3dgan, pointflow, zhou2021pvd,dmtet, shen2020interactive,yin20213dstylenet,gao2019deepspline}. However, to be practically useful for current real-world applications, 3D generative models should ideally fulfill the following requirements: \textbf{(a)} They should have the capacity to generate shapes with detailed geometry and arbitrary topology, \textbf{(b)} The output should be a textured mesh, which is a primary representation used by standard graphics software packages such as Blender~\cite{blender} and Maya~\cite{maya}, and \textbf{(c)} We should be able to leverage 2D images for supervision, as they are more widely available than explicit 3D shapes.

Prior work on 3D generative modeling has focused on subsets of the above requirements, but no method to date fulfills all of them (Tab.~\ref{tbl:concept_compare}). For example, methods that generate 3D point clouds~\cite{achlioptas2018learning, pointflow, zhou2021pvd} typically do not produce textures and have to be converted to a mesh in post-processing. Methods generating voxels often lack geometric details and do not produce texture~\cite{wu2016learning,gadelha20173d,henzler2019platonicgan,lunz2020inverse}. Generative models based on neural fields~\cite{occnet,imnet} focus on extracting geometry but disregard texture. Most of these also require explicit 3D supervision. Finally, methods that directly output textured 3D meshes~\cite{pavllo2020convmesh, pavllo2021textured3dgan} typically require pre-defined shape templates and cannot generate shapes with complex topology and variable genus. 

Recently, rapid progress in neural volume rendering~\cite{nerf} and 2D Generative Adversarial Networks (GANs)~\cite{stylegan,stylegan2,stylegan3,huang2021poegan,park2019SPADE} has led to the rise of 3D-aware image synthesis~\cite{pigan,graf,eg3d,giraffe,orel2021styleSDF,gu2021stylenerf}. However, this line of work aims to synthesize multi-view consistent images using neural rendering in the synthesis process and does not guarantee that meaningful 3D shapes can be generated.
While a mesh can potentially be obtained from the underlying neural field representation using the marching cube algorithm~\cite{marchingcube}, extracting the corresponding texture is non-trivial. 

In this work, we introduce a novel approach that aims to tackle all the requirements of a practically useful 3D generative model. Specifically, we propose {\ourmodel}, a \textbf{G}enerative model for 3D shapes that directly outputs \textbf{E}xplicit \textbf{T}extured \textbf{3D} meshes with high geometric and texture detail and arbitrary mesh topology. In the heart of our approach is a generative process that utilizes a differentiable \emph{explicit} surface extraction method~\cite{dmtet} and a differentiable rendering technique~\cite{nvdiffrec,nvdiffrast}. The former enables us to directly optimize and output textured 3D meshes with arbitrary topology, while the latter allows us to train our model with 2D images, thus leveraging powerful and mature discriminators developed for 2D image synthesis. Since our model directly generates meshes and uses a highly efficient (differentiable) graphics renderer, we can easily scale up our model to train with image resolution as high as $1024\times 1024$, allowing us to learn high-quality geometric and texture details. 

We demonstrate state-of-the-art performance for unconditional 3D shape generation on multiple categories with complex geometry from ShapeNet~\cite{shapenet}, Turbosquid~\cite{Turbosquid} and Renderpeople~\cite{renderpeople}, such as chairs, motorbikes, cars, human characters, and buildings. With explicit mesh as output representation, {\ourmodel} is also very flexible and can easily be adapted to other tasks, including: \textbf{(a)} learning to generate decomposed material and view-dependent lighting effects using advanced differentiable rendering~\cite{chen2021dibrpp}, without supervision, \textbf{(b)} text-guided 3D shape generation using CLIP~\cite{CLIP} embedding.

\begin{table}[t!]
\vspace{-4mm}
\centering
\setlength{\tabcolsep}{8pt}
\renewcommand{\arraystretch}{1.1}
\caption{{Comparison with prior works.} (NV: Novel view synthesis.)}
\resizebox{.84\textwidth}{!}{
\begin{tabular}{c|cc|ccc}
\hline
    Method & Application & Representation & Supervision & Textured mesh & Arbitrary topology\\ \hline
    OccNet~\cite{occnet} & 3D generation & Implicit  & 3D & \xmark & \cmark   \\
    PointFlow~\cite{pointflow} & 3D generation & Point cloud & 3D & \xmark & \cmark  \\
    Texture3D~\cite{pavllo2021textured3dgan} & 3D generation & Mesh & 2D & \cmark & \xmark \\
    \hline
    StyleNerf~\cite{gu2021stylenerf} & 3D-aware NV & Neural field & 2D &  \xmark & \cmark  \\
    EG3D~\cite{eg3d} & 3D-aware NV & Neural field & 2D & \xmark & \cmark  \\
    PiGAN~\cite{pigan}& 3D-aware NV & Neural field & 2D & \xmark & \cmark \\
    GRAF~\cite{graf} & 3D-aware NV & Neural field & 2D & \xmark & \cmark \\
    \hline
     Ours & 3D generation & Mesh & 2D & \cmark & \cmark \\\hline
\end{tabular}
}
\label{tbl:concept_compare}
\end{table}

\section{Related Work}
We review recent advances in 3D generative models for geometry and appearance, as well as 3D-aware generative image synthesis.

\paragraph{3D Generative Models}
In recent years, 2D generative models have achieved photorealistic quality in high-resolution image synthesis \cite{stylegan,stylegan2,stylegan3,park2019SPADE, huang2021poegan, esser2021taming, dhariwal2021diffusion}. This progress has also inspired research in 3D content generation. Early approaches aimed to directly extend the 2D CNN generators to 3D voxel grids~\cite{wu2016learning,gadelha20173d,henzler2019platonicgan,lunz2020inverse,smith2017improved}, but the high memory footprint and computational complexity of 3D convolutions hinder the generation process at high resolution. As an alternative, other works have explored point cloud~\cite{achlioptas2018learning, pointflow, zhou2021pvd,mo2019structurenet}, implicit~\cite{occnet, imnet}, or octree~\cite{ibing2021octree} representations. However, these works focus mainly on generating geometry and disregard appearance. Their output representations also need to be post-processed to make them compatible with standard graphics engines.

More similar to our work, Textured3DGAN~\cite{pavllo2020convmesh,pavllo2021textured3dgan} and DIBR~\cite{dibr} generate textured 3D meshes, but they formulate the generation as a deformation of a template mesh, which prevents them from generating complex topology or shapes with varying genus, which our method can do.
PolyGen~\cite{PolyGen} and SurfGen~\cite{luo2021surfgen} can produce meshes with arbitrary topology, but do not synthesize textures.

\paragraph{3D-Aware Generative Image Synthesis}
Inspired by the success of neural volume rendering~\cite{nerf} and implicit representations~\cite{occnet,imnet}, recent work started tackling the problem of 3D-aware image synthesis~\cite{pigan,graf,giraffe,hao2021GANcraft,gu2021stylenerf,zhou2021CIPS3D,eg3d,orel2021styleSDF,Schwarz2022,xu2021volumegan}. However, neural volume rendering networks are typically slow to query, leading to long training times~\cite{pigan, graf}, and generate images of limited resolution. GIRAFFE~\cite{giraffe} and StyleNerf~\cite{ gu2021stylenerf} improve the training and rendering efficiency by performing neural rendering at a lower resolution and then upsampling the results with a 2D CNN. However, the performance gain comes at the cost of a reduced multi-view consistency. By utilizing a dual discriminator, EG3D~\cite{eg3d} can partially mitigate this problem.
Nevertheless, extracting a textured surface from methods that are based on neural rendering is a non-trivial endeavor. 
In contrast, {\ourmodel} directly outputs textured 3D meshes that can be readily used in standard graphics engines. 

\begin{figure*}[t!]
\vspace{-2mm}
\centering
\includegraphics[width=\textwidth,trim=0 100 0 100,clip]{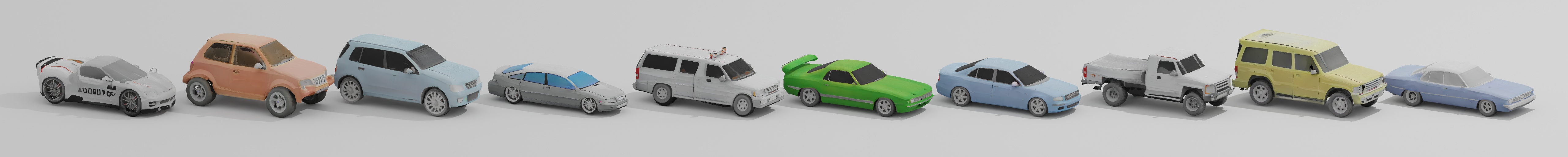}
\includegraphics[width=\textwidth,trim=0 0 0 10,clip]{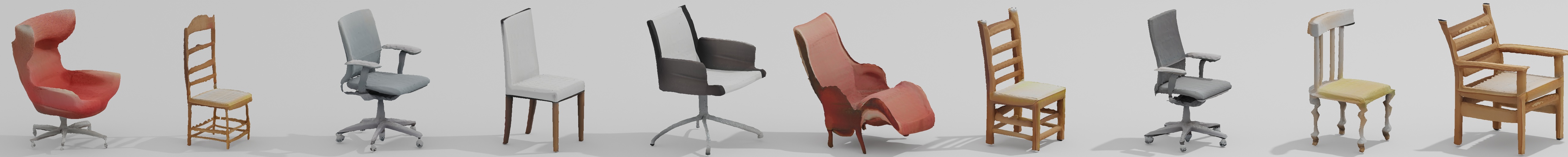}
\includegraphics[width=\textwidth,trim=0 10 0 10,clip]{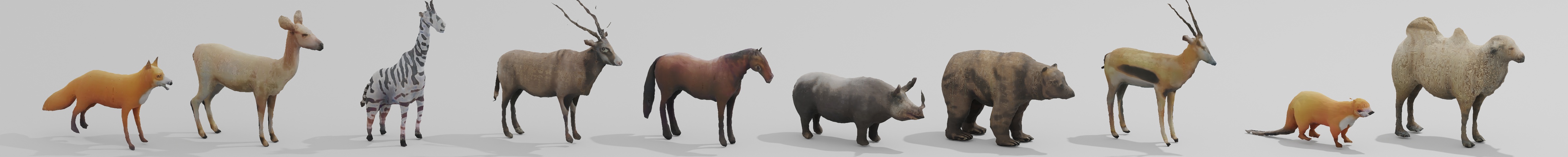}
\includegraphics[width=\textwidth,trim=0 0 0 180,clip]{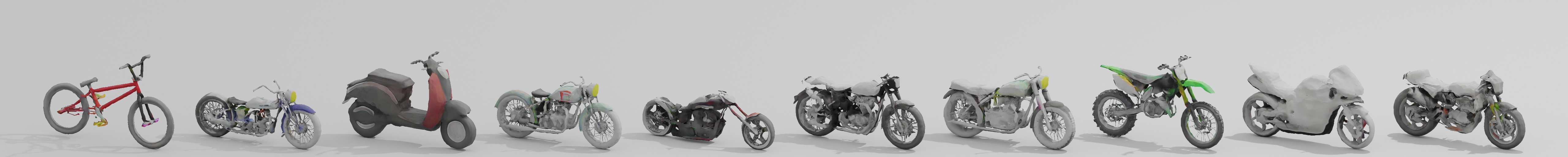}
\includegraphics[width=\textwidth,trim=0 0 0 20,clip]{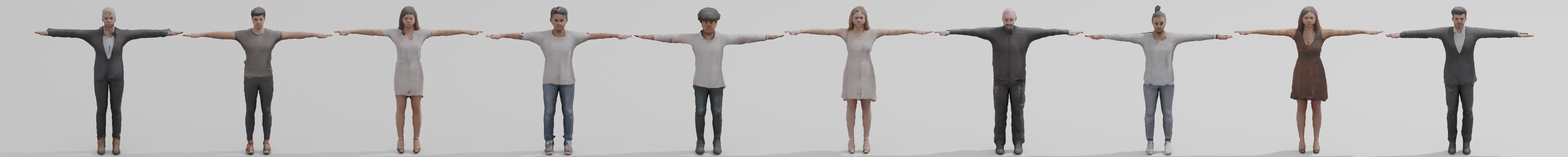}
\includegraphics[width=\textwidth,trim=0 0 0 50,clip]{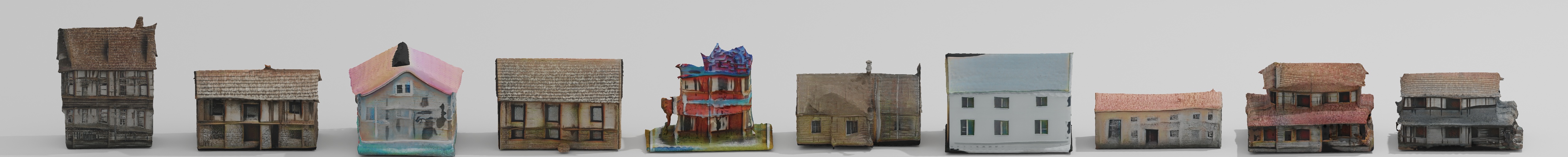}
\caption{We export our \textbf{generated shapes} and visualize them in Blender. {\ourmodel} is able to generate diverse shapes with arbitrary topology, high quality geometry, and texture.}
\label{fig:teaser_figure}
\end{figure*}

\section{Method}
\label{sec:method}
\begin{figure*}[t!]
\centering
\vspace{-5mm}
\includegraphics[width=\textwidth]{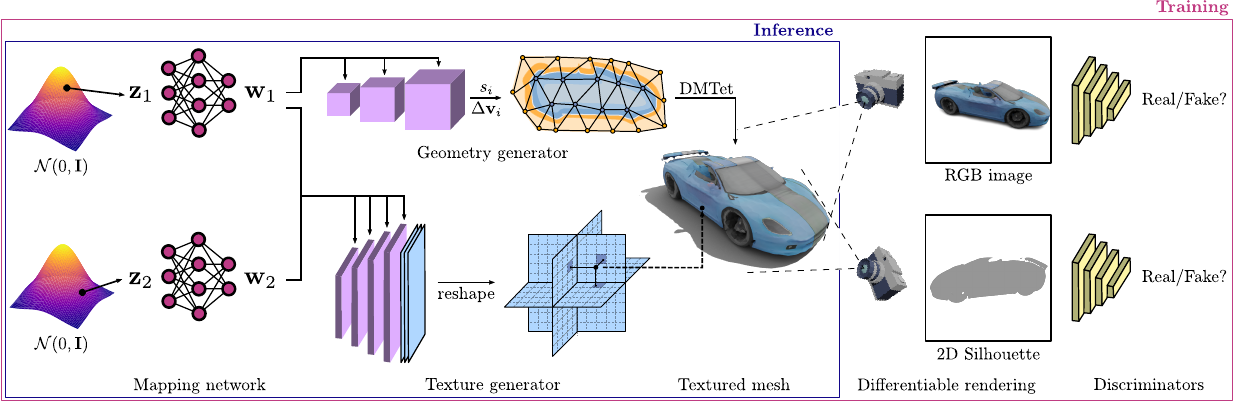}
\caption{{Overview of \textbf{{\ourmodel}:} We generate a 3D SDF and a texture field via two latent codes. We utilize DMTet~\cite{dmtet} to extract a 3D surface mesh from the SDF, and query the texture field at surface points to get colors. We train with adversarial losses defined on 2D images. In particular, we use a rasterization-based differentiable renderer~\cite{nvdiffrast} to obtain RGB images and silhouettes. We utilize two 2D discriminators, each on RGB image, and silhouette, respectively, to classify whether the inputs are real or fake. The whole model is end-to-end trainable. Note that we additionally provide an improved version of our Generator in Appendix~\ref{sec:improved_generator} and Fig.~\ref{fig:network_architecture_improved}.} 
}
\label{fig:pipeline}
\end{figure*}

We now present our {\ourmodel} framework for synthesizing textured 3D shapes. Our generation process is split into two parts: a geometry branch, which differentiably outputs a surface mesh of arbitrary topology, and a texture branch that produces a texture field that can be queried at the surface points to produce colors. The latter can be extended to other surface properties such as for example materials (Sec.~\ref{sec:material}). During training, an efficient differentiable rasterizer is utilized to render the resulting textured mesh into 2D high-resolution images. The entire process is differentiable, allowing for adversarial training from images (with masks indicating an object of interest) by propagating the gradients from the 2D discriminator to both generator branches. Our model is illustrated in Fig.~\ref{fig:pipeline}. In the following, we first introduce our 3D generator in Sec~\ref{sec:3d_generator}, before proceeding to the differentiable rendering and loss functions in Sec~\ref{sec:2d_render}.

\subsection{Generative Model of 3D Textured Meshes}
\label{sec:3d_generator}
We aim to learn a 3D generator $M, E = G(\mathbf{z})$ to map a sample from a Gaussian distribution $\mathbf{z}\in\mathcal{N}(0, \mathbf{I})$ to a mesh $M$ with texture $E$.

Since the same geometry can have different textures, and the same texture can be applied to different geometries, we sample two random input vectors $\mathbf{z}_1 \in \mathbb{R}^{512}$ and $\mathbf{z}_2 \in \mathbb{R}^{512}$. Following StyleGAN~\cite{stylegan,stylegan2,stylegan3}, we then use non-linear mapping networks $f_{\text{geo}}$ and $f_{\text{tex}}$ to map $\mathbf{z}_1$ and $\mathbf{z}_2$ to intermediate latent vectors $\textbf{w}_1 = f_{\text{geo}}(\mathbf{z}_1)$ and $\textbf{w}_2 = f_{\text{tex}}(\mathbf{z}_2)$ which are further used to produce \textit{styles} that control the generation of 3D shapes and texture, respectively. We formally introduce the generator for geometry in Sec.~\ref{sec:geometry_generator} and the texture generator in Sec.~\ref{sec:texture_generator}. 

\subsubsection{Geometry Generator}
\label{sec:geometry_generator}

We design our geometry generator to incorporate DMTet~\cite{dmtet}, a recently proposed differentiable surface representation. DMTet represents geometry as a signed distance field (SDF) defined on a deformable tetrahedral grid~\cite{gao2020deftet,gao2020beyond}, from which the surface can be differentiably recovered through marching tetrahedra~\cite{marchingtet}. Deforming the grid by moving its vertices results in a better utilization of its resolution. By adopting DMTet for surface extraction, we can produce explicit meshes with arbitrary topology and genus. We next provide a brief summary of DMTet and refer the reader to the original paper for further details.

Let ($V_T, T$) denote the full 3D space that the object lies in, where $V_T$ are the vertices in the tetrahedral grid $T$. Each tetrahedron $T_k \in T$ is defined using four vertices $\{\mathbf{v}_{a_k}, \mathbf{v}_{b_k}, \mathbf{v}_{c_k}, \mathbf{v}_{d_k}\}$, with $k\in\{1,\dots,K\}$, where $K$ is the total number of tetrahedra, and $\mathbf{v}_{i_k} \in V_T, \mathbf{v}_{i_k} \in \mathbb{R}^3$. In addition to its 3D coordinates, each vertex $\mathbf{v}_{i}$  contains the SDF value $s_i \in \mathbb{R}$ and the deformation $\Delta \mathbf{v}_{i} \in \mathbb{R}^3$ of the vertex from its initial canonical coordinate. This representation allows recovering the explicit mesh through differentiable marching tetrahedra~\cite{dmtet}, where SDF values in continuous space are computed by a barycentric interpolation of their value $s_i$ on the deformed vertices $\mathbf{v}^{\prime}_i=\mathbf{v}_i + \Delta \mathbf{v}_{i}$. 

\paragraph{Network Architecture} We map $\mathbf{w}_1 \in \mathbb{R}^{512}$ to SDF values and deformations at each vertex $\mathbf{v}_{i}$ through a series of conditional 3D convolutional and fully connected layers. Specifically, we first use 3D convolutional layers to generate a feature volume conditioned on $\mathbf{w}_1$. We then query the feature at each vertex $\mathbf{v}_{i} \in V_T$ using trilinear interpolation and feed it into MLPs that outputs the SDF value $s_i$ and the deformation $\Delta \mathbf{v}_{i}$. In cases where modeling at a high-resolution is required (e.g. motorbike with thin structures in the wheels), we further use volume subdivision following~\cite{dmtet}. 

\paragraph{Differentiable Mesh Extraction}
After obtaining $s_i$ and $\Delta \mathbf{v}_{i}$ for all the vertices, we use the differentiable marching tetrahedra algorithm to extract the explicit mesh. Marching tetrahedra determines the surface topology within each tetrahedron based on the signs of $s_i$. In particular, a mesh face is extracted when $\text{sign}(s_i) \neq \text{sign}(s_j)$, where $i, j$ denotes the indices of vertices in the edge of tetrahedron, and the vertices $\mathbf{m}_{i,j}$ of that face are determined by a linear interpolation as $\mathbf{m}_{i,j} = \frac{\mathbf{v}^{\prime}_i s_j - \mathbf{v}^{\prime}_j s_i}{ s_j - s_i}$.
Note that the above equation is only evaluated when $s_i \neq s_j$, thus it is differentiable, and the gradient from $\mathbf{m}_{i, j}$ can be back-propagated into the SDF values $s_i$ and deformations $\Delta \mathbf{v}_{i}$. With this representation, the shapes with arbitrary topology can easily be generated by predicting different signs of $s_i$.

\subsubsection{Texture Generator}
\label{sec:texture_generator}
Directly generating a texture map consistent with the output mesh is not trivial, as the generated shape can have an arbitrary genus and topology. We thus parameterize the texture as a texture field~\cite{oechsle2019texture}. Specifically, we model the texture field with a function $f_t$ that maps the 3D location of a surface point $\mathbf{p} \in \mathbb{R}^3$, conditioned on the $\mathbf{w}_2 $, to the RGB color $\mathbf{c} \in \mathbb{R}^3$ at that location. Since the texture field depends on geometry, we additionally condition this mapping on the geometry latent code $\mathbf{w}_1$, such that $\mathbf{c} = f_t(\mathbf{p}, \mathbf{w}_1 \oplus \mathbf{w}_2)$, where $\oplus$ denotes concatenation. 

\paragraph{Network Architecture} We represent our texture field using a tri-plane representation, which is efficient and expressive in reconstructing 3D objects~\cite{conv_occnet} and generating 3D-aware images~\cite{eg3d} . 
Specifically, we follow~\cite{eg3d, stylegan2} and use a conditional 2D convolutional neural network to map the latent code $\mathbf{w}_1\oplus\mathbf{w}_2$ to
three axis-aligned orthogonal feature planes of size $N\times N \times (C\times3)$, where $N=256$ denotes the spatial resolution and $C=32$ the number of channels.

Given the feature planes, the feature vector $\mathbf{f}^t \in \mathbb{R}^{32}$ of a surface point $\mathbf{p}$ can be recovered as $\mathbf{f}^t = \sum_e \rho(\pi_e(\mathbf{p}))$, where $\pi_e(\mathbf{p})$ is the projection of the point $\mathbf{p}$ to the feature plane $e$ and $\rho(\cdot)$ denotes bilinear interpolation of the features. An additional fully connected layer is then used to map the aggregated feature vector $\mathbf{f}^t$ to the RGB color $\mathbf{c}$. Note that, different from other works on 3D-aware image synthesis~\cite{eg3d,gu2021stylenerf,pigan,graf} that also use a neural field representation, we only need to sample the texture field at the locations of the surface points (as opposed to dense samples along a ray). This greatly reduces the computational complexity for rendering high-resolution images and guarantees to generate multi-view consistent images by construction.

\subsection{Differentiable Rendering and Training}
\label{sec:2d_render}
In order to supervise our model during training, we draw inspiration from Nvdiffrec~\cite{nvdiffrec} that performs multi-view 3D object reconstruction by utilizing a differentiable renderer. Specifically, we render the extracted 3D mesh and the texture field into 2D images using a differentiable renderer~\cite{nvdiffrast}, and supervise our network with a 2D discriminator, which tries to distinguish the image from a real object or rendered from the generated object. 

\paragraph{Differentiable Rendering} 
We assume that the camera distribution $\mathcal{C}$ that was used to acquire the images in the dataset is known. To render the generated shapes, we randomly sample a camera $c$ from  $\mathcal{C}$, and utilize a highly-optimized differentiable rasterizer Nvdiffrast~\cite{nvdiffrast} to render the 3D mesh into a 2D silhouette as well as an image where each pixel contains the coordinates of the corresponding 3D point on the mesh surface. These coordinates are further used to query the texture field to obtain the RGB values.  Since we operate directly on the extracted mesh, we can render high-resolution images with high efficiency, allowing our model to be trained with image resolution as high as 1024$\times$1024.

\paragraph{Discriminator \& Objective} We train our model using an adversarial objective. We adopt the discriminator architecture from StyleGAN~\cite{stylegan}, and use the same non-saturating GAN objective with R1 regularization~\cite{Mescheder2018ICML}. We empirically find that using two separate discriminators, one for RGB images and another one for silhouettes, yields better results than a single discriminator operating on both. Let $D_x$ denote the discriminator, where $x$ can either be an RGB image or a silhouette. The adversarial objective is then be defined as follows:
\begin{eqnarray}
L(D_x, G) = \mathbb{E}_{\mathbf{z}\in \mathcal{N}, c\in \mathcal{C}} [g(D_x(R(G(\mathbf{z}), c)))] + \mathbb{E}_{I_x\in p_x} [g(- D_x(I_x)) + \lambda ||\nabla D_x(I_x)||^2_2],
\label{eq:loss}
\end{eqnarray}
where $g(u)$ is defined as $g(u) = -\log (1 + \exp(-u))$, $p_x$ is the distribution of real images, $R$ denotes rendering, and  $\lambda$ is a hyperparameter. Since $R$ is differentiable, the gradients can be backpropagated from 2D images to our 3D generators.

\paragraph{Regularization} To remove internal floating faces that are not visible in any of the views, we further regularize the geometry generator with a cross-entropy loss defined between the SDF values of the neighboring vertices~\cite{nvdiffrec}: 
\begin{eqnarray}
L_{\mathrm{reg}} = \sum_{{i,j} \in \mathbb{S}_e}
H\left(\sigma(s_i), \text{sign}\left(s_j\right)\right) + H\left(\sigma(s_j), \text{sign}\left(s_i\right)\right),
\label{eq:reg}
\end{eqnarray}
where $H$ denotes binary cross-entropy loss and $\sigma$ denotes the sigmoid function. The sum in Eq.~\ref{eq:reg} is defined over the set of unique edges $\mathbb{S}_e$ in the tetrahedral grid, for which $\text{sign}(s_i) \neq \text{sign}(s_j)$. 

The overall loss function is then defined as:
\begin{eqnarray}
L = L(D_\text{rgb},  G) + L(D_\text{mask}, G) + \mu L_{\mathrm{reg}},
\label{eq:all_loss}
\end{eqnarray}
where $\mu$ is a hyperparameter that controls the level of regularization. 

\begin{table*}[t!]
\begin{minipage}{0.48\textwidth}
\centering
\renewcommand{\arraystretch}{1.3}
\begin{adjustbox}{width=\linewidth}
\begin{tabular}{lccccccc}
\toprule
\multirow{2}{*}{Category} &\multirow{2}{*}{Method} & \multicolumn{2}{c}{COV (\%, $\uparrow$)} & \multicolumn{2}{c}{MMD ($\downarrow$)} & \multicolumn{2}{c}{FID ($\downarrow$)}\\\cmidrule(lr){3-4}\cmidrule(lr){5-6}\cmidrule(lr){7-8}
&& LFD & CD & LFD & CD & Ori & 3D\\
\midrule
\multirow{8}{*}{Car}&PointFlow~\cite{pointflow} &51.91 & 57.16 & 1971 & 0.82& - & - \\
&OccNet~\cite{occnet} & 27.29 & 42.63 & 1717 & \textbf{0.61}& - & - \\ \cline{2-8}
&Pi-GAN~\cite{pigan} &0.82 & 0.55 & 6626 & 25.54 &  52.82&104.29\\
&GRAF~\cite{graf} & 1.57 & 1.57 & 6012 & 10.63&  49.95& 52.85 \\
&EG3D~\cite{eg3d} & 60.16 & 49.52 & 1527 & 0.72&  15.52& 21.89\\ \cline{2-8}
&Ours  &\textbf{66.78} & \textbf{58.39}& \textbf{1491} & 0.71& \textbf{10.25} & \textbf{10.25} \\ 
& Ours+Subdiv. & 62.48 & 55.93 & 1553 & 0.72& 12.14 & 12.14 \\ 
\cline{2-8}
& Ours (improved $G$) & 59.00 & 47.95 & 1473 & 0.81& 10.60 & 10.60 \\ 

\midrule
\multirow{8}{*}{Chair}&PointFlow~\cite{pointflow} & 49.58 & \textbf{71.87} & 3755 & \textbf{3.03}& - & -\\
&OccNet~\cite{occnet} &61.10 & 67.13 & 3494 & 3.98& - & - \\ \cline{2-8}
&Pi-GAN~\cite{pigan} &53.76 & 39.65 & 4092 & 6.65 & 65.70 & 120.53\\
&GRAF~\cite{graf} &50.23 & 39.28 & 4055 & 6.80& 43.82 &61.63\\
&EG3D~\cite{eg3d} & 58.31 & 50.14 & 3444 & 4.72 &38.87 &46.06\\ \cline{2-8}
&Ours & 69.08 &  69.91 & 3167 & 3.72& 23.28 & 23.28 \\ 
 & Ours+Subdiv. & \textbf{71.59} & 70.84 & \textbf{3163} & 3.95& \textbf{23.17} & \textbf{23.17} \\ 
\cline{2-8}
 & Ours (improved $G$) & 71.96 & 71.96 & 3125 & 3.96 & 22.41 & 22.41\\ 
\bottomrule
\end{tabular}
\end{adjustbox}
\end{minipage}
\quad
\begin{minipage}{0.48\textwidth}
\centering
\renewcommand{\arraystretch}{1.3}
\begin{adjustbox}{width=\linewidth}
\begin{tabular}{lccccccc}
\toprule
\multirow{2}{*}{Category}&\multirow{2}{*}{Method} & \multicolumn{2}{c}{COV (\%, $\uparrow$)} & \multicolumn{2}{c}{MMD ($\downarrow$)} & \multicolumn{2}{c}{FID ($\downarrow$)}\\\cmidrule(lr){3-4}\cmidrule(lr){5-6}\cmidrule(lr){7-8}
&& LFD & CD & LFD & CD &  Ori & 3D\\
\midrule
\multirow{8}{*}{Mbike} &PointFlow~\cite{pointflow} &50.68 & 63.01 & 4023 & \textbf{1.38} & - & - \\
&OccNet~\cite{occnet} & 30.14 & 47.95 & 4551 & 2.04 & - & - \\ \cline{2-8}
&Pi-GAN~\cite{pigan} &2.74 & 6.85 & 8864 & 21.08 & 72.67 &131.38\\
&GRAF~\cite{graf} & 43.84 & 50.68 & 4528 & 2.40 & 83.20 & 113.39\\
&EG3D~\cite{eg3d} & 38.36 & 34.25 & 4199 & 2.21&66.38  &89.97\\ \cline{2-8}
&Ours & \textbf{67.12} &\textbf{67.12}& 3631 &1.72 &65.60 & 65.60 \\ 
&Ours+Subdiv. & 63.01 & 61.64 & \textbf{3440} & 1.79 & \textbf{54.12} & \textbf{54.12} \\ 
\cline{2-8}
& Ours (improved $G$)& 69.86 & 65.75 & 3393 & 1.79 & 48.90 & 48.90 \\ 
\midrule
\multirow{8}{*}{Animal}&PointFlow~\cite{pointflow} &42.70 & 74.16 & 4885 & \textbf{1.68} & - & - \\
&OccNet~\cite{occnet} & 56.18 & 75.28 & 4418 & 2.39& - & - \\ \cline{2-8}
&Pi-GAN~\cite{pigan} &31.46 & 30.34 & 6084 & 8.37& 36.26&150.86\\
&GRAF~\cite{graf} & 60.67 & 61.80 & 5083 & 4.81& 42.07 & 52.48\\
&EG3D~\cite{eg3d} &74.16 & 58.43 & 4889 & 3.42 & 40.03& 83.47\\ \cline{2-8}
& Ours &\textbf{79.77} & \textbf{78.65}&  \textbf{3798}& 2.02& \textbf{28.33}&\textbf{28.33}  \\ 
& Ours+Subdiv. & 66.29 & 74.16 & 3864 & 2.03 & 28.49 & 28.49 \\
\cline{2-8}
& Ours (improved $G$) & 74.16 & 82.02 & 3767 & 1.97 & 27.18 & 27.18 \\ 
\bottomrule
\end{tabular}
\end{adjustbox}
\end{minipage}
\caption{{\bf Quantitative evaluation of generation results}:  $\uparrow$: the higher the better, $\downarrow$: the lower the better. The best scores are highlighted in bold. MMD-CD scores are multiplied by $10^3$. The results of \textit{Ours (improved $G$)} were obtained after the review process by improving the design of the generator network architecture $G$ (see Appendix~\ref{sec:improved_generator} for more details). 
}
\label{tbl:evaluation_with_baselines}
\end{table*}

\vspace{-3mm}
\begin{figure*}[t!]
\centering
\includegraphics[width=1\textwidth,trim=0 0 0 0,clip]{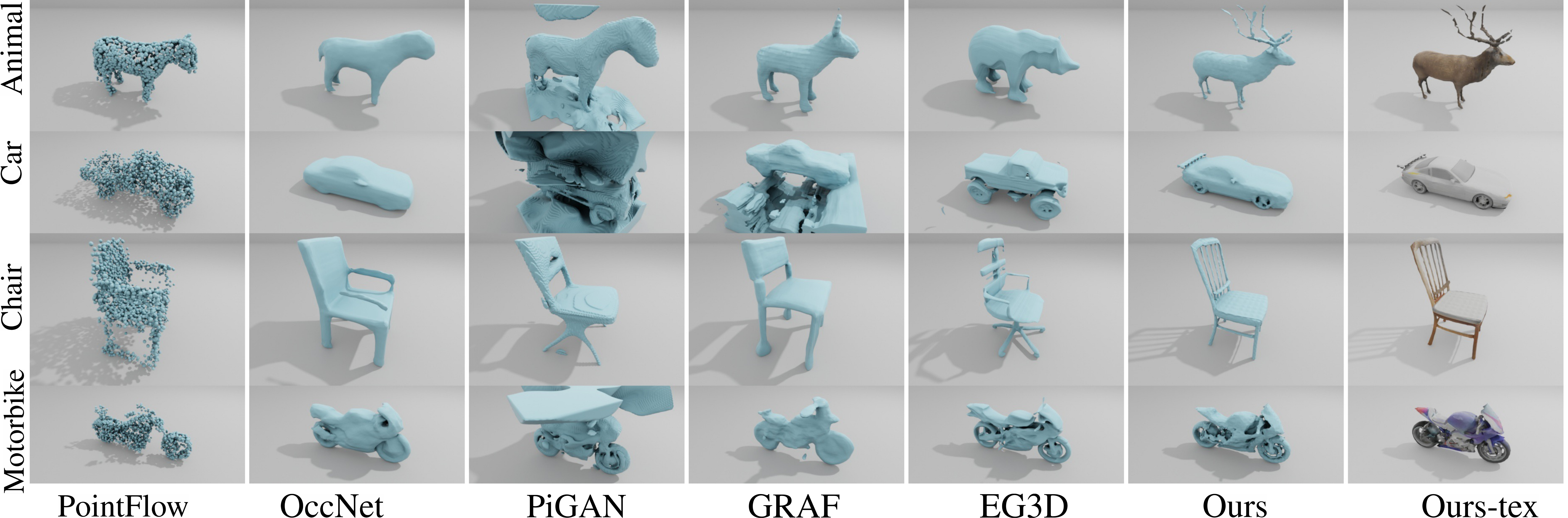}
\caption{Qualitative comparison of {\ourmodel} to the baseline methods in terms of extracted 3D geometry. {\ourmodel} is able to generate shapes with much higher geometric detail across all categories.}
\label{fig:our_results_geo}
\end{figure*}

\section{Experiments}
\label{sec:exp}
We conduct extensive experiments to evaluate our model. We first compare the quality of the 3D textured meshes generated by {\ourmodel} to the existing methods using the ShapeNet~\cite{shapenet} and Turbosquid~\cite{Turbosquid}
datasets. Next, we ablate our design choices in Sec.~\ref{sec:exp_ablation}. Finally, we demonstrate the flexibility of {\ourmodel} by adapting it to downstream applications in Sec.~\ref{sec:exp_application}. Additional experimental results and implementation details are provided in Appendix. 

\subsection{Experiments on Synthetic Datasets}
\label{sec:exp_synthetic}
\paragraph{Datasets} For evaluation on ShapeNet~\cite{shapenet}, we use three categories with complex geometry -- \emph{Car}, \emph{Chair}, and \emph{Motorbike}, which contain 7497, 6778, and 337 shapes, respectively. We randomly split each category into training (70\%), validation (10 \%), and test (20 \%), and further remove from the test set shapes that have duplicates in the training set. To render the training data, we randomly sample camera poses from the upper hemisphere of each shape. For the \emph{Car} and \emph{Chair} categories, we use 24 random views, while for \emph{Motorbike} we use 100 views due to less number of shapes. As models in ShapeNet only have simple textures, we also evaluate {\ourmodel} on an \emph{Animal} dataset (442 shapes) collected from TurboSquid~\cite{Turbosquid}, where textures are more detailed and we split it into training, validation and test as defined above. Finally, to demonstrate the versatility of {\ourmodel}, we also provide qualitative results on the \emph{House} dataset collected from Turbosquid (563 shapes), and \emph{Human Body} dataset from Renderpeople~\cite{renderpeople} (500 shapes). We train a separate model on each category.

\paragraph{Baselines} We compare {\ourmodel} to two groups of works: {\bf 1)} 3D generative models that rely on 3D supervision: PointFlow~\cite{pointflow} and OccNet~\cite{occnet}. Note that these methods only generate geometry without texture. {\bf 2)} 3D-aware image generation methods: GRAF~\cite{graf}, PiGAN~\cite{pigan}, and EG3D~\cite{eg3d}.

\paragraph{Metrics} To evaluate the quality of our synthesis, we consider both the geometry and texture of the generated shapes. For geometry, we adopt metrics from~\cite{achlioptas2018learning} and use both Chamfer Distance (CD) and Light Field Distance~\cite{chen2003visual} (LFD) to compute the Coverage score and Minimum Matching Distance. For OccNet~\cite{occnet}, GRAF~\cite{graf}, PiGAN~\cite{pigan} and EG3D~\cite{eg3d}, we use marching cubes to extract the underlying geometry. For PointFlow~\cite{pointflow}, we use Poisson surface reconstruction to convert a point cloud into a mesh when evaluating LFD. To evaluate texture quality, we adopt the FID~\cite{heusel2017gans} metric commonly used to evaluate image synthesis. In particular, for each category, we render the test shapes into 2D images, and also render the generated 3D shapes from each model into 50k images using the same camera distribution. We then compute FID on the two image sets. As the baselines from 3D-aware image synthesis~\cite{graf,pigan,eg3d} do not directly output textured meshes, we compute FID score in two ways: (\textbf{i}) we use their neural volume rendering to obtain 2D images, which we refer to as FID-Ori, and (\textbf{ii}) we extract the mesh from their neural field representation using marching cubes, render it, and then use the 3D location of each pixel to query the network to obtain the RGB values. We refer to this score, that is more aware of the actual 3D shape, as FID-3D. Further details on the evaluation metrics are available in the Appendix~\ref{sec:evaluation_metrics}.

\begin{figure*}[t!]
\centering
\includegraphics[width=1\textwidth,trim=0 0 0 0,clip]{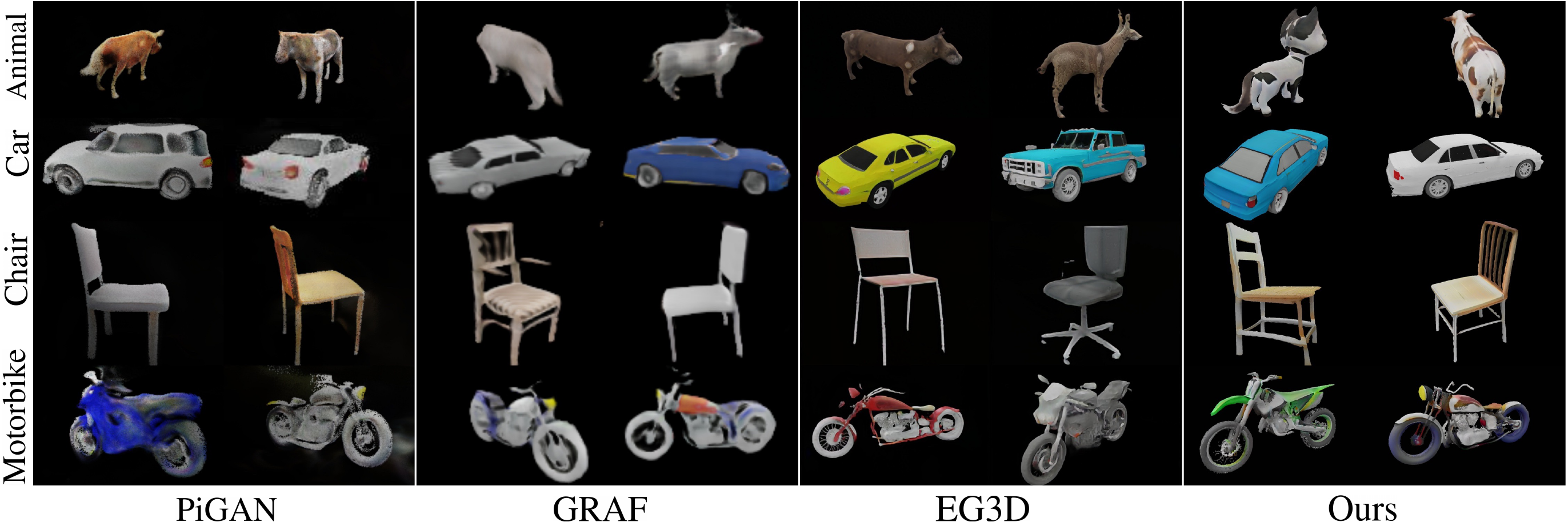}
\caption{Qualitative comparison of {\ourmodel} to the baseline methods in terms of generated 2D images. {\ourmodel} generates sharp textures with high level of detail.}
\label{fig:compare_gen_img}
\end{figure*}

\begin{figure*}[t!]
\centering
\includegraphics[width=0.24\textwidth,trim=0 0 0 0,clip]{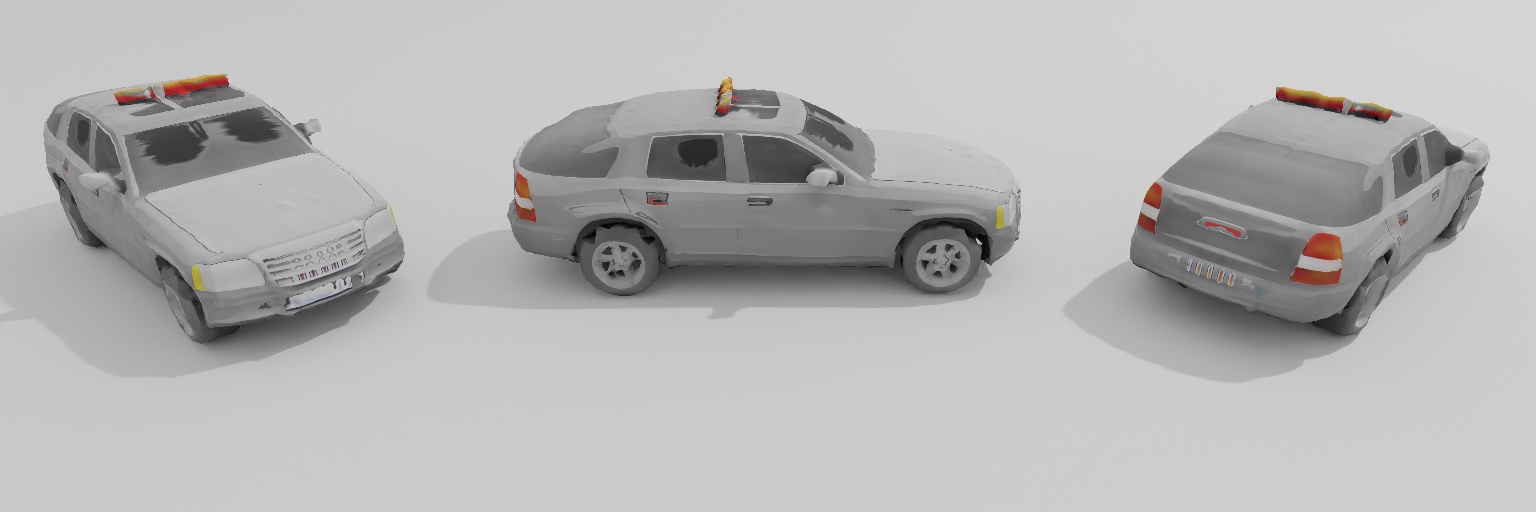}
\includegraphics[width=0.24\textwidth,trim=0 0  0 0,clip]{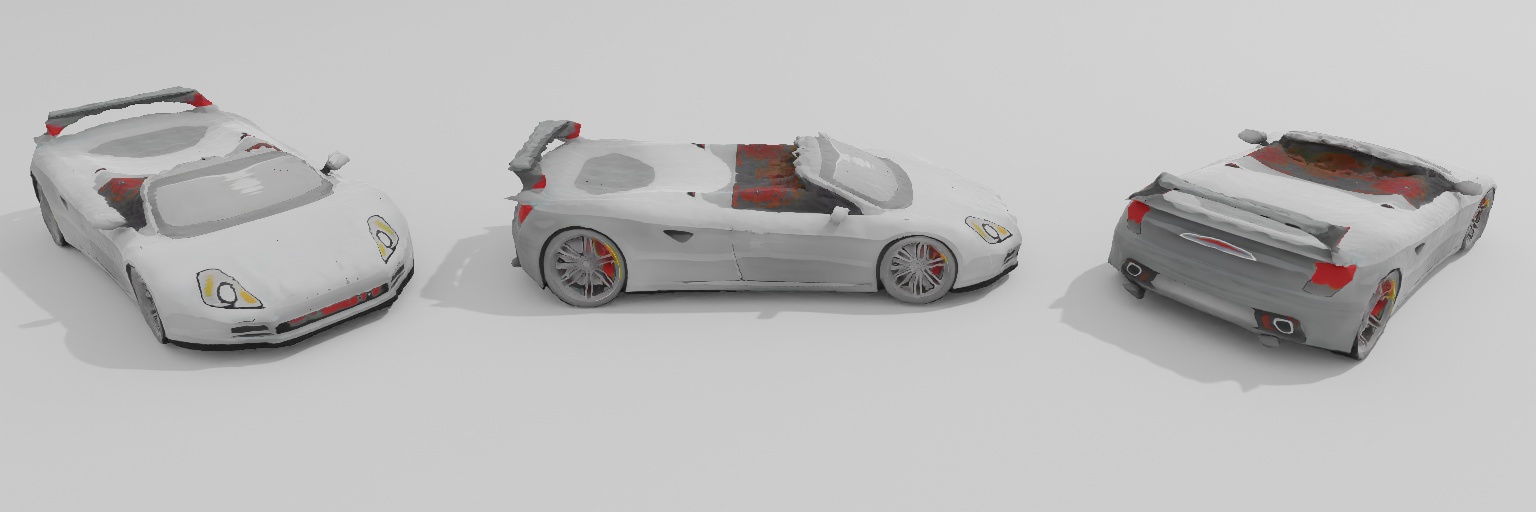}
\includegraphics[width=0.24\textwidth,trim=0 0 0 0,clip]{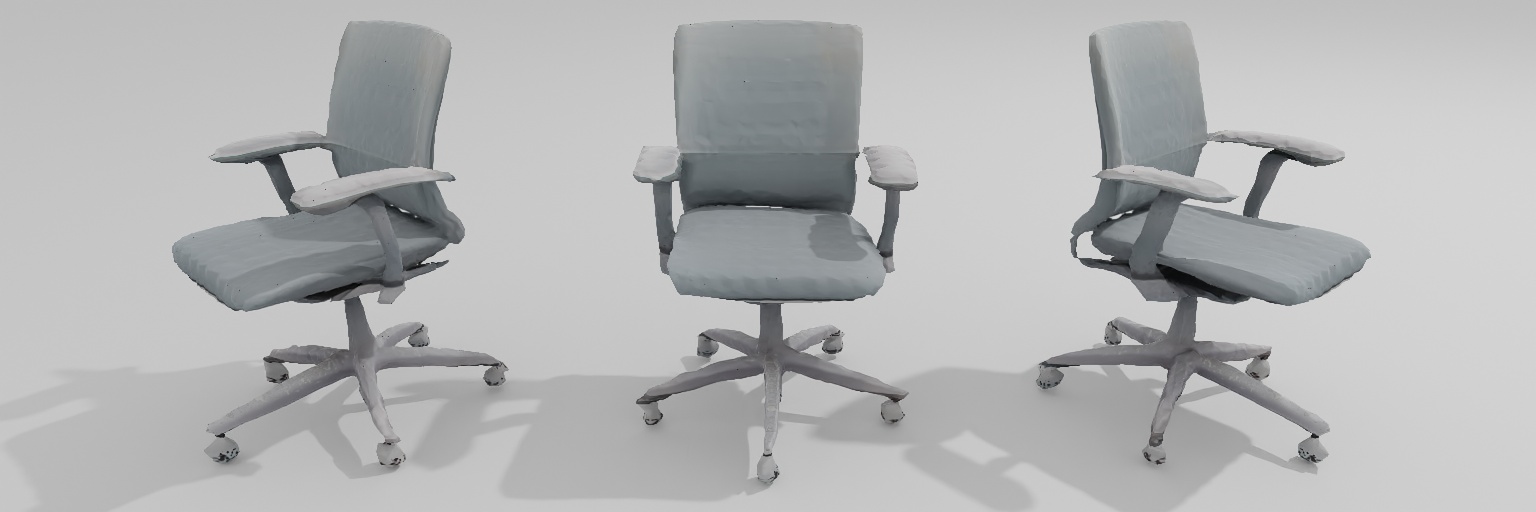}
\includegraphics[width=0.24\textwidth,trim=0 0 0 0,clip]{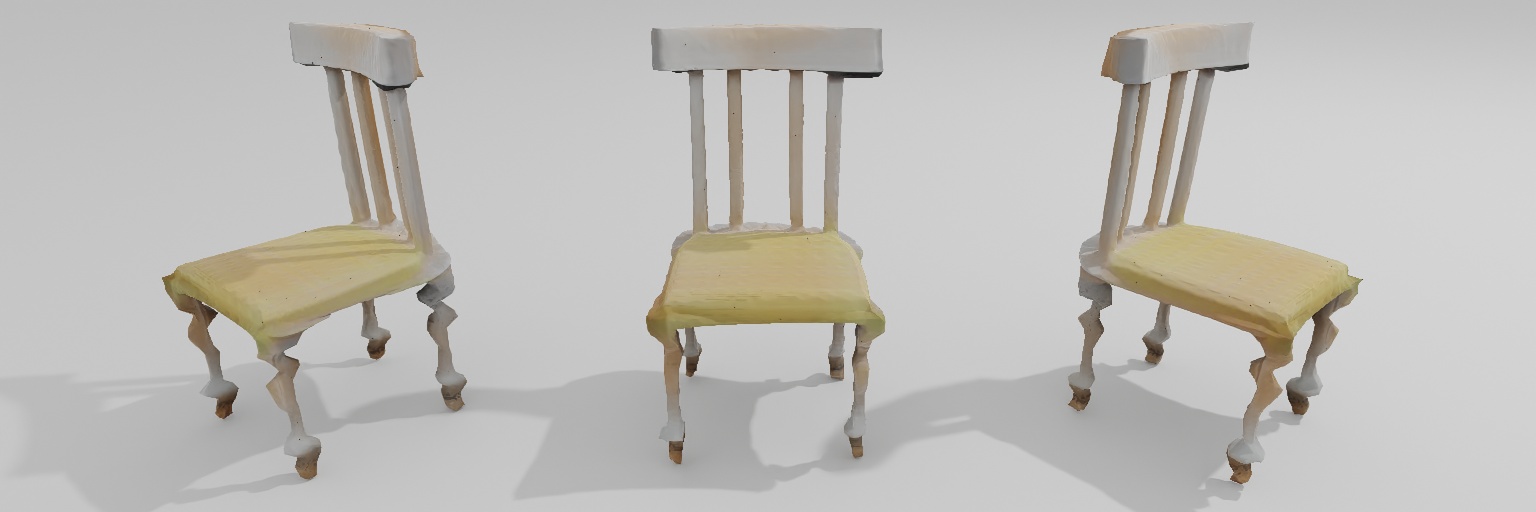}
\includegraphics[width=0.24\textwidth,trim=0 0  0 0,clip]{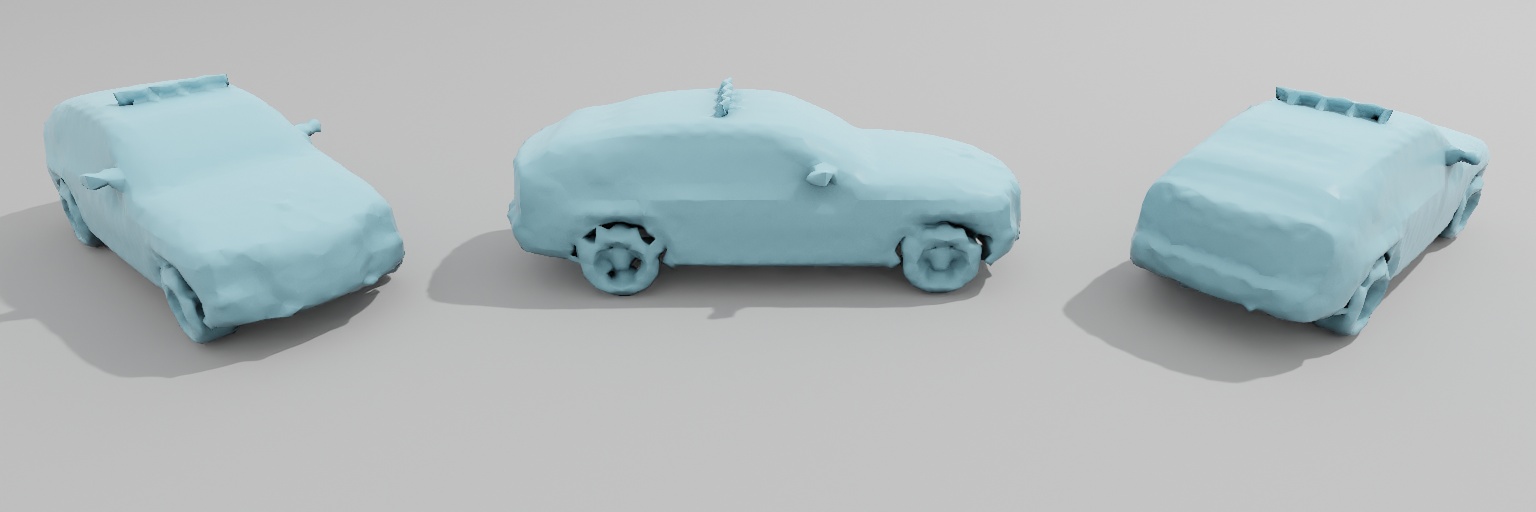}
\includegraphics[width=0.24\textwidth,trim=0 0  0 0,clip]{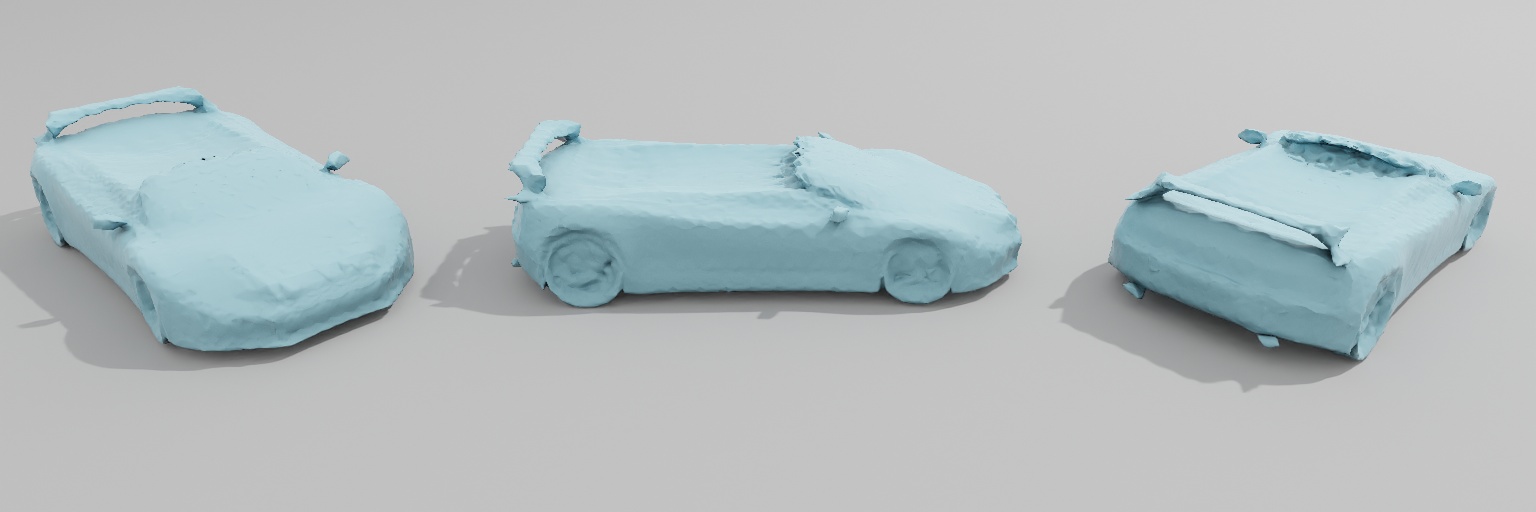}
\includegraphics[width=0.24\textwidth,trim=0 0 0 0,clip]{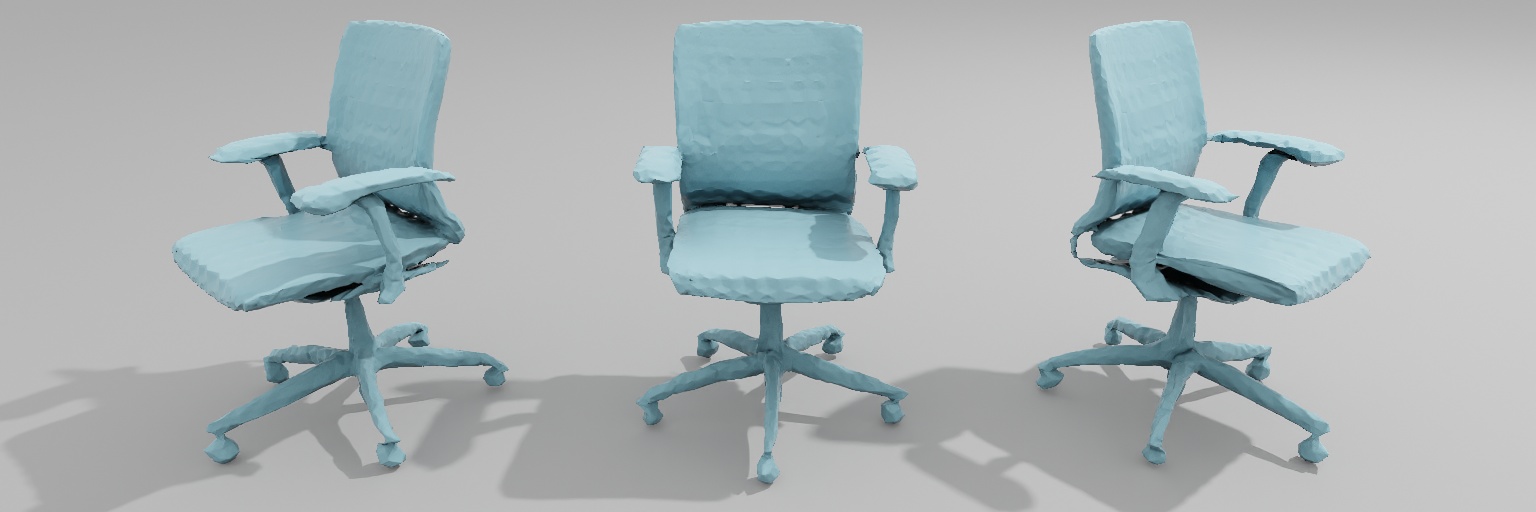}
\includegraphics[width=0.24\textwidth,trim=0 0 0 0,clip]{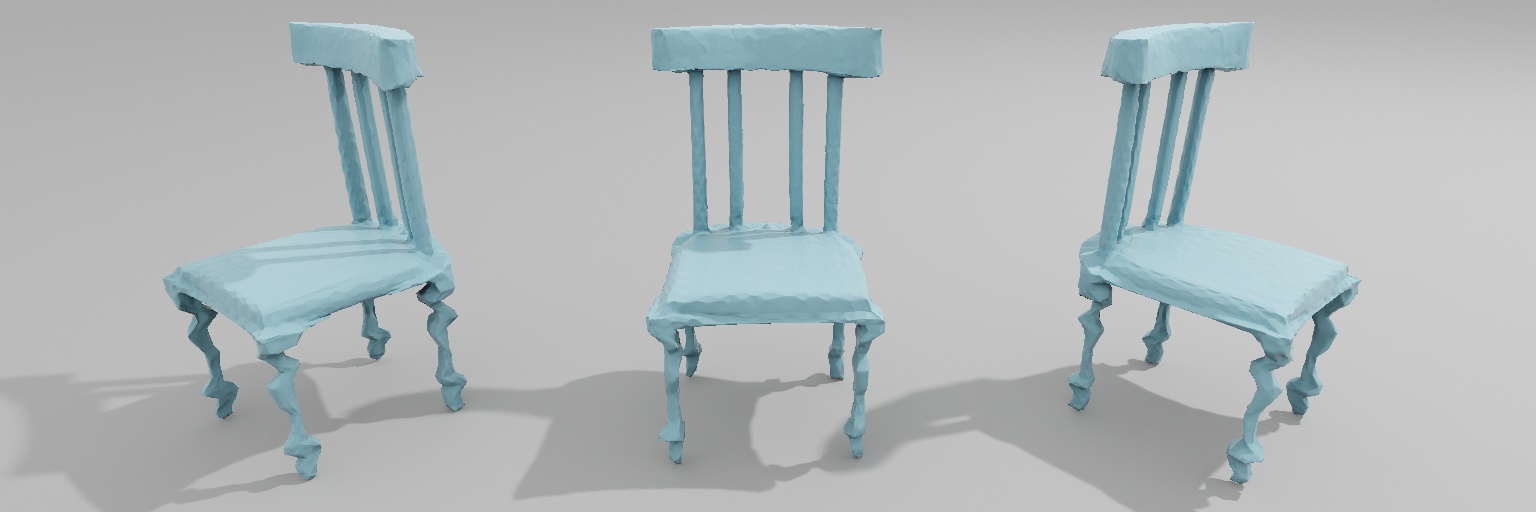}
\includegraphics[width=0.24\textwidth,trim=0 50 0 50,clip]{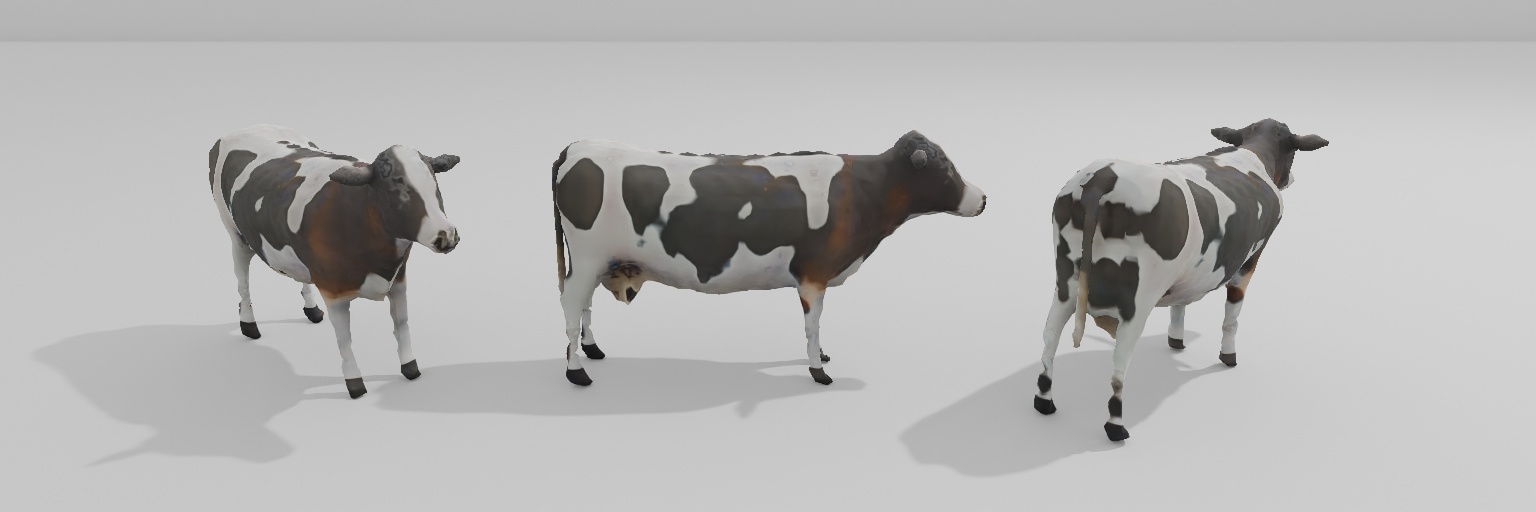}
\includegraphics[width=0.24\textwidth,trim=0 50 0 50,clip]{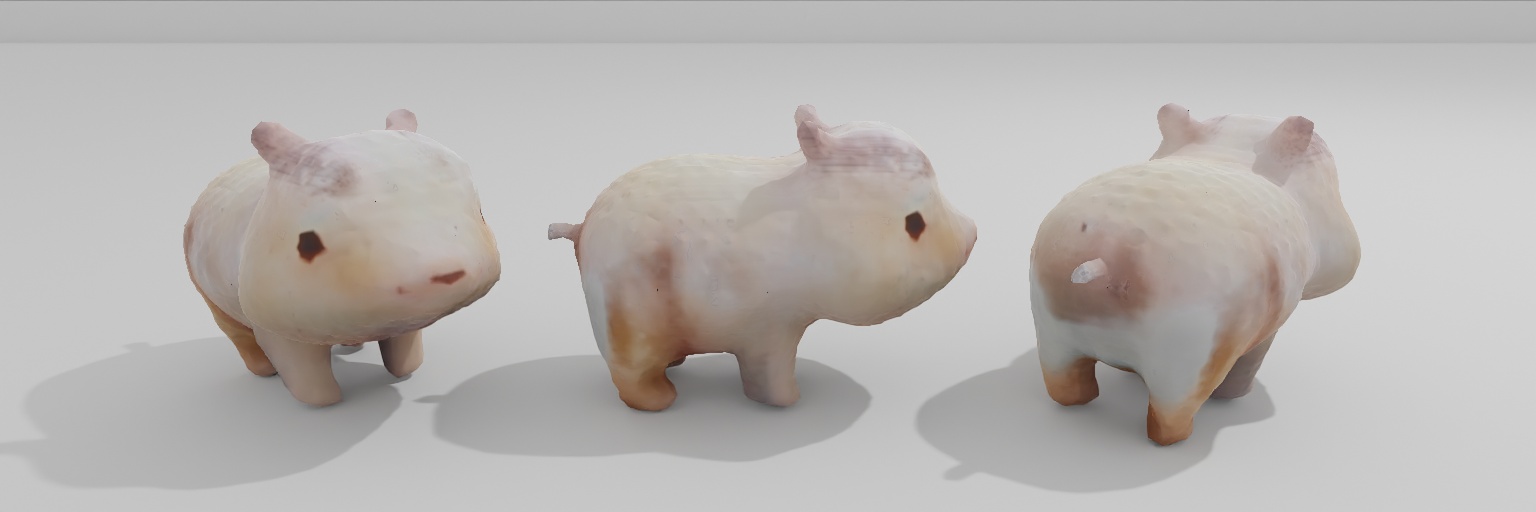}
\includegraphics[width=0.24\textwidth,trim=0 50 0 50,clip]{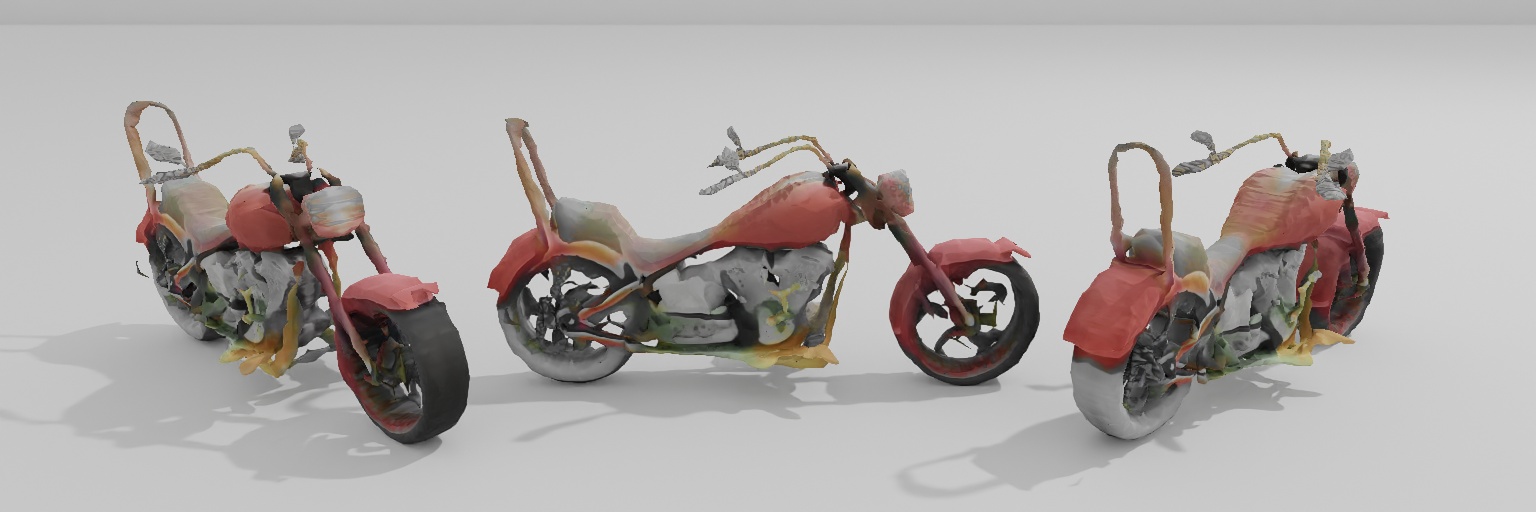}
\includegraphics[width=0.24\textwidth,trim=0 50 0 50,clip]{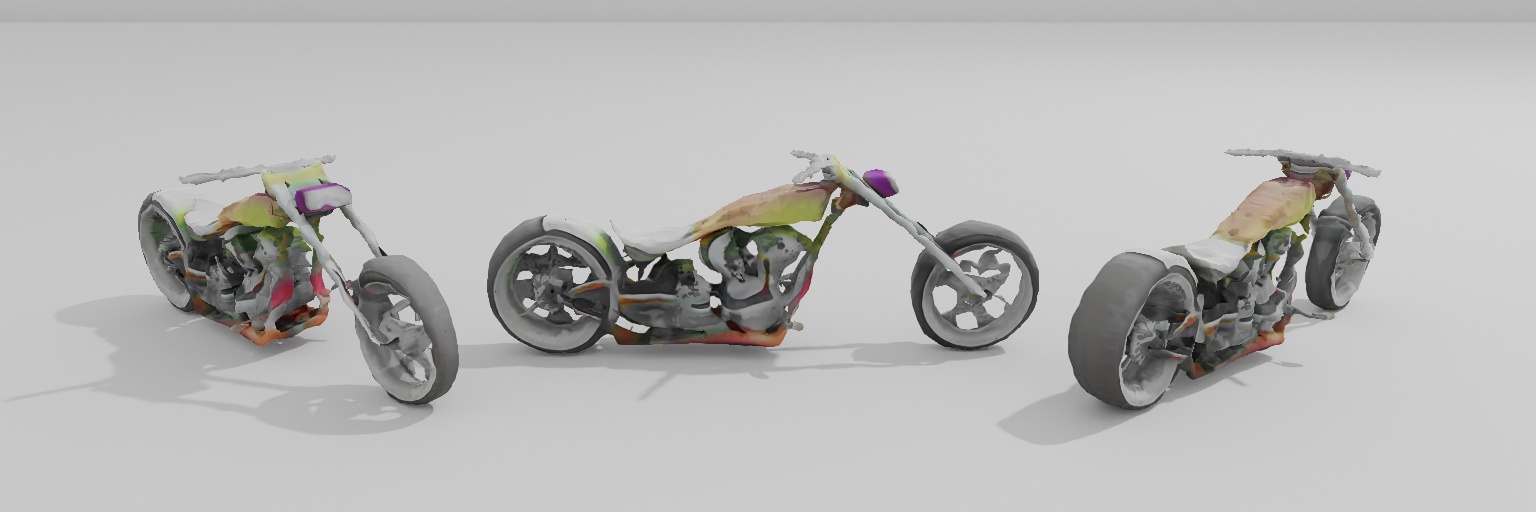}
\includegraphics[width=0.24\textwidth,trim=0 50 0 50,clip]{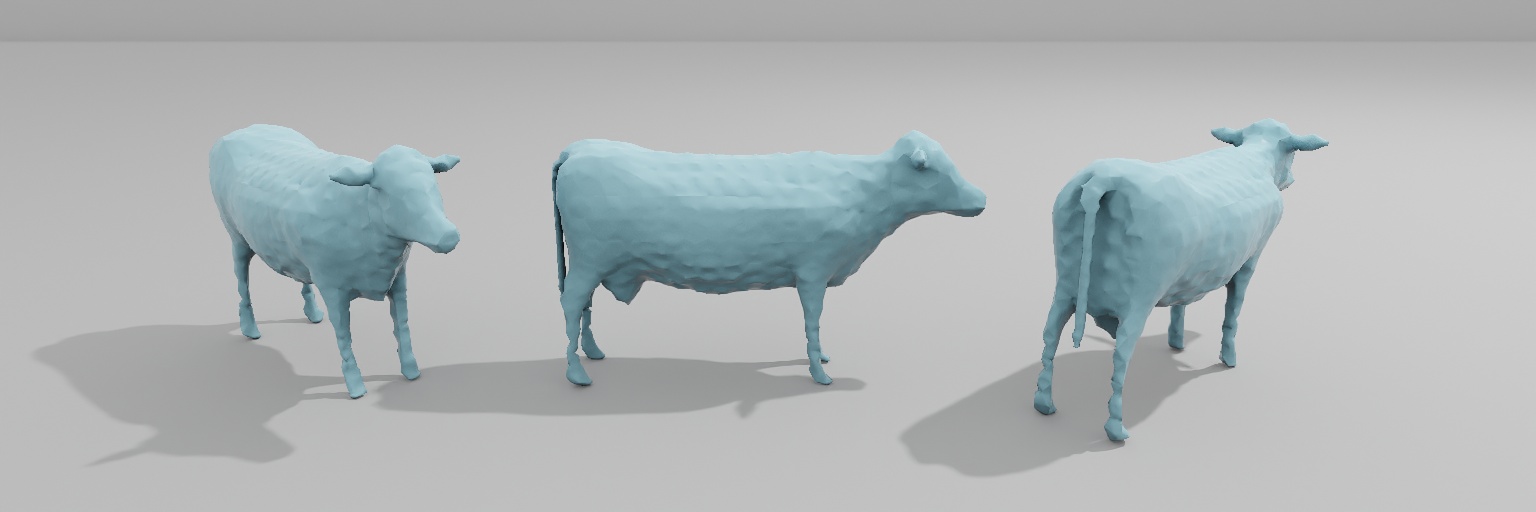}
\includegraphics[width=0.24\textwidth,trim=0 50 0 50,clip]{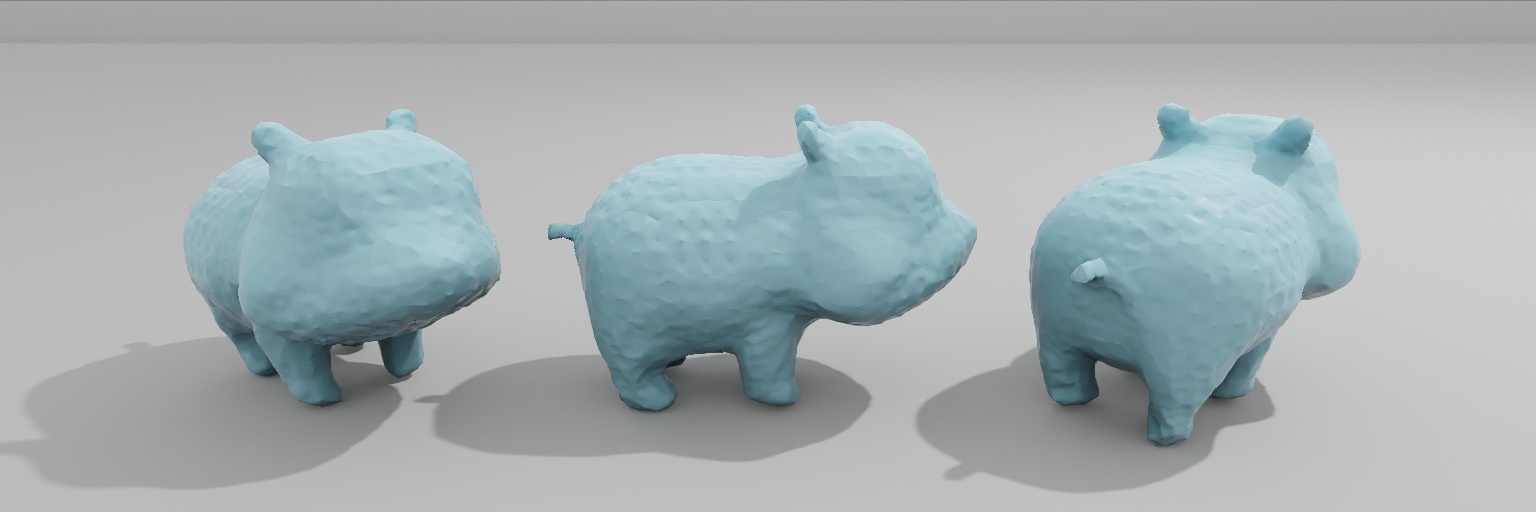}
\includegraphics[width=0.24\textwidth,trim=0 50 0 50,clip]{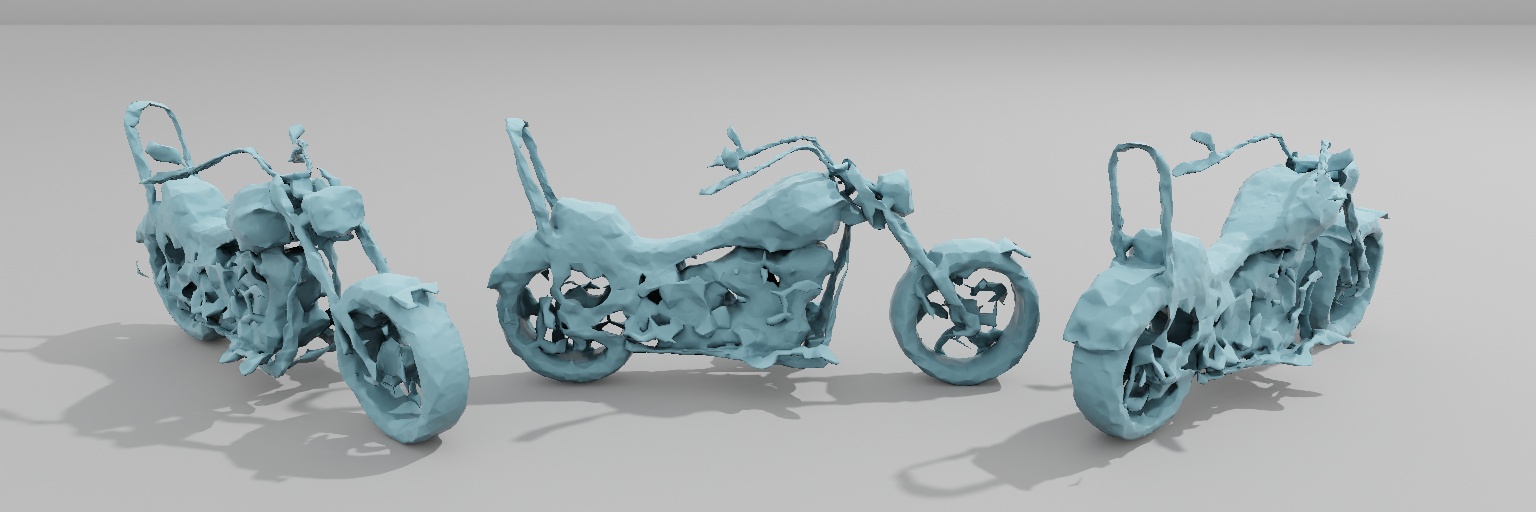}
\includegraphics[width=0.24\textwidth,trim=0 50 0 50,clip]{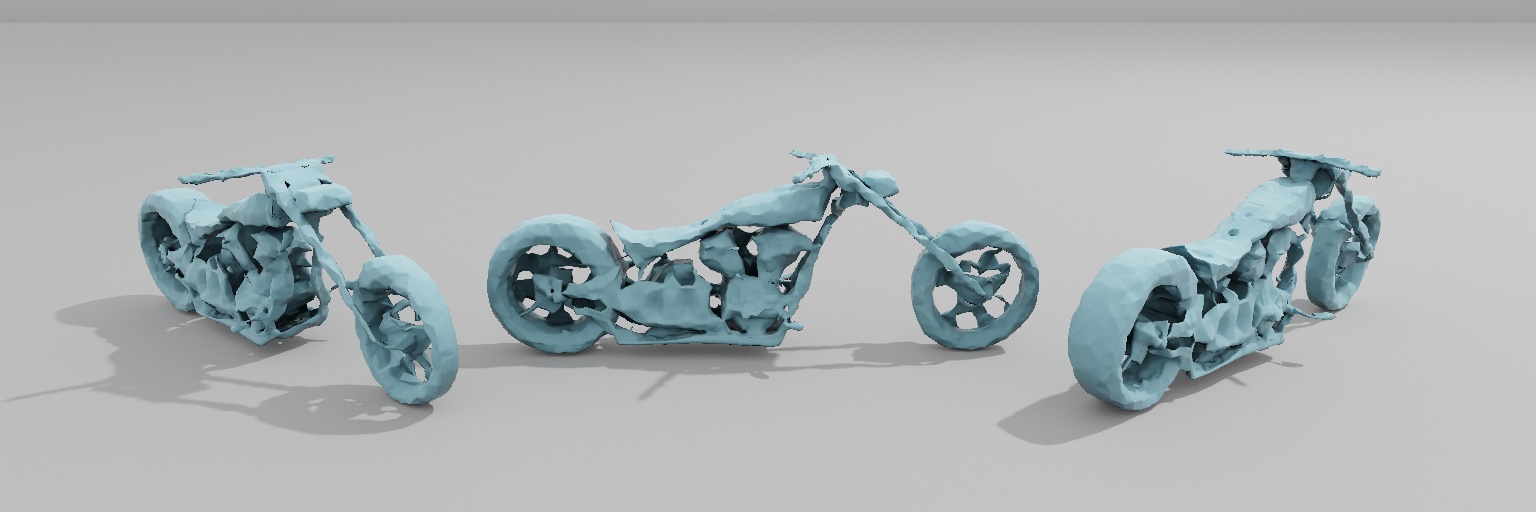}
\includegraphics[width=0.24\textwidth,trim=0 0 0 0,clip]{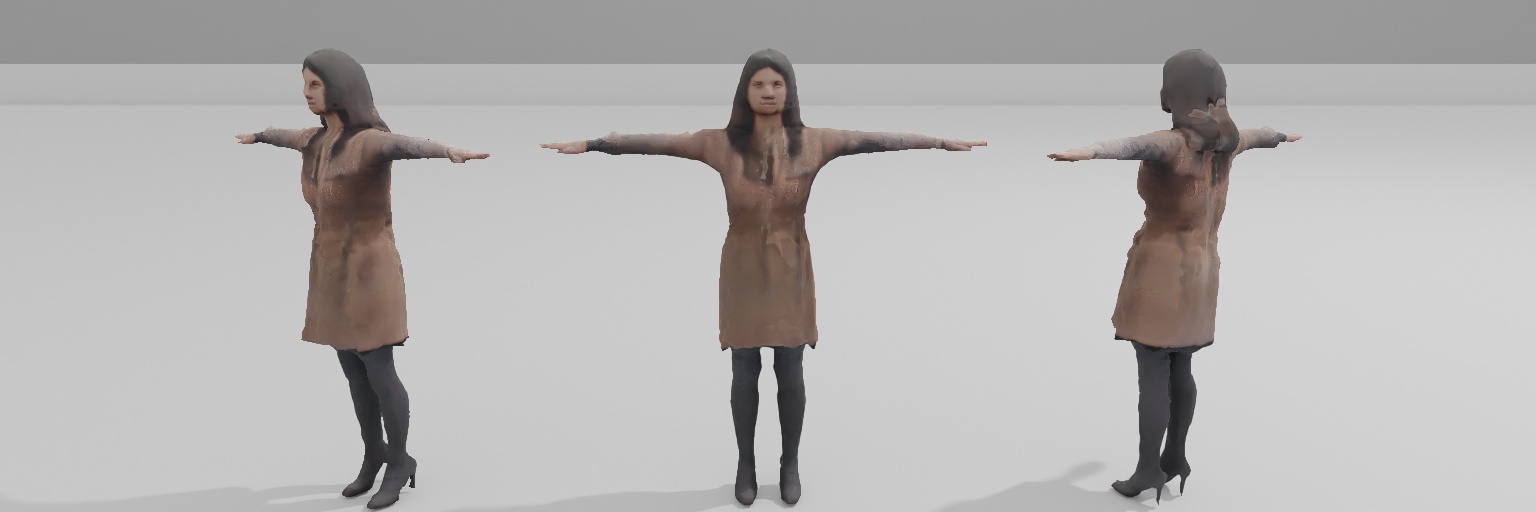}
\includegraphics[width=0.24\textwidth,trim=0 0 0 0,clip]{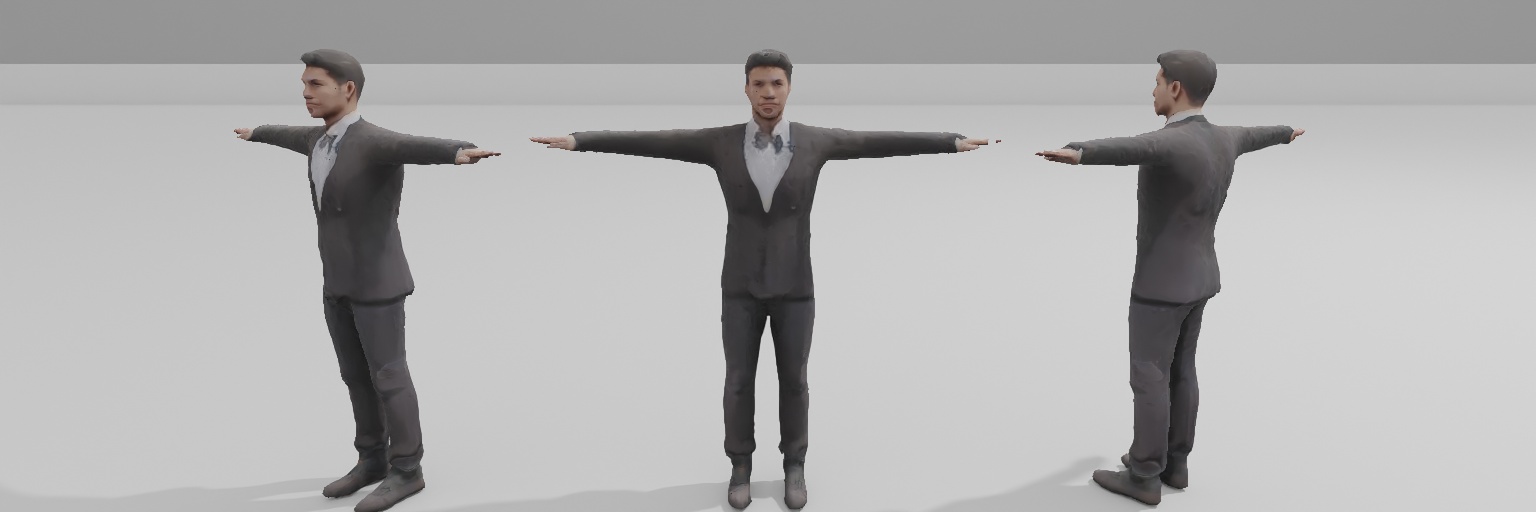}
\includegraphics[width=0.24\textwidth,trim=0 0 0 0,clip]{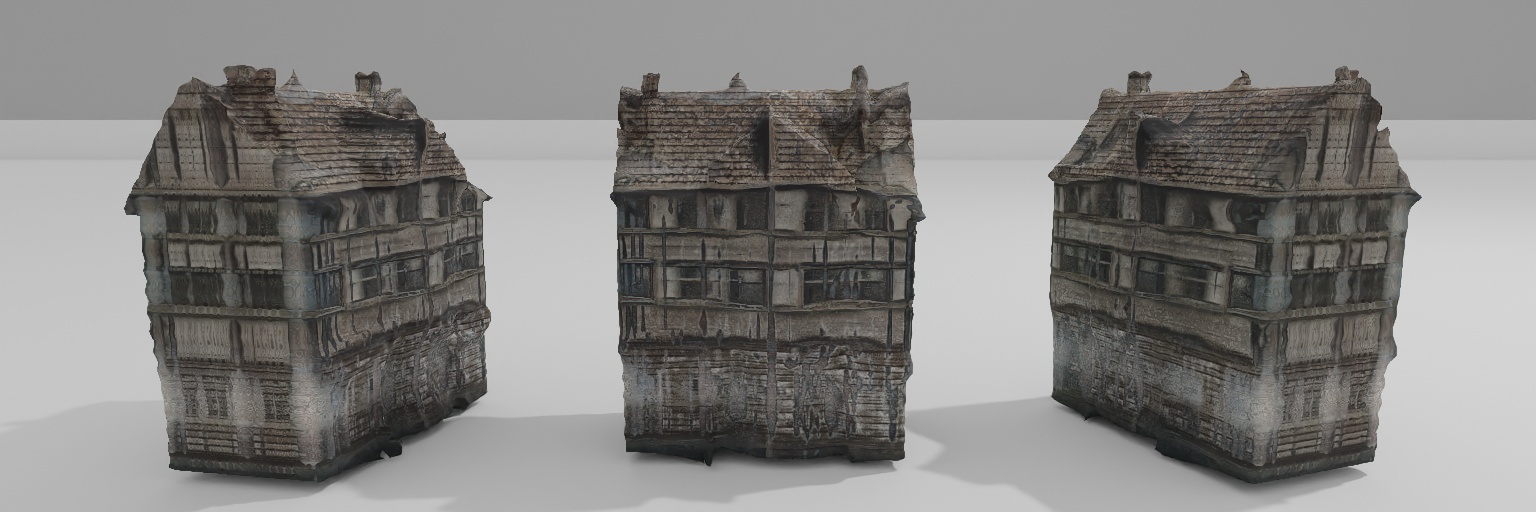}
\includegraphics[width=0.24\textwidth,trim=0 0 0 0,clip]{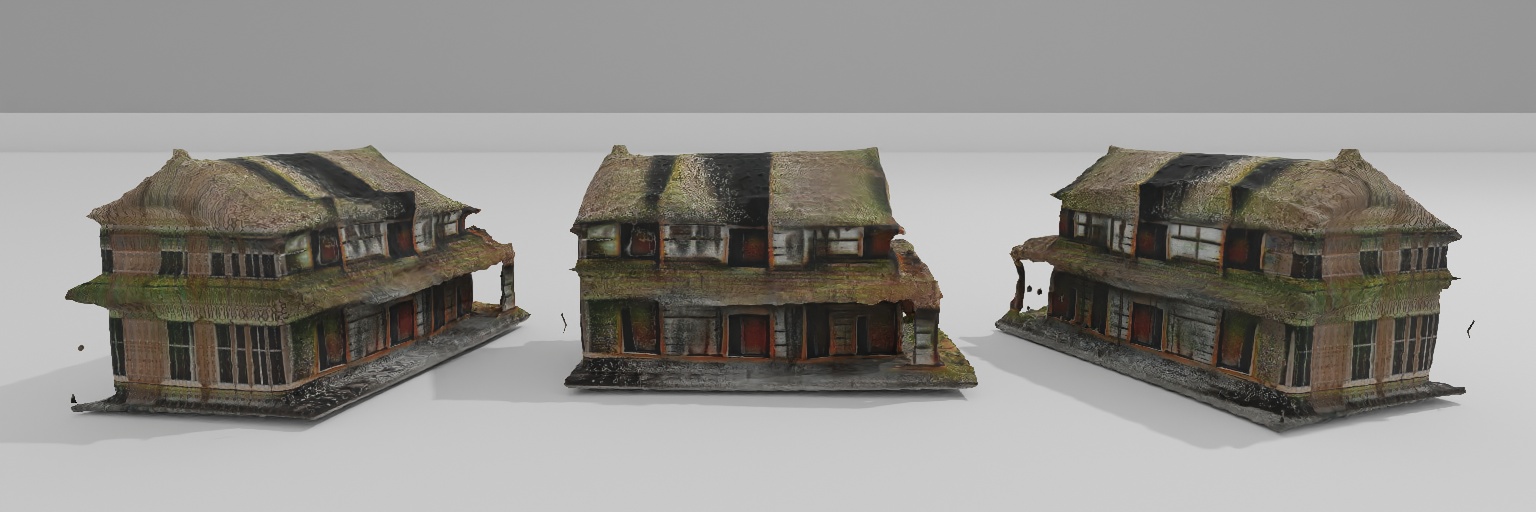}
\includegraphics[width=0.24\textwidth,trim=0 0 0 0,clip]{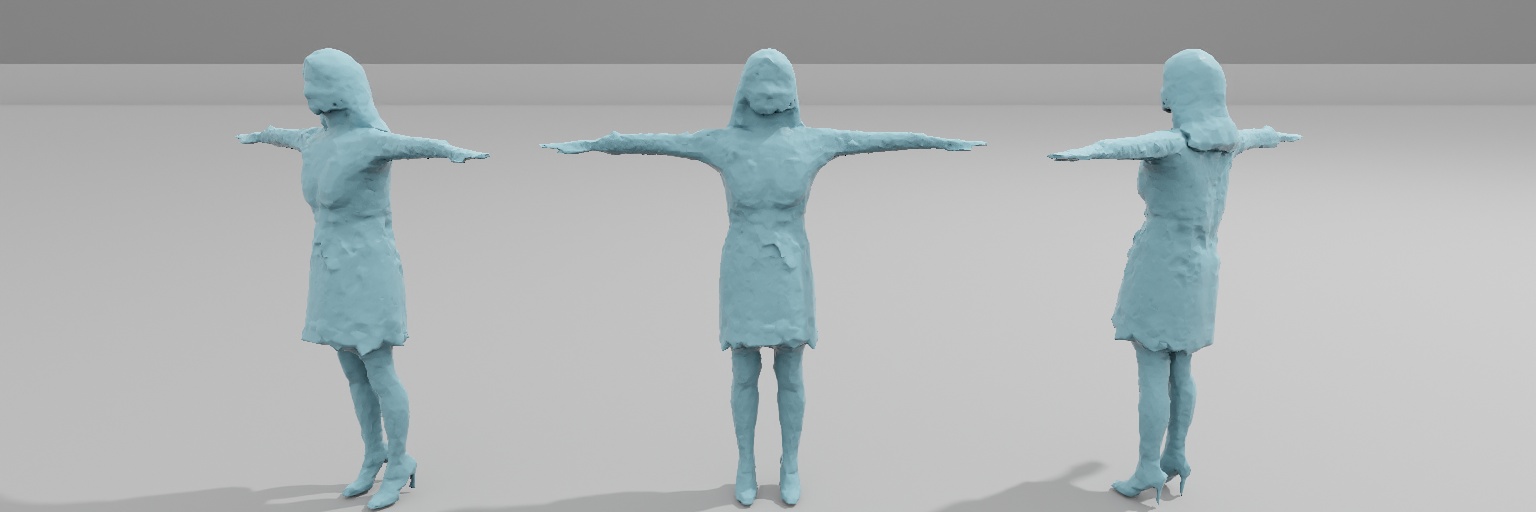}
\includegraphics[width=0.24\textwidth,trim=0 0 0 0,clip]{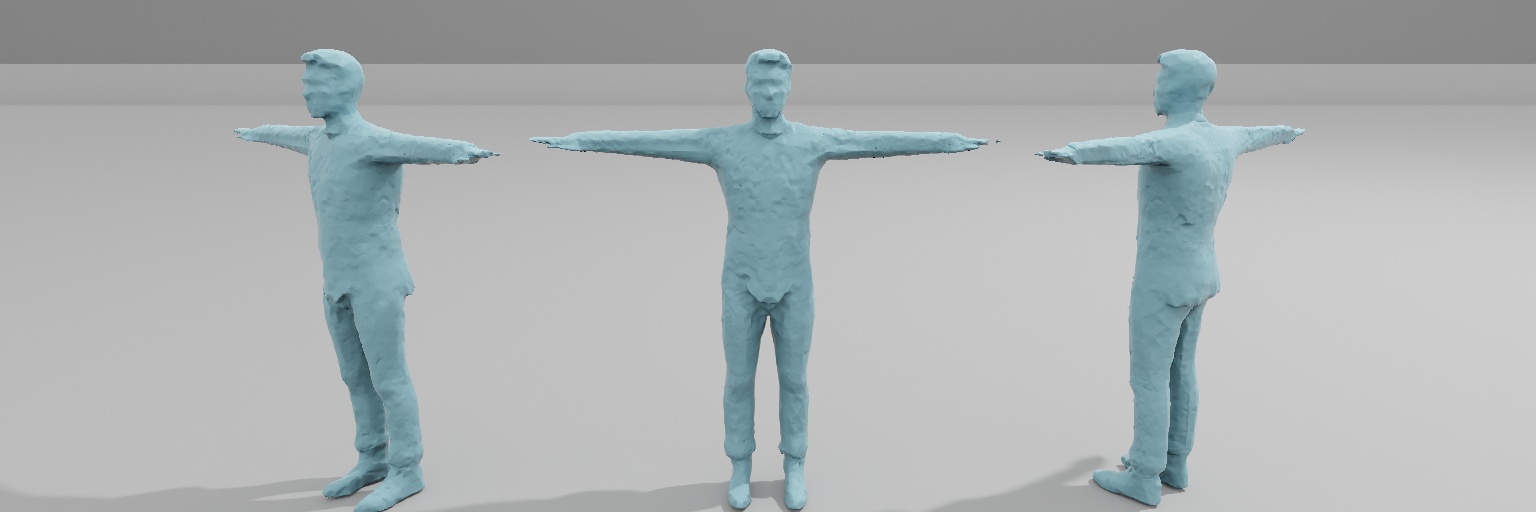}
\includegraphics[width=0.24\textwidth,trim=0 0 0 0,clip]{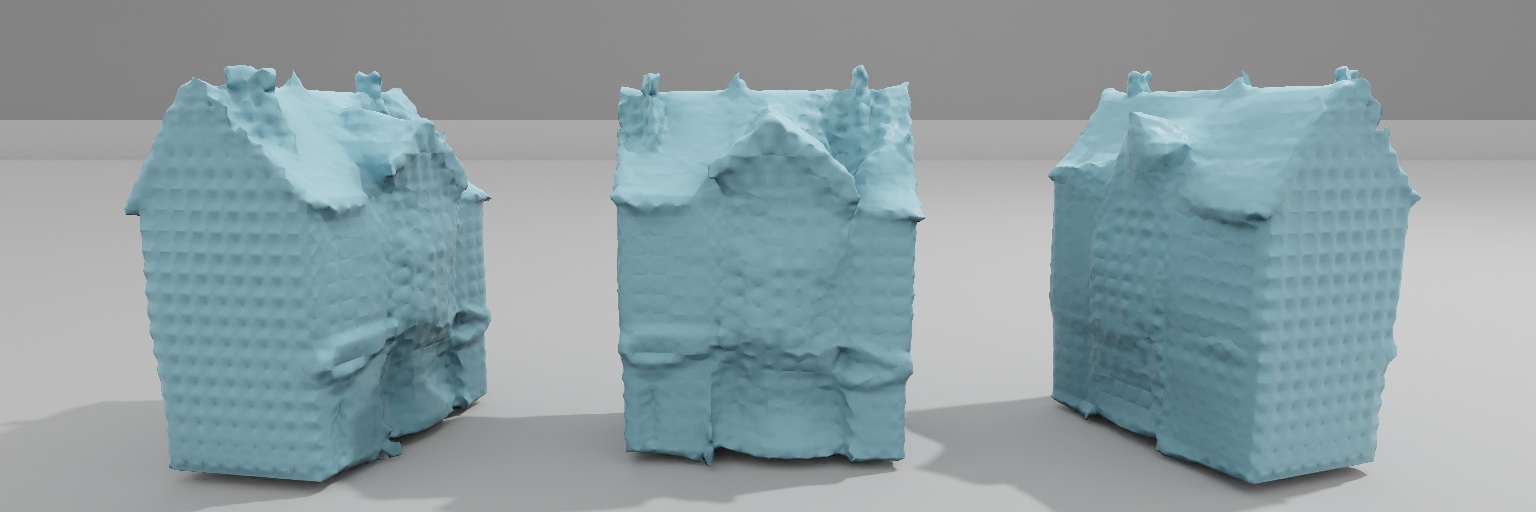}
\includegraphics[width=0.24\textwidth,trim=0 0 0 0,clip]{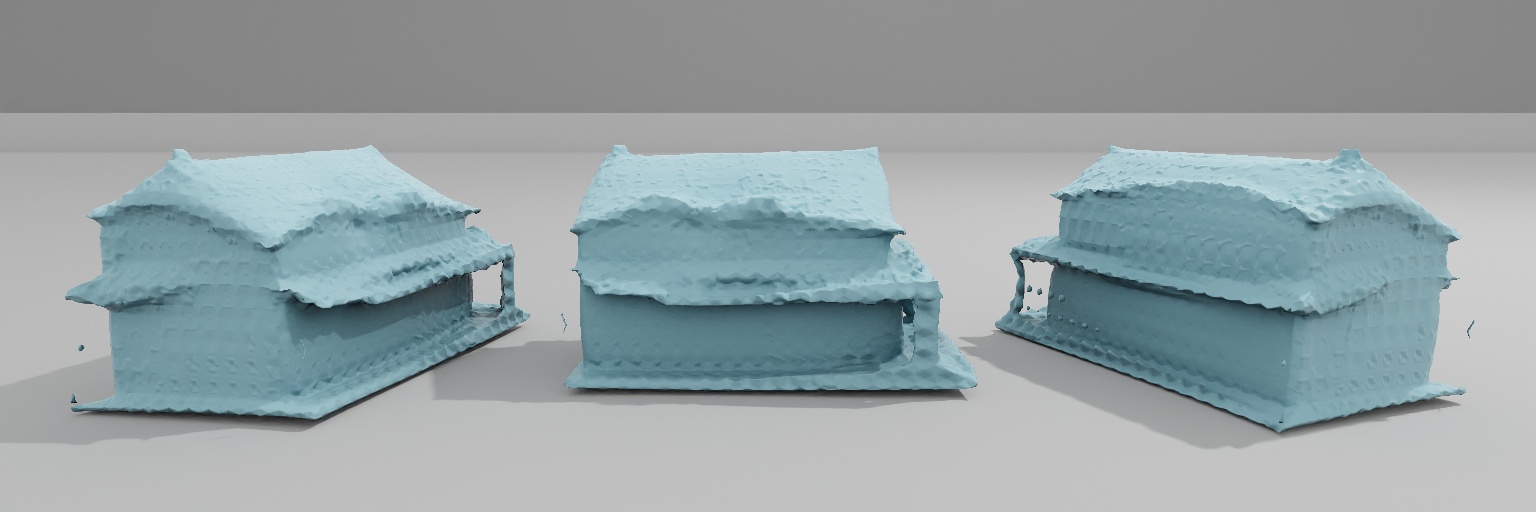}
\caption{\textbf{Shapes generated by {\ourmodel} rendered in Blender.} {\ourmodel} generates high-quality shapes with diverse texture, high-quality geometry, and complex topology. Zoom-in for details.}
\vspace{-1mm}
\label{fig:our_results}
\end{figure*}

\paragraph{Experimental Results} We provide quantitative results in Table.~\ref{tbl:evaluation_with_baselines} and qualitative examples in Fig.~\ref{fig:our_results_geo} and Fig.~\ref{fig:compare_gen_img}. Additional results are available in the supplementary video. Compared to OccNet~\cite{occnet} that uses 3D supervision during training, {\ourmodel} achieves better performance in terms of both diversity (COV) and quality (MMD), and our generated shapes have more geometric details. PointFlow~\cite{pointflow} outperforms {\ourmodel} in terms of MMD on CD, while {\ourmodel} is better in MMD on LFG. We hypothesize that this is because PointFlow directly optimizes on point locations, which favours CD. {\ourmodel} also performs favourably when compared to 3D-aware image synthesis methods, we achieve significant improvements over PiGAN~\cite{pigan} and GRAF~\cite{graf} in terms of all metrics on all datasets. Our generated shapes also contain more detailed geometry and texture. Compared with recent work EG3D~\cite{eg3d}. We achieve comparable performance on generating 2D images (FID-ori), while we significantly improve on 3D shape synthesis in terms of FID-3D, which demonstrates the effectiveness of our model on learning actual 3D geometry and texture.

Since we synthesize textured meshes, we can export our shapes into Blender\footnote{We use  xatlas~\cite{xatlas} to get texture coordinates for the extracted mesh, from where we can warp our 3D mesh into a 2D plane and obtain the corresponding 3D location on the  mesh surface for any position on the 2D plane. We then discretize the 2D plane into an image, and for each pixel, we query the texture field using corresponding 3D location to obtain the RGB color to get the texture map.}. We show rendering results in Fig.~\ref{fig:teaser_figure} and~\ref{fig:our_results}. {\ourmodel} is able to generate shapes with diverse and high quality geometry and topology, very thin structures (motorbikes), as well as complex textures on cars, animals, and houses.

\begin{figure*}[t!]
\centering
\includegraphics[width=\textwidth,trim=0 20 0 150,clip]{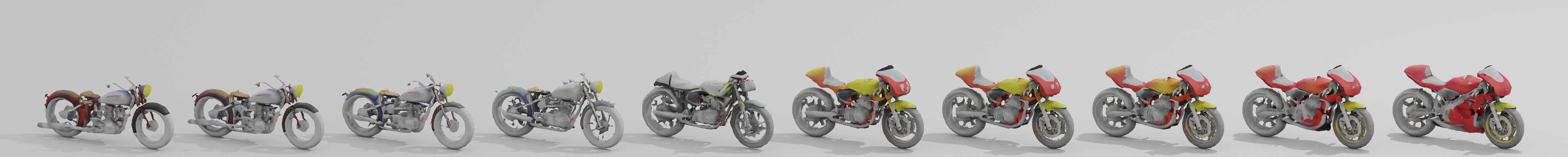}
\includegraphics[width=\textwidth,trim=0 100 0 40,clip]{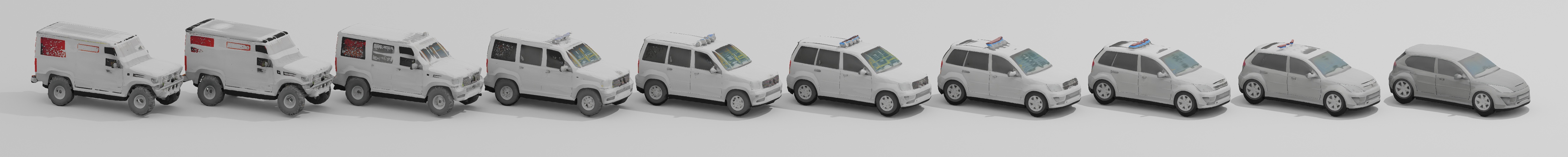}
\caption{ \textbf{Shape interpolation.} We interpolate both geometry and texture latent codes from left to right.}
\label{fig:interpolation}
\end{figure*}

\begin{figure*}[t!]
\centering
\includegraphics[width=\textwidth,trim=0 20 0 50,clip]{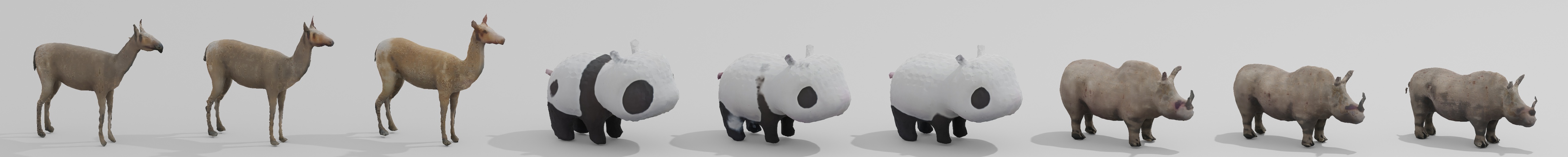}
\caption{\textbf{Shape variation.} We locally perturb each latent code to generate different shapes.}
\label{fig:local_variation}
\end{figure*}

\textbf{Shape Interpolation} {\ourmodel} also enables shape interpolation, which can be useful for editing purposes. We explore the latent space of {\ourmodel} in Fig.~\ref{fig:interpolation}, where we interpolate the latent codes to generate each shape from left to right. {\ourmodel} is able to faithfully generate a smooth and meaningful transition from one shape to another. We further explore the local latent space by slightly perturbing the latent codes to a random direction. {\ourmodel} produces novel and diverse shapes when applying local editing in the latent space (Fig.~\ref{fig:local_variation}). 

\subsection{Ablations}
\label{sec:exp_ablation}
We ablate our model in two ways: {\bf 1)} w/ and w/o volume subdivision, {\bf 2)} training using different image resolutions. Further ablations are provided in the Appendix~\ref{sec:additional_ablation}.

\begin{wraptable}[11]{r}{0.48\textwidth}
\vspace{-5mm}
\resizebox{0.9\linewidth}{!}{
\begin{tabular}{lcccccc}
\toprule
\multirow{2}{*}{Class} & \multirow{2}{*}{Img Res}   & \multicolumn{2}{c}{COV (\%, $\uparrow$)} & \multicolumn{2}{c}{MMD ($\downarrow$)} &  \multirow{2}{*}{FID ($\downarrow$)} \\\cmidrule(lr){3-4}\cmidrule(lr){5-6}
& & LFD & CD & LFD & CD &  \\ 
\midrule
  \multirow{3}{*}{Car} & 128$^2$  & 9.28 & 8.25 & 2224 & 1.30& 39.21  \\ 
 & 512$^2$ & 52.32 & 44.13 & 1593 & 0.80& 13.19\\ 
 & 1024$^2$ &\textbf{66.78} & \textbf{58.39} & \textbf{1491} & \textbf{0.71}& \textbf{10.25} \\ 
\midrule
   \multirow{3}{*}{Chair} & 128$^2$  &38.25 & 33.98 & 3886 & 5.90 &  43.04  \\ 
 & 512$^2$   & 68.80 & \textbf{69.92} & \textbf{3149} & 3.90 &  30.16  \\ 
 & 1024$^2$  & \textbf{69.08} & 67.87 & 3167 & \textbf{3.74}& \textbf{23.28} \\
\midrule
 \multirow{2}{*}{Mbike} & 512$^2$ &\textbf{68.49} & \textbf{65.75} & \textbf{3421} & 1.74 & 74.04\\ 
 & 1024$^2$ &67.12 & 64.38 & 3631 & \textbf{1.73} &\textbf{65.60}  \\ 
\midrule
\multirow{2}{*}{Animal}  & 512$^2$ &77.53 & \textbf{78.65} & 3828 & \textbf{2.01} &29.75 \\ 
 & 1024$^2$ &\textbf{79.78} & \textbf{78.65} & \textbf{3798} & 2.03& \textbf{28.33} \\ 
\bottomrule
\end{tabular}
\vspace{-5mm}
\caption{\footnotesize {\bf Ablating the image resolution.}  $\uparrow$:  higher is better, $\downarrow$: lower is better.
}
\label{tbl:ablation_volsub_imgres}
}
\end{wraptable}

\paragraph{Ablation of Volume Subdivision} 
As shown in Tbl.~\ref{tbl:evaluation_with_baselines}, volume subdivision significantly improves the performance on classes with thin structures (e.g., motorbikes), while not getting gains on other classes. We hypothesize that the initial tetrahedral resolution  is already sufficient to capture the detailed geometry on Chairs and Cars, and hence the subdivision cannot provide further improvements.

\paragraph{Ablating Different Image Resolutions}
We ablate the effect of the training image resolution in Tbl.~\ref{tbl:ablation_volsub_imgres}. As expected, increased image resolution improves the performance in terms of FID and shape quality, as the network can see more details, which are often not available in the low-resolution images. This corroborates the importance of training with higher image resolution, which are often hard to make use of for implicit-based methods.

\begin{figure*}[t!]
\centering
\includegraphics[width=\textwidth]{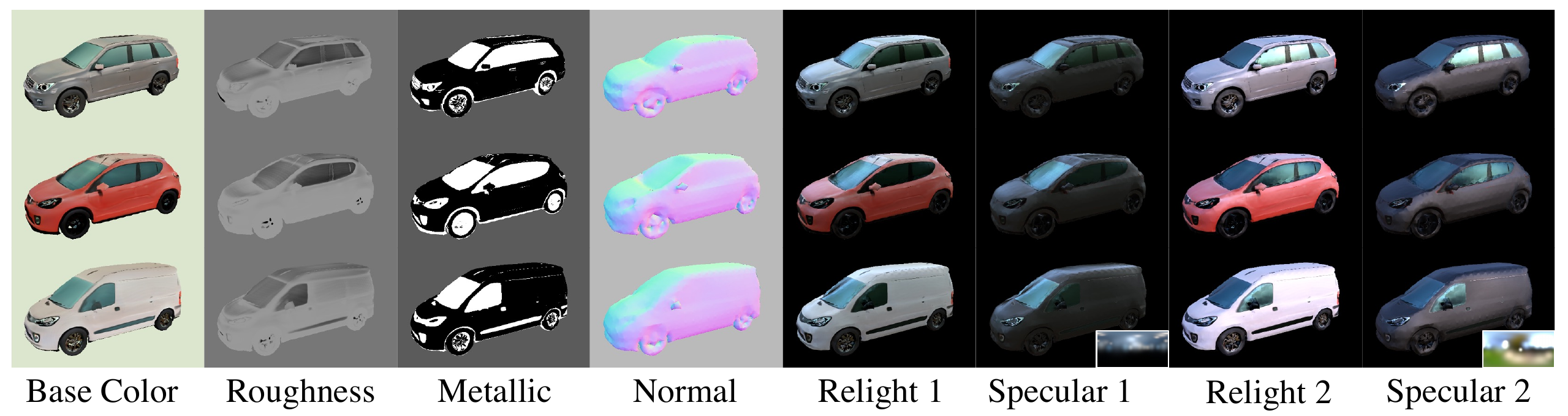}
\caption{\textbf{Material generation and relighting.} Despite being unsupervised, our model generates reasonable material properties, and can be realistically rendered with real-world HDR panoramas (bottom right). Normals are computed from the generated mesh. Note how specular effects change under two different lighting conditions. 
} 
\label{fig:material}
\end{figure*}

\subsection{Applications}
\label{sec:exp_application}
\subsubsection{Material Generation for View-dependent Lighting Effects}
\label{sec:material}

{\ourmodel} can easily be extended to also generate surface materials that are directly usable in modern graphics engines. In particular, we follow the widely used Disney BRDF \cite{burley2012physically,karis2013real} and describe the materials in terms of the base color ($\mathbb{R}^3$), metallic ($\mathbb{R}$), and roughness ($\mathbb{R}$) properties. As a result, we repurpouse our texture generator to now output a 5-channel reflectance field (instead of only RGB). To accommodate differentiable rendering of materials, we adopt an efficient spherical Gaussian (SG) based deferred rendering pipeline~\cite{chen2021dibrpp}. Specifically, we rasterize the reflectance field into a G-buffer, and randomly sample an HDR image from a set of real-world outdoor HDR panoramas $\mathcal{S}_\text{light} = \{L_{SG}\}_K$, where $L_{SG} \in \mathbb{R}^{32 \times 7}$ is obtained by fitting 32 SG lobes to each panorama. The SG renderer~\cite{chen2021dibrpp} then uses the camera $c$ to render an RGB image with view-dependent lighting effects, which we feed into the discriminator during training. Note that {\ourmodel} does not require material supervision during training and learns to generate decomposed materials in an unsupervised manner.

We provide qualitative results of generated surface materials in Fig.~\ref{fig:material}. Despite unsupervised, {\ourmodel} discovers interesting material decomposition, e.g., the windows are correctly predicted with a smaller roughness value to be more glossy than the car's body, and the car's body is discovered as more dielectric while the window is more metallic.
Generated materials enable us to produce realistic relighting results, which can account for complex specular effects under different lighting conditions.

\subsubsection{Text-Guided 3D Synthesis}
\label{sec:text_guided_synthesis}
\begin{figure*}[t!]
\centering
\includegraphics[width=\textwidth]{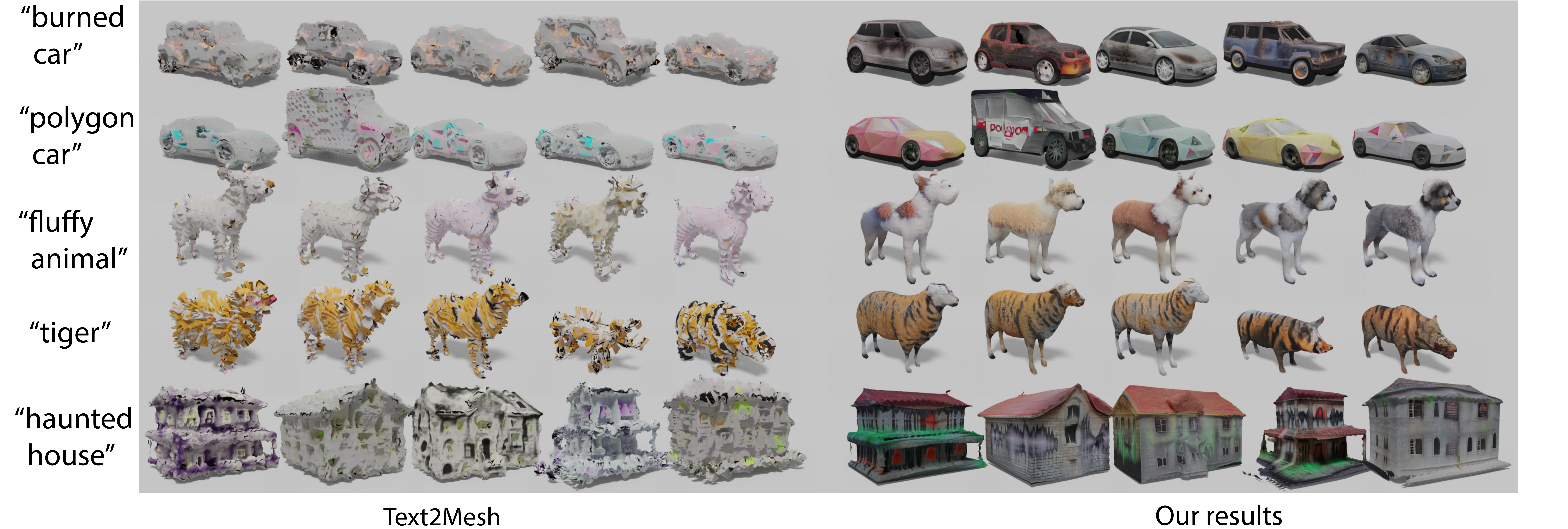}
\caption{\textbf{Text-guided 3D synthesis.} Note that Text2Mesh~\cite{Text2Mesh} requires 3D mesh geometry as input. To fulfil the requirement, we provide our generated geometry as its input mesh. }
\label{fig:text2mesh}
\end{figure*}

Similar to image GANs, {\ourmodel} also supports text-guided 3D content synthesis by fine-tuning a pre-trained model under the guidance of CLIP~\cite{CLIP}. Note that our final synthesis result is a textured 3D mesh. To this end, we follow the dual-generator design from styleGAN-NADA~\cite{stylegan-nada}, where a trainable copy $G_t$ and a frozen copy $G_f$ of the pre-trained generator are adopted. During optimization $G_t$ and $G_f$ both render images from 16 random camera views.  Given a text query, we sample 500 pairs of noise vectors $\mathbf{z}_1$ and $\mathbf{z}_2$. For each sample, we optimize the parameters of $G_t$ to minimize the directional CLIP loss~\cite{stylegan-nada} (the source text labels are ``car'', ``animal'' and ``house'' for the corresponding categories), and select the samples with minimal loss. To accelerate this process, we first run a small number of optimization steps for the 500 samples, then choose the top 50 samples with the lowest losses, and run the optimization for 300 steps. The results and comparison against a SOTA text-driven mesh stylization method, Text2Mesh~\cite{Text2Mesh}, are provided in Fig.~\ref{fig:text2mesh}. 
Note that, \cite{Text2Mesh} requires a mesh of the shape as an input to the method.  We provide our generated meshes from the frozen generator as input meshes to it. Since it needs mesh vertices to be dense to synthesize surface details with vertex displacements, we further subdivide the input meshes with  mid-point subdivision to make sure each mesh has 50k-150k vertices on average.

\vspace{-2mm}
\section{Conclusion}
\label{sec:conclusion}
\vspace{-2mm}
We introduced {\ourmodel}, a novel 3D generative model that is able to synthesize high-quality 3D textured meshes with arbitrary topology. {\ourmodel } is trained using only 2D images as supervision. We experimentally demonstrated significant improvements on generating 3D shapes over previous state-of-the-art methods on multiple categories.  We hope that this work brings us one step closer to democratizing 3D content creation using A.I.. 

\vspace{-2mm}
\paragraph{Limitations} While {\ourmodel} makes a significant step towards a practically useful 3D generative model of 3D textured shapes, it still has some limitations. In particular, we still rely on 2D silhouettes as well as the knowledge of camera distribution during training. As a consequence, {\ourmodel} was currently only evaluated on synthetic data. A promising extension could use the advances in instance segmentation and camera pose estimation to mitigate this issue and extend {\ourmodel} to real-world data. {\ourmodel} is also trained per-category; extending it to multiple categories in the future, could help us better represent the inter-category diversity.

\vspace{-2mm}
\paragraph{Broader Impact} 
We proposed a novel 3D generative model that generates 3D textured meshes, which can be readily imported into current graphics engines. Our model is able to generate shapes with arbitrary topology, high quality textures and rich geometric details, paving the path for democratizing A.I. tool for 3D content creation. As all machine learning models, {\ourmodel} is also prone to biases introduced in the training data. Therefore, an abundance of caution should be applied when dealing with sensitive applications, such as generating 3D human bodies, as {\ourmodel} is not tailored for these applications. We do not recommend using {\ourmodel} if privacy or erroneous recognition could lead to potential misuse or any other harmful applications. Instead, we do encourage practitioners to carefully inspect and de-bias the datasets before training our model to depict a fair and wide distribution of possible skin tones, races or gender identities. 

\section{Disclosure of Funding}
This work was funded by NVIDIA. Jun Gao, Tianchang Shen, Zian Wang and Wenzheng Chen acknowledge additional revenue in the form of student scholarships from University of Toronto and the Vector Institute, which are not in direct support of this work. 
{\small
\bibliographystyle{plain}
\bibliography{egbib}
}
\clearpage
\setcounter{section}{0}
\setcounter{figure}{0}
\setcounter{table}{0}
\renewcommand{\thesection}{\Alph{section}}
\renewcommand\thefigure{\Alph{figure}}   
\renewcommand\thetable{\Alph{table}} 

\newcommand{\x}{\mathbf{x}}
\newcommand{\dir}{\bm{\omega}}
\newcommand{\wi}{\dir_\mathrm{i}}
\newcommand{\wo}{\dir_\mathrm{o}}
\newcommand{\wh}{\dir_\mathrm{h}}
\newcommand{\Li}{L_\mathrm{i}}
\newcommand{\Le}{L_\mathrm{e}}
\newcommand{\Lo}{L_\mathrm{o}}
\newcommand{\normal}{\mathbf{n}}
\newcommand{\thetai}{\theta_\mathrm{i}}
\newcommand{\dd}{\mathrm{d}}
\newcommand{\brdf}{f_\mathrm{r}}
\newcommand{\hemis}{{\mathcal{H}^2}}
\newcommand{\sphere}{{\mathcal{S}^2}}
\newcommand{\object}{\mathcal{M}}
\newcommand{\basecolor}{\mathbf{c}_\text{base}}
\newcommand{\albedo}{a}
\newcommand{\diffAlbedo}{\mathbf{c}_\text{diff}}
\newcommand{\specAlbedo}{\mathbf{c}_\text{spec}}
\newcommand{\rough}{\beta}
\newcommand{\metal}{m}
\newcommand{\amplitude}{\bm{\mu}}
\newcommand{\sharpness}{\lambda}
\newcommand{\axis}{\bm{\xi}}
\newcommand{\sg}{\mathcal{G}}
\newcommand{\real}{\mathbb{R}}
\newcommand{\ndf}{D}
\newcommand{\fresnel}{F}
\newcommand{\geom}{G}
\newcommand{\microfacet}{M}
\newcommand{\intersecMap}{X}
\newcommand{\texturemap}{T}
\newcommand{\albedomap}{A}
\newcommand{\normalmap}{N}
\newcommand{\vismap}{V}
\newcommand{\woMap}{W_\mathrm{o}}
\newcommand{\unit}{[0,1]}
\newcommand{\envmap}{\Li}
\newcommand{\rasterizer}{R}%
\newcommand{\shader}{S}%
\newcommand{\objectParam}{\bm{\Theta}}
\newcommand{\materialParam}{\bm{\theta}}
\newcommand{\shapeParam}{\bm{\pi}}
\newcommand{\lightParam}{\bm{\gamma}}
\newcommand{\pix}{p}
\newcommand{\pixel}{p}
\newcommand{\vis}{v}
\newcommand{\surfFeatures}{\mathbf{f}}
\newcommand{\camera}{\mathbf{c}}
\newcommand{\image}{I}
\newcommand{\network}{F}%
\newcommand{\networkParam}{\bm{\vartheta}}

\newcommand{\light}{L}
\newcommand{\loss}{\mathcal{L}}
\newcommand{\rt}{Monte Carlo}
\newcommand{\rtshort}{MC}
\newcommand{\sgshort}{SG}

\newcommand{\etal}{et al. }

\section*{\LARGE Appendix}
In this Appendix, we first provide detailed description of the {\ourmodel} network architecture (Sec.~\ref{sec:mapping_network}-~\ref{sec:rendering_discriminator}) along with the training procedure and hyperparameters (Sec.~\ref{sec:training_hyperparameters}). We then describe the datasets (Sec.~\ref{sec:datasets}), baselines (Sec.~\ref{sec:baselines}), and evaluation metrics (Sec.~\ref{sec:evaluation_metrics}). Additional qualitative results, ablation studies, robustness analysis, and results on the real dataset are available in Sec.~\ref{sec:more_exp_results}. Details and additional results of the material generation for view-dependent lighting effects are provided in Sec.~\ref{sec:exp_sg}. Sec~\ref{sec:exp_text2mesh} contains more information about the text-guided shape generation experiments as well as more additional qualitative results. The readers are also kindly referred to the accompanying video (\textit{demo.mp4}) that includes 360-degree renderings of our results (more than 400 generated shapes for each category), detailed zoom-ins, interpolations, material generation, and shapes generated with text-guidance. 

\section{Details of Our Model}
\label{sec:model_details}
In Sec.~\ref{sec:method} we have provided a high level description of {\ourmodel}. Here, we provide the implementation details that were omitted due to the lack of space. Please consult the Figure~\ref{fig:network_architecture} and Figure~2 in the main paper for more context. Source code is available at our \href{https://nv-tlabs.github.io/GET3D/}{project webpage} 

\subsection{Mapping Network}
\label{sec:mapping_network}
Following StyleGAN~\cite{stylegan,stylegan2}, our mapping networks $f_\text{geo}$ and $f_\text{tex}$ are 8-layer MLPs in which each fully-connected layer has 512 hidden dimensions and a leaky-ReLU activation (Figure~\ref{fig:network_architecture}). The mapping networks are used to map the randomly sampled noise vectors $\mathbf{z}_1 \in \mathbb{R}^{512}$ and $\mathbf{z}_2 \in \mathbb{R}^{512}$ to the latent vectors $\mathbf{w}_1 \in \mathbb{R}^{512}$ and $\mathbf{w}_2 \in \mathbb{R}^{512}$ as $\mathbf{w}_1 = f_\text{geo}(\mathbf{z}_1)$ and $\mathbf{w}_2 = f_\text{tex}(\mathbf{z}_2)$.

\subsection{Geometry Generator}
\label{sec:geometry_gen}

The geometry generator of {\ourmodel} starts from a randomly initialized feature volume $\mathbf{F}_\text{geo} \in \mathbb{R}^{4\times4\times4\times256}$ that is shared across the generated shapes, and is learned during training. Through a series of four modulated 3D convolution blocks (\emph{ModBlock3D} in Figure~\ref{fig:network_architecture}), the initial volume is up-sampled to a feature volume $\mathbf{F}^{\prime}_\text{geo} \in \mathbb{R}^{32\times32\times32\times64}$ that is conditioned on $\mathbf{w}_1$. Specifically, in each \emph{ModBlock3D}, the input feature volume is first upsampled by a factor of two using trilinear interpolation. It is then passed through a small 3D ResNet, where the residual path uses a 3D convolutional layer with kernel size 1x1x1, and the main path applies two \emph{conditional} 3D convolutional layers with kernel size 3x3x3. To perform the conditioning, we follow StyleGAN2~\cite{stylegan2} and first map the latent vector $\mathbf{w}_1$ to \emph{style} $\mathbf{h}$ through a learned affine transformation (\textbf{A} in Figure~\ref{fig:network_architecture}). The \textit{style} $\mathbf{h}$ is then used to modulate (\textbf{M}) and demodulate (\textbf{D}) the weights of the convolutional layers as:

\begin{eqnarray}
    \textbf{M}: \theta^{\prime}_{i, j, k, l, m} &=& h_{i} \cdot   \omega_{i, j, k, l, m}, \\ 
    \textbf{D}: \theta^{\prime\prime}_{i, j, k, l, m} &=& \theta^{\prime}_{i, j, k, l, m} / \sqrt{\sum_{i, k, l, m} {\theta^{\prime}_{i, j, k, l, m} }^2},
    \label{eq:modulate}
\end{eqnarray}
where $\theta$ and $\theta^{\prime\prime}$ are the original and modulated weight, respectively. $h_i$ is the \textit{style} corresponding to the $i$th input channel, $j$ is the output channel dimension, and $k, l, m$ denote the spatial dimension of the 3D convolutional filter. 

Once we obtain the final feature volume $\mathbf{F}^{\prime}_\text{geo}$, the feature vector $\mathbf{f}^{\prime}_\text{geo} \in \mathbb{R}^{64}$ of each vertex $\mathbf{v}$ in the tetrahedral grid can be obtained through trilinear interpolation. We additionally feed the coordinates of the point $\mathbf{p}$ to a $[\sin(\mathbf{p}), \cos(\mathbf{p})]$ positional encoding (PE) and concatenate the output with the feature vector $\mathbf{f}^{\prime}_\text{geo}$. To decode the concatenated feature vector into the vertex offset $\Delta\mathbf{v}\in \mathbb{R}^3$ or the SDF value $s \in \mathbb{R}$, we pass it through three conditional FC layers (\emph{ModFC} in Figure~\ref{fig:network_architecture}). The modulation and demodulation in these layers is done analogously to Eq.~\ref{eq:modulate}. All the layers, except for the last, are followed by the leaky-ReLU activation function. In the last layer, we  apply \texttt{tanh} to either normalize the SDF prediction $s$ to be within [-1, 1], or normalize the $\Delta\mathbf{v}$ to be within [-$\frac{1}{\text{tet-res}}$, $\frac{1}{\text{tet-res}}$], where \text{tet-res} denotes the resolution of our tetrahedral grid, which we set to 90 in all the experiments. 

Note that for simplicity, we remove all the noise vector from StyleGAN~\cite{stylegan,stylegan2} and only have stochasticity in the input $\textbf{z}$. Furthermore, following practices from DEFTET~\cite{gao2020deftet} and DMTET~\cite{dmtet}, we us two copies of the geometry generator. One generates the vertex offsets $\Delta\mathbf{v}$, while the other outputs the SDF values $s$. The architecture of both is the same, except for the output dimension and activation function of the last layer.

\begin{wrapfigure}[12]{R}{0.315\textwidth}
\centering
\vspace{-4mm}
\includegraphics[width=0.98\textwidth,clip]{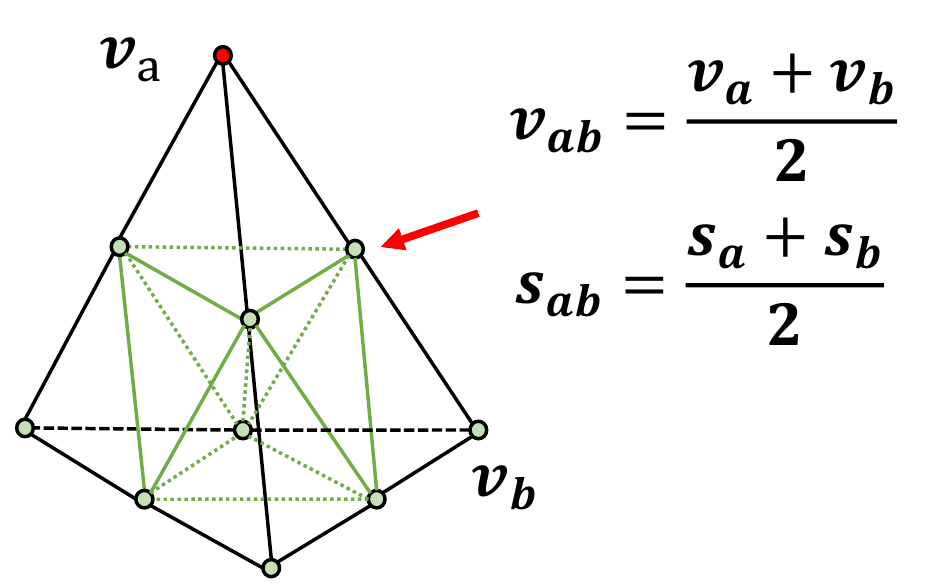}
\vspace{-2mm}
\caption{\footnotesize With volume subdivision, each tetrahedron is divided into 8 smaller tetrahedra by connecting midpoints. }
\vspace{-1mm}
\label{fig:vol_subdiv}
\end{wrapfigure}
\paragraph{Volume Subdivision:} In cases where modeling at a high-resolution is required (e.g. motorbike with thin structures in the wheels), we further use volume subdivision following DMTET~\cite{dmtet}. As illustrated in Fig.~\ref{fig:vol_subdiv}, we first subdivide the tetrahedral grid and compute SDF values of the new vertices (midpoints) by averaging the SDF values on the edge. Then we identify tetrahedra that have vertices with different SDF signs. These are the tetrahedra that intersect with the underlying surface encode by SDF. To refine the surface at increased grid resolution after subdivision, we further predict the residual on SDF values and deformations to update $s$ and $\Delta \mathbf{v}$ of the vertices in \textbf{identified} tetrahedra. Specifically, we use an additional 3D convolutional layer to upsample feature volume $ \mathbf{F}^{\prime}_\text{geo}$ to $ \mathbf{F}^{\prime\prime }_\text{geo}$ of shape $64\times64\times64\times8$ conditioned on $w_1$. Then, following the steps described above, we use trilinear interpolation to obtain per-vertex feature, concatenate it with PE and decode the residuals $\delta s$  and $\delta \mathbf{v}$  using conditional FC layers. The final SDF and vertex offset are computed as:
\begin{eqnarray}
s^{\prime} = s+\delta s, \
\Delta\mathbf{v}^{\prime} = \Delta\mathbf{v} + \delta \mathbf{v}.
\end{eqnarray}

\begin{figure*}[t!]
\centering
\includegraphics[width=\textwidth]{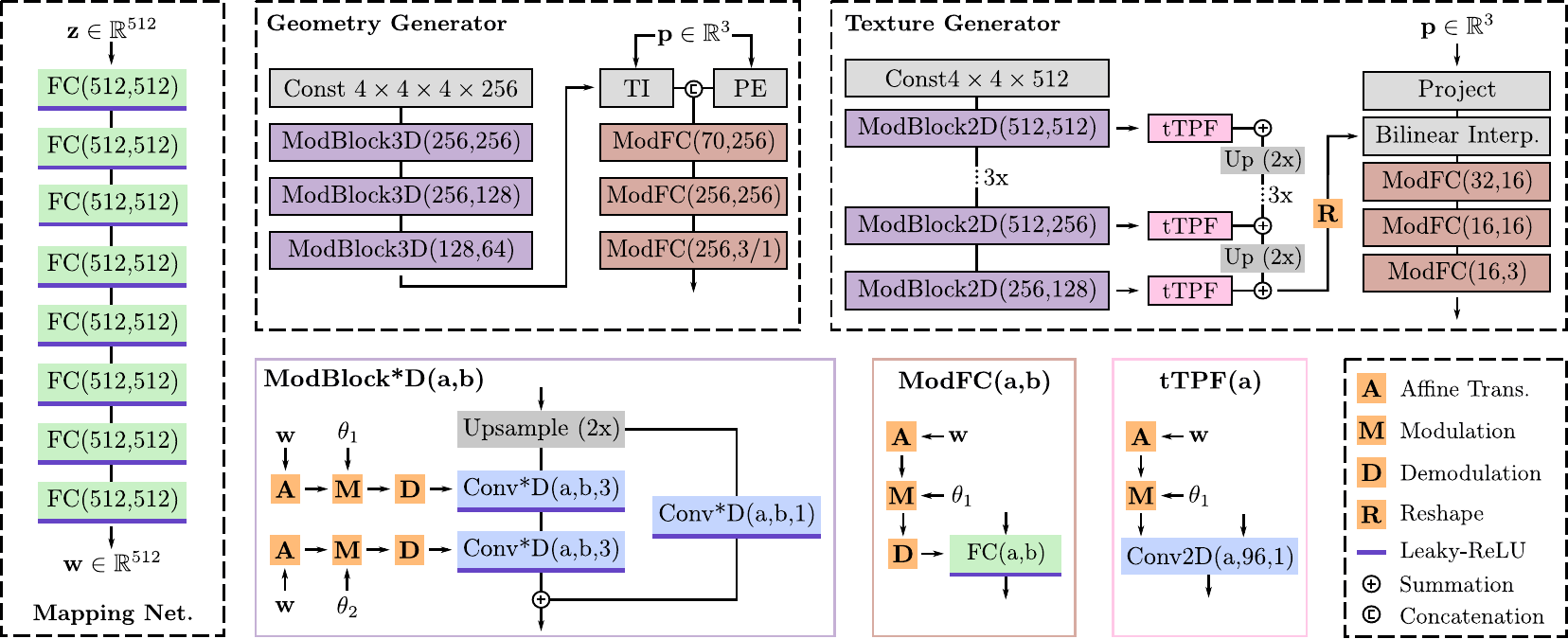}
\caption{\textbf{Network architecture of {\ourmodel}}. TI and PE denote trilinear interpolation and positional encoding, respectively. FC($a,b$) represents a fully connected layer with $a$ and $b$ denoting the input and output dimension, respectively. Similarly, Conv3D($a,b,c$) denotes a 3D convolutional layer with $a$ input channels, $b$ output channels, and kernel dimension $c\times c \times c$. In the Texture Generator, the block ModBlock2D(512,512) is repeated four times. All convolutional layers have stride 1.} 
\label{fig:network_architecture}
\end{figure*}

\subsection{Texture Generator}
\label{sec:texture_generator_suppl}
We adapt the generator architecture from StyleGAN2~\cite{stylegan2} to generate a tri-plane representation of the texture field. Similar as in the geometry generator, we start from a randomly initialized feature grid $\mathbf{F}_\text{tex} \in \mathbb{R}^{4\times4\times512}$ that is shared across the shapes, and is learned during training. This initial feature grid is up-sampled to a feature grid $\mathbf{F}^\prime_\text{tex} \in \mathbb{R}^{256\times256\times96}$ that is conditioned on $\mathbf{w}_1$ and $\mathbf{w}_2$. Specifically, we use a series of six modulated 2D convolution blocks (\emph{ModBlock2D} in Figure~\ref{fig:network_architecture}). The \emph{ModBlock2D} blocks are the same as the \emph{ModBlock3D} blocks, except that the convolution is 2D and that the conditioning is on $\mathbf{w}_1\oplus\mathbf{w}_2$, where $\oplus$ denotes concatenation. Additionally, the output of each \emph{ModBlock2D} block is passed through a conditional \emph{tTPF} layer that applies a conditional 2D convolution with kernel size 1x1. Note that, following the practices from StyleGAN2~\cite{stylegan2}, the conditioning in the \emph{tTPF} layers is performed only through \emph{modulation} of the weights (no \emph{demodulation}).

The output of the last \emph{tTPF} layer is then reshaped into three axis-aligned feature planes of size $256\times256\times32$.

To obtain the feature $\mathbf{f}_\text{tex} \in \mathbb{R}^{32}$ of a surface point $\mathbf{p}  \in \mathbb{R}^{3}$, we first project $\mathbf{p}$ onto each plane, perform bilinear interpolation of the features, and finally sum the interpolated features:
\begin{equation}
\mathbf{f}_\text{tex} = \sum_e \rho(\pi_e(\mathbf{p})),
\label{eq:triplane_query}
\end{equation}
where $\pi_e(\mathbf{p})$ is the projection of the point $\mathbf{p}$ to the feature plane $e$ and $\rho(\cdot)$ denotes bilinear interpolation of the features. Color $\mathbf{c} \in \mathbb{R}^3$ of the point $\mathbf{p}$ is then decoded from $\mathbf{f}^t $ using three conditional FC layers (\emph{ModFC}) conditioned on $\mathbf{w}_1\oplus\mathbf{w}_2$. The hidden dimension of each layer is 16.  Following StyleGAN2~\cite{stylegan2}, we do not apply normalization to the final output. 

\subsection{2D Discriminator}
\label{sec:rendering_discriminator}
We use two discriminators to train {\ourmodel}: one for the RGB output and one for the 2D silhouettes. For both, we use exactly the same architecture as the discriminator in StyleGAN~\cite{stylegan}. Empirically, we have observed that conditioning the discriminator on the camera pose leads to canonicalization of the shape orientations. However, discarding this conditioning only slightly affects the performance, as shown in Section~\ref{sec:additional_ablation}. In fact, we primarily use this conditioning to enable the evaluation of geometry using evaluation metrics, which assume that the shapes are generated in the canonical frame.

\begin{figure*}[t!]
\centering
\includegraphics[width=\textwidth]{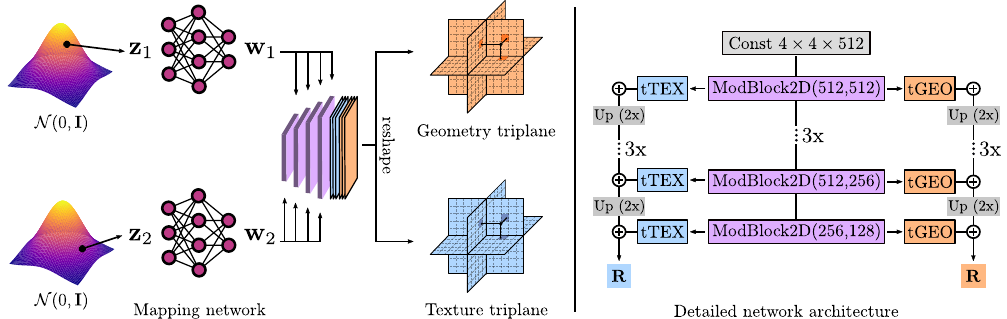}
\caption{\textbf{Improved generator architecture of {\ourmodel}}. High-level overview \emph{left} and detailed architecture \emph{right}. Different to the model architecture proposed in the main paper, the new generator shares the same backbone network for both geometry and texture generation. This improves the information flow and enables better disentanglement of the geometry and texture.}
\label{fig:network_architecture_improved}
\end{figure*}

\subsection{Improved Generator}
\label{sec:improved_generator}
\begin{figure*}[t!]
\centering
\includegraphics[width=\textwidth]{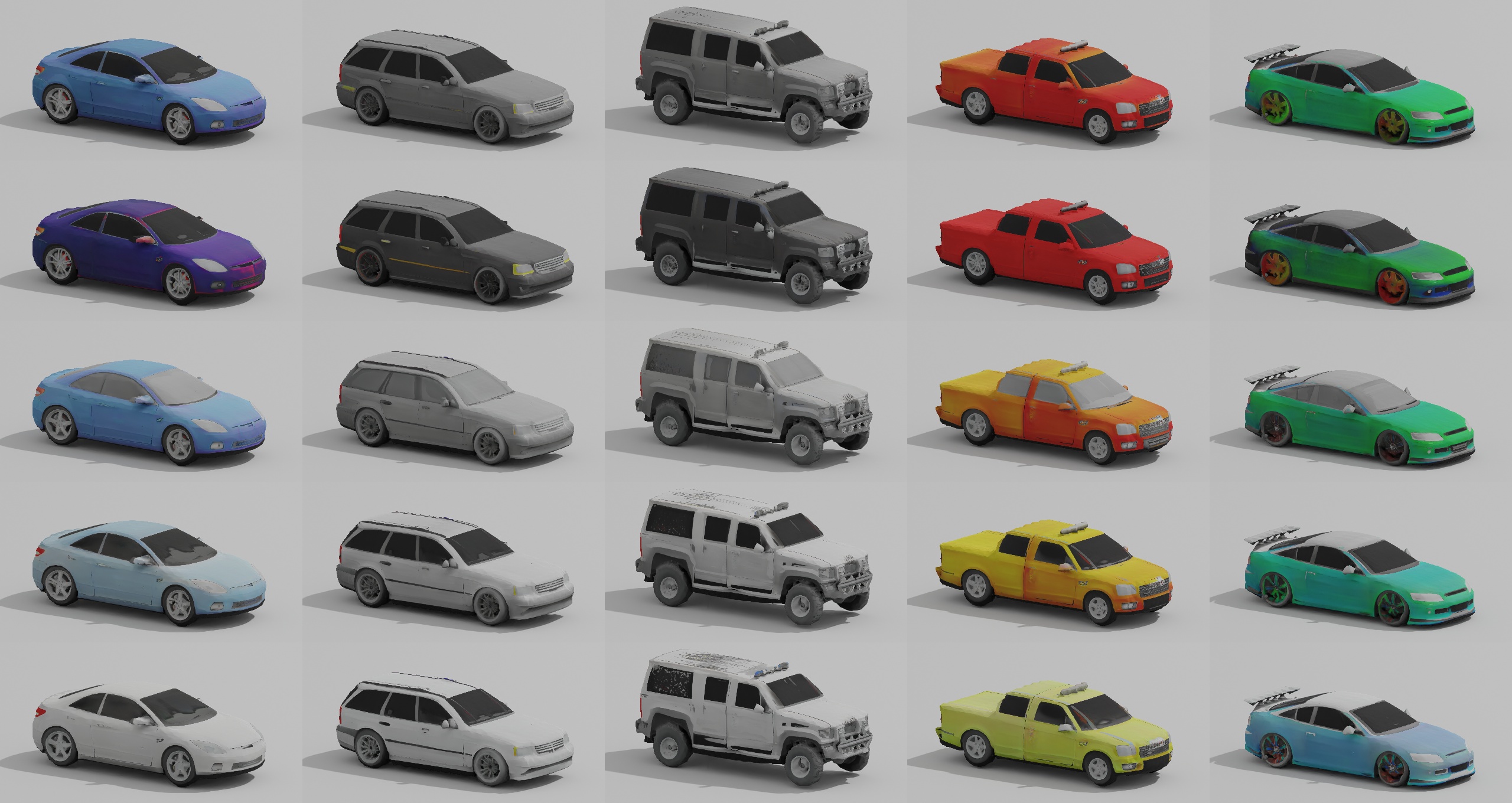}
\caption{\textbf{Disentanglement of geometry and texture achieved by the original model depicted in Fig.~\ref{fig:pipeline}}. In each row, we show shapes generated from the same texture latent code, while changing the geometry latent code. In each column, we show shapes generated from the same geometry latent code, while changing the texture code. The original model fails to achieve good disentanglement.} 
\label{fig:old_model_swap}
\end{figure*}

The motivation for sampling two noise vectors ($\mathbf{z}_1$, $\mathbf{z}_2$) in the generator is to enable disentanglement of the geometry and texture, where geometry is to be treated as a first-class citizen. Indeed, the geometry should only be controlled by the geometry latent code, while the texture should be able to not only adapt to the changes in the texture latent code, but also to the changes in geometry, i.e. a change in the geometry latent should propagate to the texture. However, in the original design of the {\ourmodel} generator (c.f. Sec.~\ref{sec:method} and Fig.~\ref{fig:pipeline}) the information flow from the geometry to the texture generator is very limited---concatenation of the two latent codes (Fig.~\ref{fig:network_architecture}). Such a weak connection makes it hard to learn the disentanglement of geometry and texture and the texture generator can even learn to ignore the texture latent code (Fig.~\ref{fig:old_model_swap}.). 

This empirical observation motivated us to improve the design of the generator network, after the initial submission, by improving the information flow, which in turn better supports the disentanglement of the geometry and texture. To this end, our improved generator shares the same backbone network for both geometry and texture generation, as shown in Fig.~\ref{fig:network_architecture_improved}. In particular, we follow SemanticGAN~\cite{semanticGAN} and use StyleGAN2~\cite{stylegan2} backbone. Each ModBlock2D (modulated with the geometry latent code $\mathbf{w}_1$), now has two tTPF branches, one for generating the geometry feature (\emph{tGEO}), and the other for generating texture features (\emph{tTEX}). The output of this backbone network are two feature triplanes, one for geometry and one for texture. To predict the SDF value and deformation for each vertex in the tetrahedral grid, we project the vertex onto each of the geometry triplanes, obtain its feature vector using Eq.~\ref{eq:triplane_query}, and finally use a ModFC to decode $s_i$ and $\Delta \mathbf{v}_i$. The prediction of the color in the texture branch remains unchanged.

\begin{figure*}[t!]
\centering
\includegraphics[width=\textwidth]{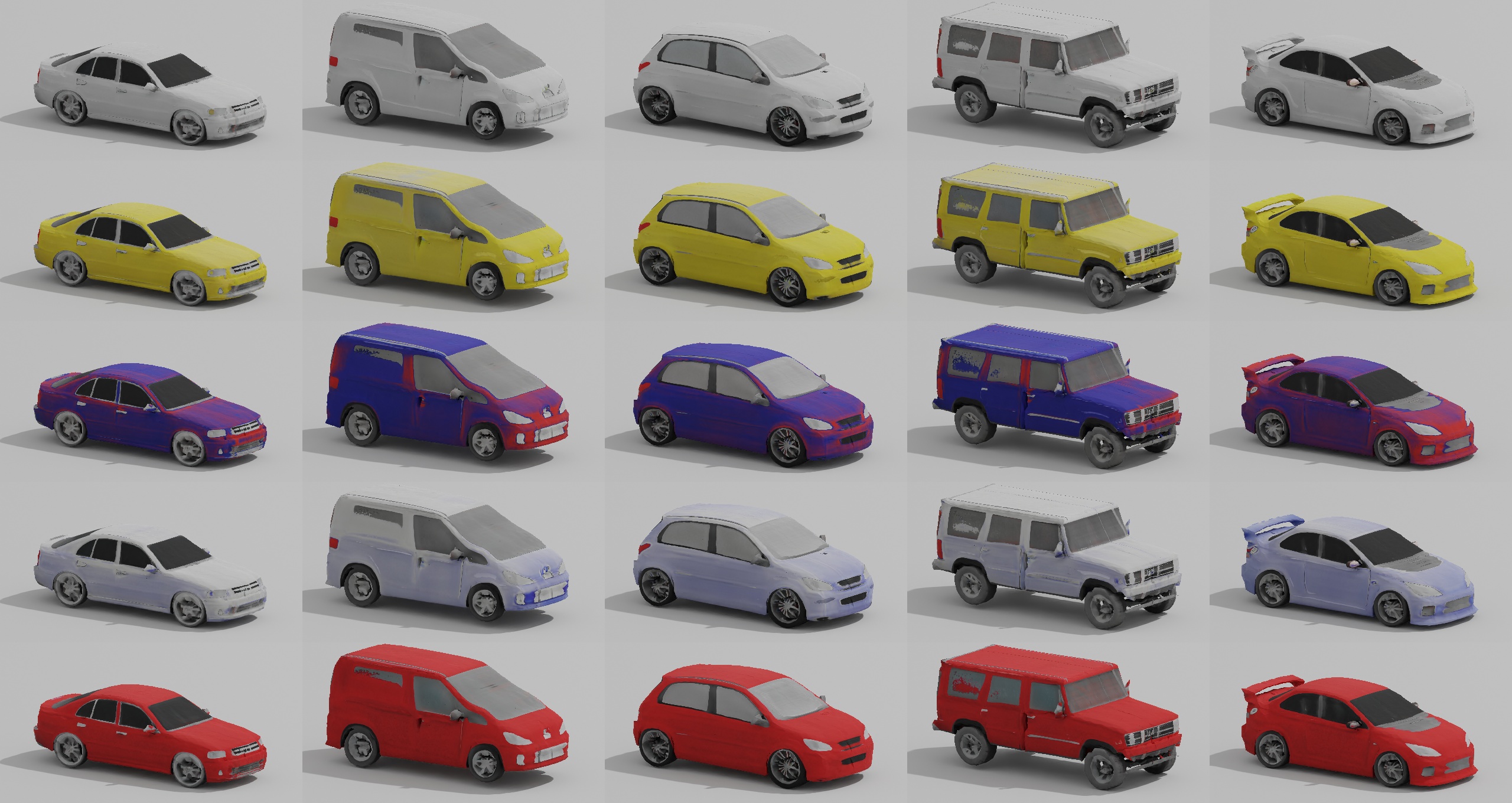}
\caption{\textbf{Disentanglement of geometry and texture achieved by the improved model depicted in Fig.~\ref{fig:network_architecture_improved}}. In each row, we show shapes generated from the same texture latent code, while changing the geometry latent code. In each column, we show shapes generated from the same geometry latent code, while changing the texture code. The disentanglement in this model is poor. Comparing with Fig.~\ref{fig:old_model_swap}, this improved model achieves significant better disentanglement of geometry and texture.} 
\label{fig:new_model_swap}
\end{figure*}

\begin{figure*}[t!]
\centering
\includegraphics[width=\textwidth,trim=0 0 0 0,clip]{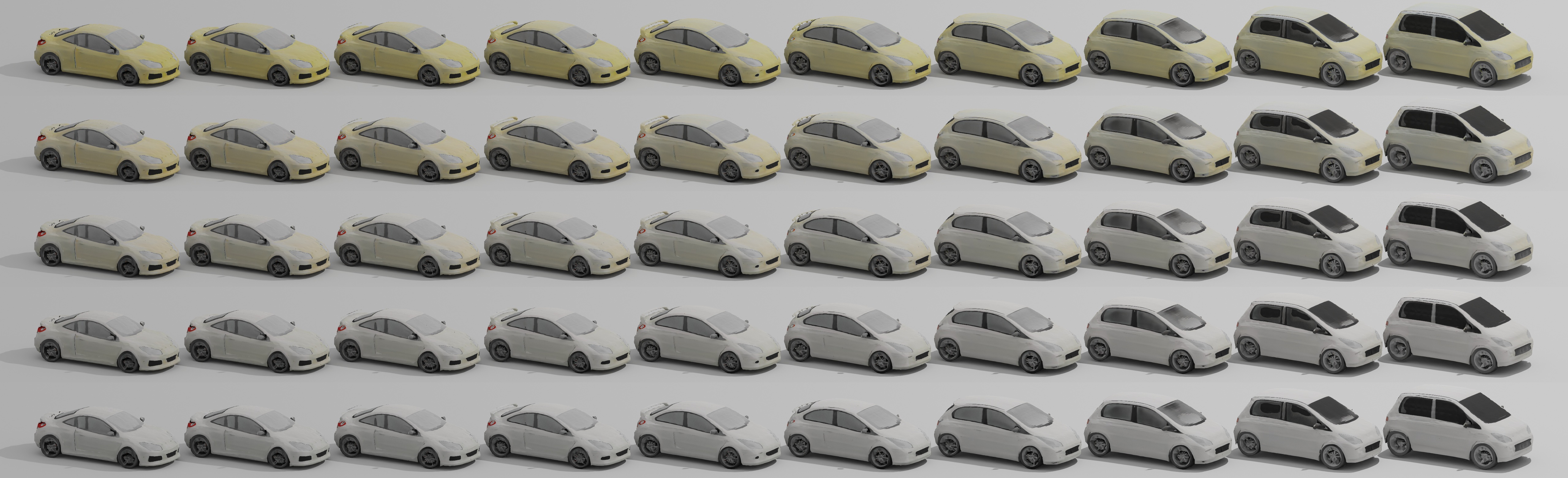}
\caption{\textbf{Shape Interpolation.} We interpolate the latent code from top-left corner to the bottom-right corner. In each row, we keep the texture latent code fixed and interpolate the geometry latent code. In each column, we keep the geometry latent code fixed and interpolate the texture latent code. {\ourmodel} adequately disentangles geometry and texture, while also providing a meaningful interpolation for both geometry or texture.}
\label{fig:qualitative_results_inter_1}
\end{figure*}

Qualitative result of the geometry and texture disentanglement achieved with this improved generator is depicted in Fig.~\ref{fig:new_model_swap} and~\ref{fig:qualitative_results_inter_1}. Shared backbone network allows us to achieve much better disentanglement of geometry and texture (Fig.~\ref{fig:old_model_swap} vs Fig.~\ref{fig:new_model_swap}), while also achieving better quantitative metrics on the task of unconditional generation (Tab.~\ref{tbl:evaluation_with_baselines}).

\subsection{Training Procedure and Hyperparameters}
\label{sec:training_hyperparameters}
We implement {\ourmodel} on top of the official PyTorch implementation of StyleGAN2~\cite{stylegan2}\footnote{StyleGan3: \url{https://github.com/NVlabs/stylegan3} (NVIDIA Source Code License)}. Our training configuration largely follows StyleGAN2~\cite{stylegan2} including: using a minibatch standard deviation in the discriminator, exponential moving average for the generator, non-saturating logistic loss, and R1 Regularization. 
We train {\ourmodel} along with the 2D discriminators from scratch, without progressive training or initialization from pretrained checkpoints. Most of our hyper-parameters are adopted form styleGAN2~\cite{stylegan2}. Specifically, we use Adam optimizer with learning rate 0.002 and  $\beta = 0.9$. For R1 regularization, we set the regularization weight $\gamma$ to 3200 for chair, 80 for car, 40 for animal, 80 for motorbike,  80 for renderpeople, and 200 for house.  We follow StyleGAN2~\cite{stylegan2} and use lazy regularization, which applies R1 regularization to discriminators only every 16 training steps.  Finally, we set the hyperparameter $\mu$ that controls the SDF regularization to 0.01 in all the experiments. We train our model using a batch size of 32 on 8 A100 GPUs for all the experiments. Training a single model takes about 2 days to converge. 

\section{Experimental Details}
\label{sec:exp_details}

\subsection{Datasets}
\label{sec:datasets}
We evaluate {\ourmodel} on ShapeNet~\cite{shapenet}, TurboSquid~\cite{Turbosquid}, and RenderPeople~\cite{renderpeople} datasets. In the following, we provide their detailed description and the preprocessing steps that were used in our evaluation. Detailed statistic of the datasets is available in Table~\ref{tbl:dataset}. 

\textbf{ShapeNet}\footnote{The ShapeNet license is explained at \url{https://shapenet.org/terms}}~\cite{shapenet} contains more than 51k shapes from 55 different categories and is the most commonly used dataset for benchmarking 3D generative models\footnote{Herein, we used ShapeNet v1 Core subset obtained from \url{https://shapenet.org/}}. Prior work~\cite{pointflow, zhou2021pvd} typically uses the categories \emph{Airplane}, \emph{Car}, and \emph{Chair} for evaluation. Herein, we replace the category \emph{Airplane} with \emph{Motorcycle}, which has more complex geometry and contains shapes with varying genus. \emph{Car}, \emph{Chair}, and \emph{Motorcycle} contain 7497, 6778, and 337 shapes, respectively. We random split the shapes of each category into training~(70\%), validation~(10\%), and test~(20\%) and remove from the test set shapes that have duplicates in the training set.

\textbf{TurboSquid}\footnote{\url{https://www.turbosquid.com}, we obtain consent via an agreement with TurboSquid, and following license at \url{https://blog.turbosquid.com/turbosquid-3d-model-license/}}~\cite{Turbosquid} is a large collection of various 3D shapes with high-quality geometry and texture, and is thus well suited to evaluate the capacity of {\ourmodel} to generate shapes with high-quality details. To this end, we use the category \emph{Animal} that contains 442 textured shapes with high diversity ranging from cats, dogs, and lions, to bears and deer~\cite{dmtet,yin2021_3DStyleNet}. We again randomly split the shapes into training~(70\%), validation~(10\%), and test~(20\%) set. Additionally, we provide qualitative results on the category \emph{House} that contains 563 shapes. Since we perform only qualitative evaluation on \emph{House}, we use all the shapes for training.

\textbf{RenderPeople}\footnote{We follow the license of Renderpeople \url{https://renderpeople.com/general-terms-and-conditions/}}~\cite{renderpeople} is a large dataset containing photorealistic 3D models of real-world \emph{humans}. We use it to showcase the capacity of {\ourmodel} to generate high-quality and diverse characters that can be used to populate virtual environments, such as games or even movies. In particular, we use 500 models from the whole dataset for training and only perform qualitative analysis. 

\paragraph{Preprocessing} To generate the data, we first scale each shape such that the longest edge of its bounding-box equals $\mathbf{e}_m$, where $\mathbf{e}_m = 0.9$ for \emph{Car}, \emph{Motorcycle}, and \emph{Human}, $\mathbf{e}_m = 0.8$ for \emph{House}, and $\mathbf{e}_m = 0.7$ for \emph{Chair} and \emph{Animal}.
For methods that use 2D supervision (Pi-GAN, GRAF, EG3D, and our model {\ourmodel}),
we then render the RGB images and silhouettes from camera poses sampled from the upper hemisphere of each object. Specifically, we sample 24 camera poses for \emph{Car} and \emph{Chair}, and 100 poses for \emph{Motorcycle}, \emph{Animal}, \emph{House}, and \emph{Human}. The rotation and elevation angles of the camera poses are sampled uniformly from a specified range (see Table~\ref{tbl:dataset}). For all camera poses, we use a fixed radius of 1.2 and the fov angle of $49.13^{\circ}$. We render the images in Blender~\cite{blender} using a fixed lighting, unless specified differently. 

For the methods that rely on 3D supervision, we follow their preprocessing pipelines~\cite{pointflow,occnet}. Specifically, for Pointflow~\cite{pointflow} we randomly sample 15k points from the surface of each shape, while for OccNet~\cite{occnet} we convert the shapes into watertight meshes by rendering depth frames from random camera poses and performing TSDF fusion.

\begin{table}[!t]
\centering
\def\arraystretch{1.25}
\begin{tabular}{lcccc}
\toprule
Dataset & \# Shapes & \# Views per shape & Rotation Angle & Elevation Angle \\
\midrule
ShapeNet Car &7497 & 24 & [0, $2\pi$] & [$\frac{1}{3}\pi$, $\frac{1}{2}\pi$]\\
ShapeNet Chair &6778 & 24 & [0, $2\pi$]&[$\frac{1}{3}\pi$, $\frac{1}{2}\pi$]\\
ShapeNet Motorbike & 337 & 100 & [0, $2\pi$] &[$\frac{1}{3}\pi$, $\frac{1}{2}\pi$]\\
Turbosquid Animal &442 & 100 & [0, $2\pi$] &[$\frac{1}{4}\pi$, $\frac{1}{2}\pi$]\\
Turbosquid House &563 & 100 & [0, $2\pi$] &[$\frac{1}{3}\pi$, $\frac{1}{2}\pi$]\\
Renderpeople &500 & 100 & [0, $2\pi$] &[$\frac{1}{3}\pi$, $\frac{1}{2}\pi$]\\
\bottomrule
\end{tabular}
\caption{\textbf{Dataset statistics}.}
\label{tbl:dataset}
\end{table}
\subsection{Baselines}
\label{sec:baselines}
\textbf{PointFlow}~\cite{pointflow} is a 3D point cloud generative model based on continuous normalizing flows. It models the generative process by learning a distribution of distributions. Where the former, denotes the distribution of shapes, and the latter the distribution of points given a shape~\cite{pointflow}. PointFlow generates only the geometry, which is represented in the form of a point cloud. To generate the results of~\cite{pointflow}, we use the original source code provided by the authors\footnote{PointFlow: \url{https://github.com/stevenygd/PointFlow} (MIT License)} and train the models on our data. To compute the metrics based on LFD, we convert the output point clouds (10k points) to a mesh representation using Open3D~\cite{Zhou2018open3d} implementation of Poisson surface reconstruction~\cite{kazhdan2006poisson}.

\textbf{OccNet}~\cite{occnet} is an implicit method for 3D surface reconstruction, which can also be applied to unconditional generation of 3D shapes. OccNet is an autoencoder that learns a continuous mapping from 3D coordinates to occupancy values, from which an explicit mesh can be extracted using marching cubes~\cite{marchingcube}. When applied to unconditional 3D shape generation, OccNet is trained as a variational autoencoder. To generate the results of~\cite{occnet}, we use the original source code provided by the authors\footnote{OccNet: \url{https://github.com/autonomousvision/occupancy_networks} (MIT License)} and train the models on our data.

\textbf{GRAF}~\cite{graf} is a generative model that tackles the problem of 3D-aware image synthesis. GRAF's underlying representation is a neural radiance field---conditioned on the shape and appearance latent codes---parameterized using a multi-layer perceptron with positional encoding. To synthesize novel views, GRAF utilizes a neural volume rendering approach similar to Nerf~\cite{nerf}. In our evaluation, we use the source code provided by the authors\footnote{GRAF: \url{https://github.com/autonomousvision/graf} (MIT License)} and train GRAF models on our data.

\textbf{Pi-GAN}~\cite{pigan} similar to GRAF, Pi-GAN also tackles the problem of 3D-aware image synthesis, but uses a Siren~\cite{sitzmann2019siren} network---conditioned on a randomly sampled noise vector---to parameterize the neural radiance field. To generate the results of Pi-GAN~\cite{pigan}, we use the original source code provided by the authors\footnote{Pi-GAN: \url{https://github.com/marcoamonteiro/pi-GAN} (License not provided)} and train the models on our data.

\textbf{EG3D}~\cite{eg3d} is a recent model for 3D-aware image synthesis. Similar to our method, EG3D builds upon the StyleGAN formulation and uses a tri-plane representation to parameterize the underlying neural radiance field. To improve the efficiency and to enable synthesis at higher resolution, EG3D utilizes neural rendering at a lower resolution and then upsamples the output using a 2D CNN. The source code of EG3D was provided to us by the authors. To generate the results, we train and evaluate EG3D on our data.

\subsection{Evaluation Metrics}
\label{sec:evaluation_metrics}
To evaluate the performance, we compare both the texture and geometry of the \textit{generated} shapes $S_g$ to the \textit{reference} ones $S_r$. 

\subsubsection{Evaluating the Geometry}

To evaluate the geometry, we use all shapes of the test set as $S_r$, and synthesize five times as many generated shapes, such that $|S_g| = 5|S_r|$, where $|\cdot|$ denotes the cardinality of a set.
Following prior work~\cite{pointflow, imnet}, we use Chamfer Distance $d_\text{CD}$ and Light Field Distance $d_\text{LFD}$~\cite{chen2017lfd} to measure the similarity of the shapes, which is in turn used to compute Coverage (COV) and Minimum Matching Distance (MMD) evaluation metrics. 

Let $X \in S_g$ denote a generated shape and $Y \in S_r$ a reference one. To compute $d_\text{CD}$, we first randomly sample $N=2048$ points $X_p \in \mathbb{R}^{N \times 3}$ and $Y_p \in \mathbb{R}^{N \times 3}$ from the surface of the shapes $X$ and $Y$, respectively\footnote{For PointFlow~\cite{pointflow}, we directly use $N$ points generated by the model.} . The $d_\text{CD}$ can then be computed as:
\begin{equation}
    d_\text{CD}(X_p,Y_p) =  \sum_{\mathbf{x} \in X_p} \min_{\mathbf{y} \in Y_p} ||\mathbf{x} - \mathbf{y}||^2_2 + \sum_{\mathbf{y} \in Y_p} \min_{\mathbf{x} \in X_p} ||\mathbf{x} - \mathbf{y}||^2_2.
\end{equation}
While Chamfer distance has been widely used in the field of 3D generative models and reconstruction~\cite{dibr,gao2020deftet,dmtet}, LFD has received a lot attention in computer graphics~\cite{chen2017lfd}. Inspired by human perception, LFD measures the similarity between the 3D shapes based on their appearance from different viewpoints. In particular, LFD renders the shapes $X$ and $Y$ (represented as explicit meshes) from a set of selected viewpoints, encodes the rendered images using Zernike moments and Fourier descriptors, and computes the similarity over these encodings. Formal definition of LFD is available in~\cite{chen2017lfd}. In our evaluation, we use the official implementation to compute $d_\text{LFD}$\footnote{LFD: \url{https://github.com/Sunwinds/ShapeDescriptor/tree/master/LightField/3DRetrieval_v1.8/3DRetrieval_v1.8} (License not provided)}.

We combine these similarity measures with the evaluation metrics proposed in \cite{achlioptas2018learning}, which are commonly used to evaluate 3D generative models:

\begin{itemize}
    \item \textbf{Coverage (COV)} measures the fraction of shapes in the \emph{reference} set that are matched to at least one of the shapes in the \emph{generated} set. Formally, COV is defined as
    \begin{equation}
    \text{COV} (S_g, S_r) = \frac{|\{\argmin_{X \in S_r}D(X,Y) \, | Y  \in S_g  \}|}{|S_r|} ,
    \end{equation}
    where the distance metric D can be either $d_\text{CD}$ or $d_\text{LFD}$. Intuitively, COV measures the diversity of the generated shapes and is able to detect mode collapse. However, COV does not measure the quality of individual generated shapes. In fact, it is possible to achieve high COV even when the generated shapes are of very low quality.  
    
    \item \textbf{Minimum Matching Distance (MMD)} complements COV metric, by measuring the quality of the individual generated shapes. Formally, MMD is defined as 
    \begin{equation}
    \text{MMD} (S_g, S_r) = \frac{1}{|S_r|} \sum_{X \in S_r} \min_{Y \in S_g}  D(X, Y),
    \end{equation}
    where D can again be either $d_\text{CD}$ or $d_\text{LFD}$. Intuitively, MMD measures the quality of the generated shapes by comparing their geometry to the closest reference shape.

\end{itemize}

\subsubsection{Evaluating the Texture and Geometry}

To evaluate the quality of the generated textures, we adopt the Fréchet Inception Distance (FID) metric, commonly used to evaluate the synthesis quality of 2D images. In particular, for each category,  we render 50k views of the generated shapes (one view per shape) from the camera poses randomly sampled from the predefined camera distribution, and use all the images in the test set. We then encode these images using a pretrained Inception v3~\cite{szegedy2016rethinking} model\footnote{Inception network checkpoint path: \url{http://download.tensorflow.org/models/image/imagenet/inception-2015-12-05.tgz}}, where we consider the output of the last pooling layer as our final encoding. The FID metric can then be computed as:
\begin{equation}
    \text{FID}(S_g, S_r) =  ||\bm{\mu}_g - \bm{\mu}_r||_2^2 + \text{Tr}[\bm{\Sigma}_g + \bm{\Sigma}_r - 2(\bm{\Sigma}_g \bm{\Sigma}_r)^{1/2}]||,
\end{equation}
where Tr denotes the trace operation. $\bm{\mu}_g$ and $\bm{\Sigma}_g$ are the mean value and covariance matrix of the generated image encoding, while $\bm{\mu}_r$ and $\bm{\Sigma}_r$ are obtained from the encoding of the test images.

As briefly discussed in the main paper, we use two variants of FID, which differ in the way in which the 2D images are rendered. In particular, for FID-Ori, we directly use the neural volume rendering of the 3D-aware image synthesis methods to obtain the 2D images. This metric favours the baselines that were designed to directly generate valid 2D images through neural rendering. Additionally, we propose a new metric, FID-3D, which puts more emphasis on the overall quality of the generated 3D shape. Specifically, for the baselines which do not output a textured mesh, we extract the geometry from their underlying neural field using marching cubes~\cite{marchingcube}. Then, we find the intersection point of each pixel ray with the generated mesh and use the 3D location of the intersected point to query the RGB value from the network. In this way, the rendered image is a more faithful representation of the underlying 3D shape and takes the quality of both geometry and texture into account. Note that FID-3D and FID-Ori are identical for methods that directly generate textured 3D meshes, as it is the case with {\ourmodel}.
\section{Additional Results on the Unconditioned Shape Generation}
\label{sec:more_exp_results}
In this section we provide additional results on the task of unconditional 3D shape generation. First, we perform additional qualitative comparison of {\ourmodel} with the baselines in Section~\ref{sec:additional_qualitative_comparison}. Second, we present further qualitative results of {\ourmodel} in Section~\ref{sec:additional_results_ours}. Third, we provide additional ablation studies in Section~\ref{sec:additional_ablation}. We also analyse the robustness and effectiveness of {\ourmodel}. Specifically, in Sec.~\ref{sec:noisy_camera} and~\ref{sec:noisy_mask}, we evaluate {\ourmodel} trained with noisy cameras and 2D silhouettes predicted by 2D segmentation networks. We further provide addition experiments on StyleGAN generated realistic dataset from GANverse3D~\cite{ganverse3d} in Sec.~\ref{sec:real_image}. Finally,
we provide additional comparison with EG3D~\cite{eg3d} on human character generation in Sec.~\ref{sec:compare_on_human}.

\subsection{Additional Qualitative Comparison with the Baselines}
\label{sec:additional_qualitative_comparison}
\paragraph{Comparing the Geometry of Generated Shapes} We provide additional visualization of the 3D shapes generated by {\ourmodel} and compare them to the baseline methods in Figure~\ref{fig:additional_gen_geo}. {\ourmodel} is able to generate shapes with complex geometry, different topology, and varying genus. When compared to the baselines, the shapes generated by {\ourmodel} contain more details and are more diverse.  

\paragraph{Comparing the Synthesized Images} We provide additional results on the task of 2D image generation in Figure~\ref{fig:additional_gen_imgs}. Even though {\ourmodel} is not designed for this task, it produces comparable results to the strong baseline EG3D~\cite{eg3d}, while significantly outperforming other baselines, such as PiGAN~\cite{pigan} and GRAF~\cite{graf}. Note that {\ourmodel} directly outputs 3D textured meshes, which are compatible with standard graphics engines, while extracting such representation from the baselines is non-trivial.

\begin{figure*}[t!]
\centering
\includegraphics[width=0.96\textwidth,trim=0 100 0 100,clip]{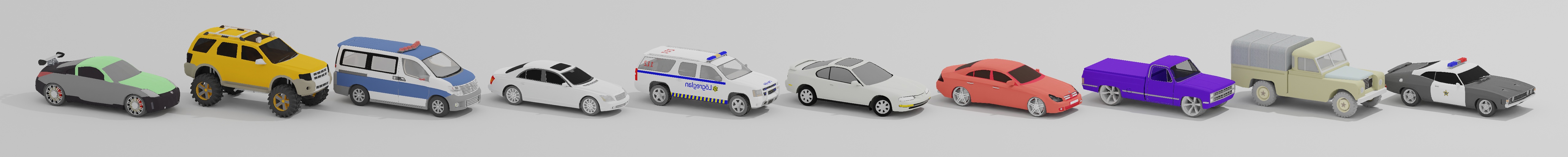}
\includegraphics[width=0.96\textwidth,trim=0 0 0 10,clip]{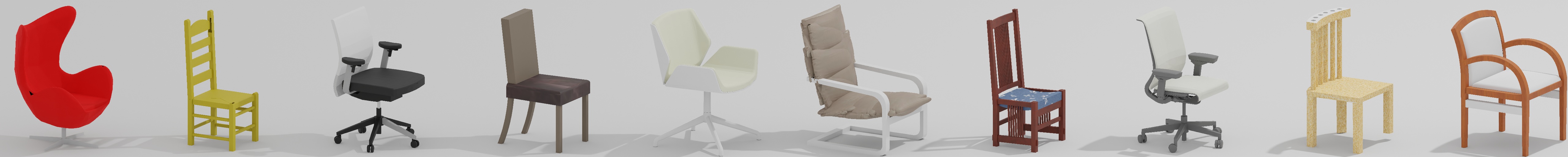}
\includegraphics[width=0.96\textwidth,trim=0 10 0 10,clip]{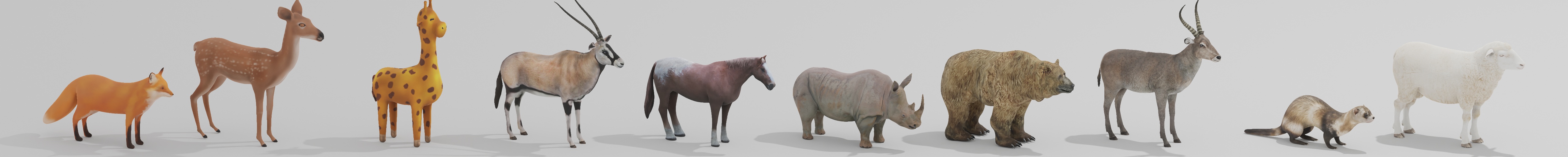}
\includegraphics[width=0.96\textwidth,trim=0 0 0 200,clip]{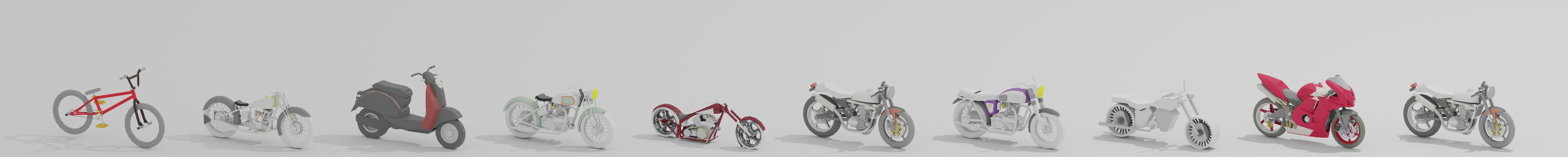}
\includegraphics[width=0.96\textwidth,trim=0 0 0 20,clip]{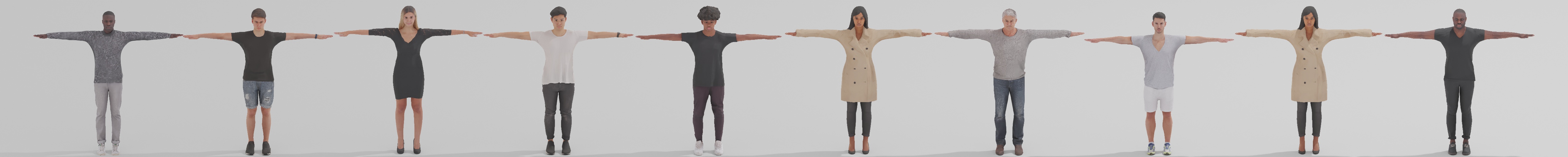}
\includegraphics[width=0.96\textwidth,trim=0 0 0 50,clip]{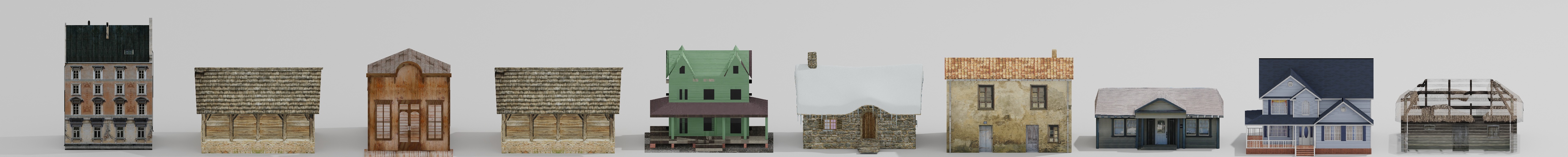}
\caption{\footnotesize \textbf{Shape retrieval of our} \textbf{generated shapes}. We retrieve the closest shape in the training set for each of shapes we showed in the Figure~\ref{fig:teaser_figure}. Our generator is able to generate novel shapes that are different from the training set}
\label{fig:teaser_figure_retrieval}
\end{figure*}

\begin{figure*}[t!]
\centering
\includegraphics[width=\textwidth]{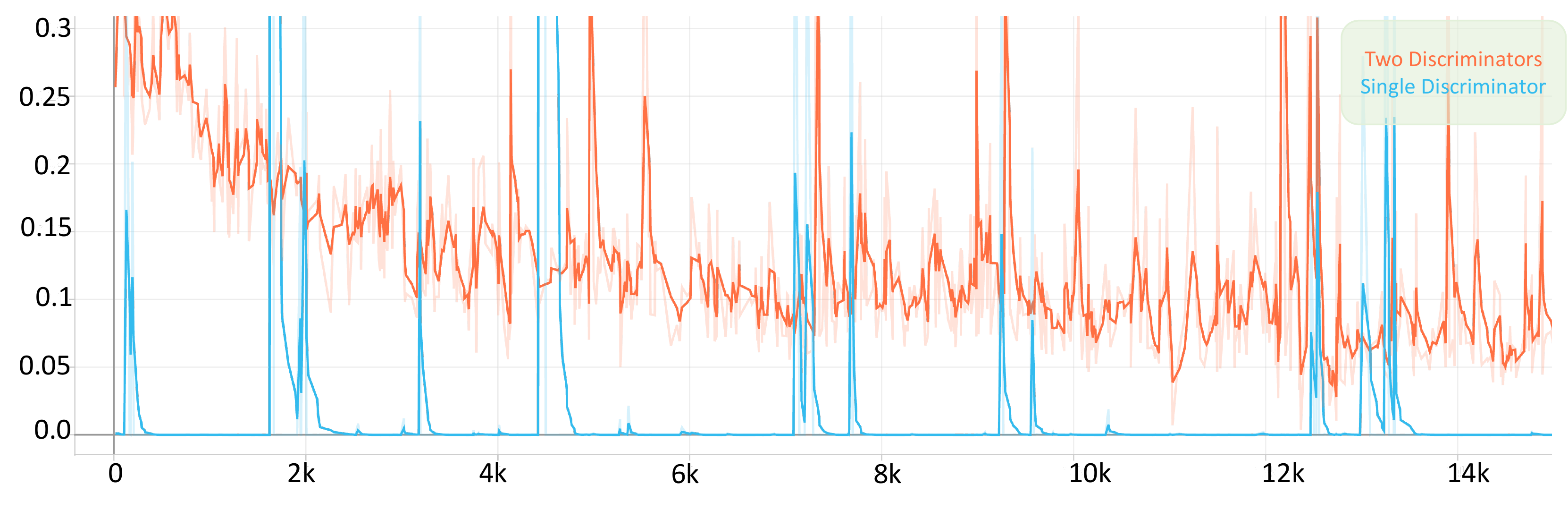}
\caption{\textbf{Training loss curve for discriminator.} We compare the training dynamics of using a single discriminator on both RGB image and 2D silhouette, with the ones using two discriminators for each image, respectively. The horizontal axis represents the number of images that the discriminators have seen during training (mod by 1000). Two discriminators greatly reduce training instability and help us obtain good results. 
} 
\label{fig:loss_curve_d}
\end{figure*}

\begin{figure*}[t!]
\centering
\includegraphics[width=\textwidth]{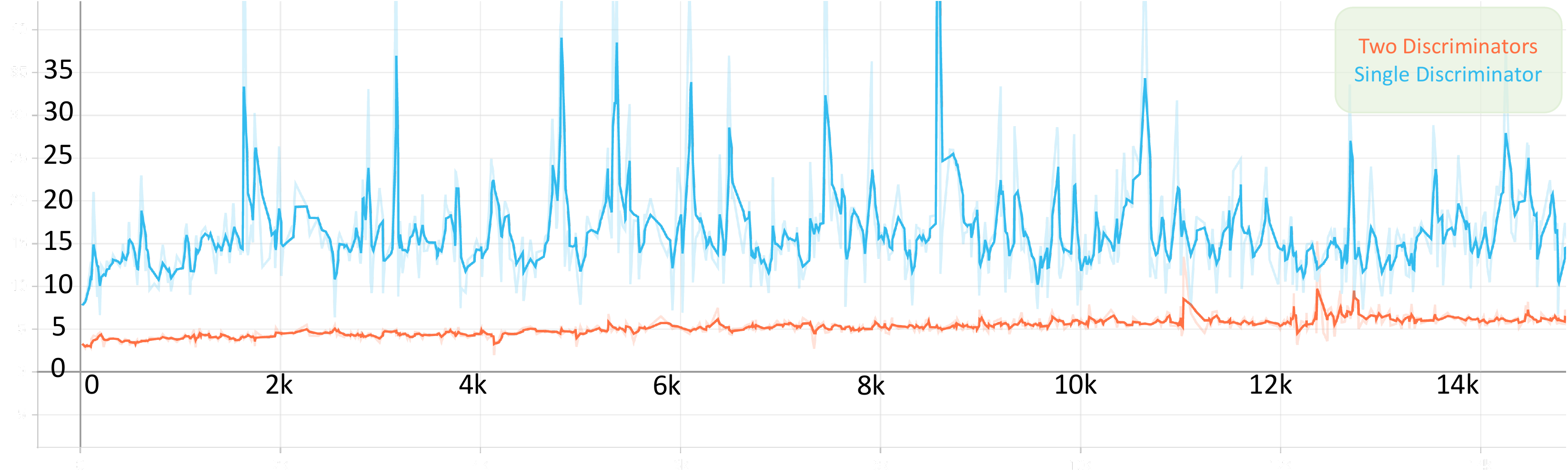}
\caption{\textbf{Training loss curve for generator.} We compare the training dynamics for using single discriminator on both RGB image and 2D silhouette with two discriminators for each image, respectively. The horizontal axis represents the number of images discriminator have seen during training (mod by 1000).
} 
\label{fig:loss_curve_g}
\end{figure*}

\subsection{Additional Qualitative Results of {\ourmodel}}
\label{sec:additional_results_ours}
We provide additional visualizations of the generated geometry and texture in Figures~\ref{fig:qual_car}-\ref{fig:qual_people}. {\ourmodel} can generate high quality shapes with diverse textures across all the categories, from chairs, cars, and animals, to motorbikes, humans, and houses. Accompanying video (\textit{demo.mp4}) contains further visualizations, including detailed $360^\circ$ turntable animations for 400+ shapes and interpolation results.

\paragraph{Closest Shape Retrieval} To demonstrate that {\ourmodel} is capable of generating novel shapes, we perform shape retrieval for our generated shapes. In particular, we retrieve the closest shape in the training set for each of shapes we showed in the Figure~\ref{fig:teaser_figure} by measuring the CD between the generated shape and all training shapes. Results are provided in Figure~\ref{fig:teaser_figure_retrieval}. All generated shapes in Figure~\ref{fig:teaser_figure} significantly differ from their closest shape in the training set, exhibiting different geometry and texture, while still maintaining the quality and diversity.

\paragraph{Volume Subdivision} We provide further qualitative results highlighting the benefits of volume subdivision in Figure~\ref{fig:ablate_vol_subdivision}. Specifically, we compare the shapes generated with and without volume subdivision on ShapeNet motorbike category. Volume subdivision enables {\ourmodel} to generate finer geometric details like handle and steel wire, which are otherwise hard to represent. 

\subsection{Additional Ablations Studies}
\label{sec:additional_ablation}
We now provide additional ablation studies in an attempt to further justify our design choices. In particular, we first discuss the design choice of using two dedicated discriminators for RGB images and 2D silhouettes, before ablating the impact of adding the camera pose conditioning to the discriminator. 

\subsubsection{Using Two Dedicated Discriminators}
We empirically find that using a single discriminator on both RGB image and silhouettes introduces significant training instability, which leads to divergence when training {\ourmodel}. We provide a comparison of the training dynamics in Figure~\ref{fig:loss_curve_d} and~\ref{fig:loss_curve_g}, where we depict the loss curves for the generator and discriminator. We hypothesize that the instability might be caused by the fact that a single discriminator has access to both geometry (from 2D silhouettes) and texture (from RGB image) of the shape, when classifying whether the image is real or not. Since we randomly initialize our geometry generator, the discriminator can quickly overfit to one aspect---either geometry or texture---and thus produces bad gradients for the other branch. A two-stage approach in which two discriminators would be used in the first stage of the training, and a single discriminator in the later stage, when the model has already learned to produce meaningful shapes, is an interesting research direction, which we plan to explore in the future.

\begin{table}
\resizebox{0.45\linewidth}{!}{
\begin{tabular}{lc}
\toprule
Model & FID\\
\midrule
{\ourmodel} w.o. Camera Condition & 11.63 \\
{\ourmodel} w/ Camera Condition & 10.25 \\
\bottomrule
\end{tabular}
\caption{\footnotesize {\bf Ablations on using camera condition}: We ablate using camera condition for discriminator. We train the model on Shapenet Car dataset.
}
\label{tbl:ablation_camera_condition}
}
\end{table}
\subsubsection{Ablation on Using Camera Condition for Discriminator}
Since we are mainly operating on synthetic datasets in which the shapes are aligned to a canonical direction, we condition the discriminators on the camera pose of each image. In this way, {\ourmodel} learns to generate shapes in the canonical orientation, which simplifies the evaluation when using metrics that assume that the input shapes are canonicalized. We now ablate this design choice. Specifically, we train another model without the conditioning and evaluate its performance in terms of FID score. Quantitative results are given in Table.~\ref{tbl:ablation_camera_condition}. We observe that removing the camera pose conditioning, only slightly degrades the performance of {\ourmodel} (-1.38 FID). This confirms that our model can be successfully trained without such conditioning, and that the primary benefit of using it is the easier evaluation.
\begin{table}[ht]
\centering
{
\begin{tabular}{lc}
\toprule
Method & FID \\
\midrule
{\ourmodel} - original & 10.25 \\
\midrule
{\ourmodel} - noisy cameras & 19.53\\
{\ourmodel} - predicted 2D silhouettes (Mask-Black) & 29.68\\
{\ourmodel} - predicted 2D silhouettes (Mask-Random) & 33.16\\
\bottomrule
\end{tabular}
{\caption{Additional quantitative results for noisy cameras and using predicted 2D silhouettes on Shapenet Car dataset.}
\label{tbl:add_exp_noisy_camera}
}

}
\end{table}

\begin{figure*}[t!]

\centering
\includegraphics[width=0.24\textwidth,trim=0 0 0 0,clip]{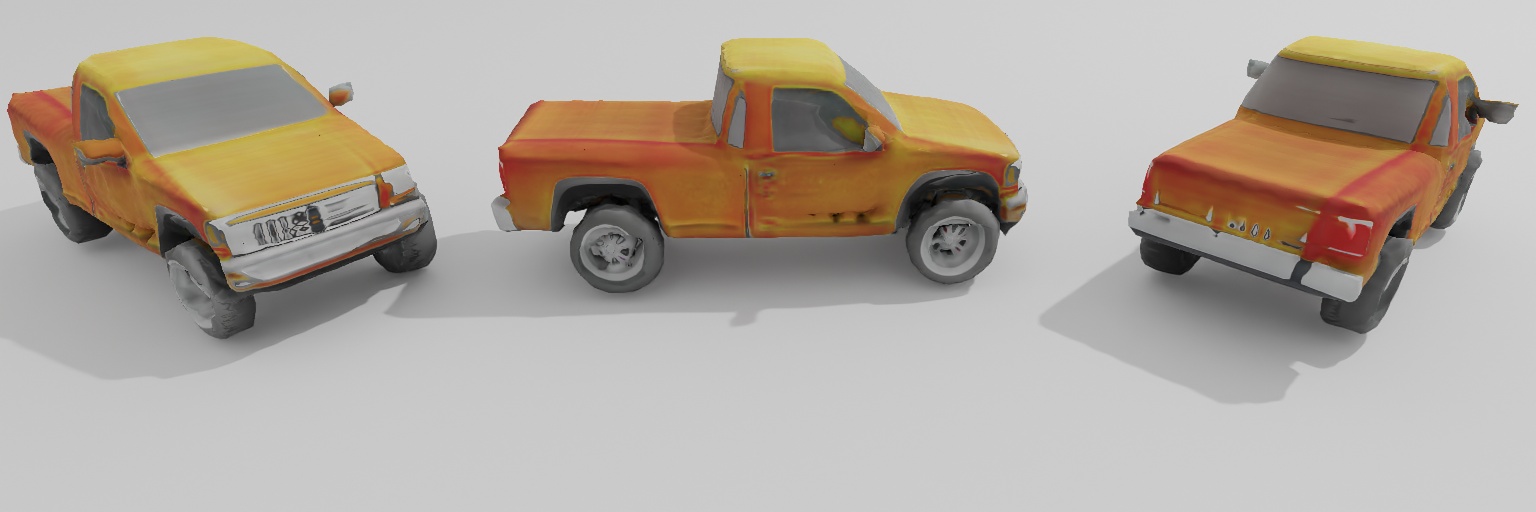}
\includegraphics[width=0.24\textwidth,trim=0 0 0 0,clip]{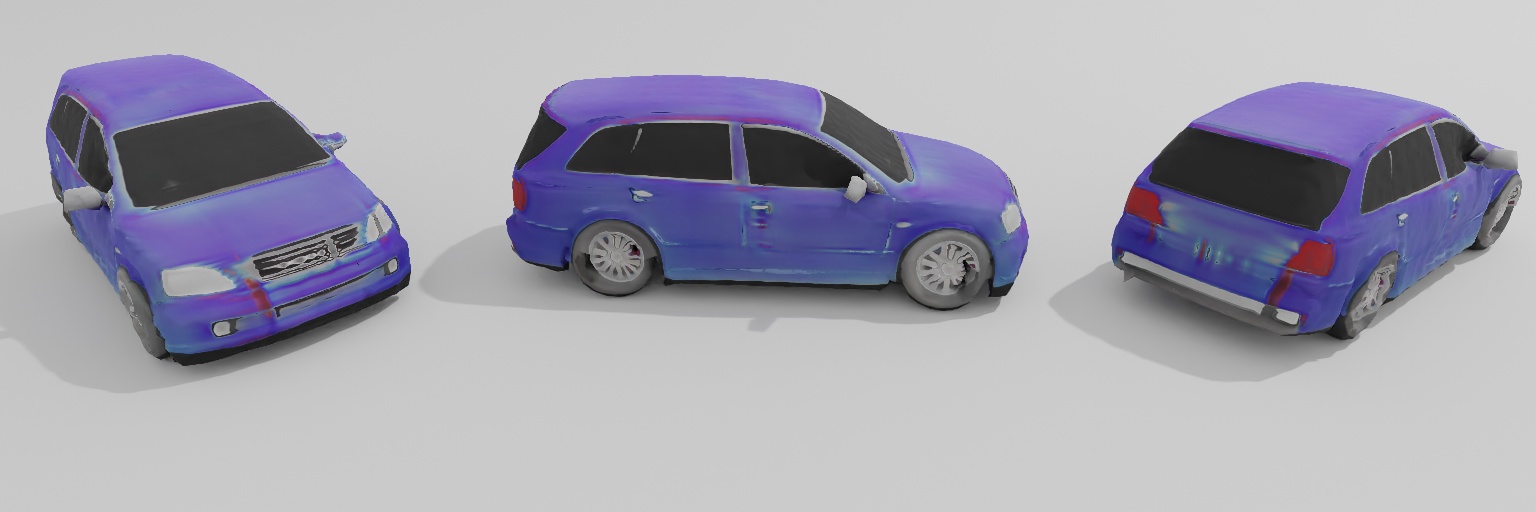}
\includegraphics[width=0.24\textwidth,trim=0 0 0 0,clip]{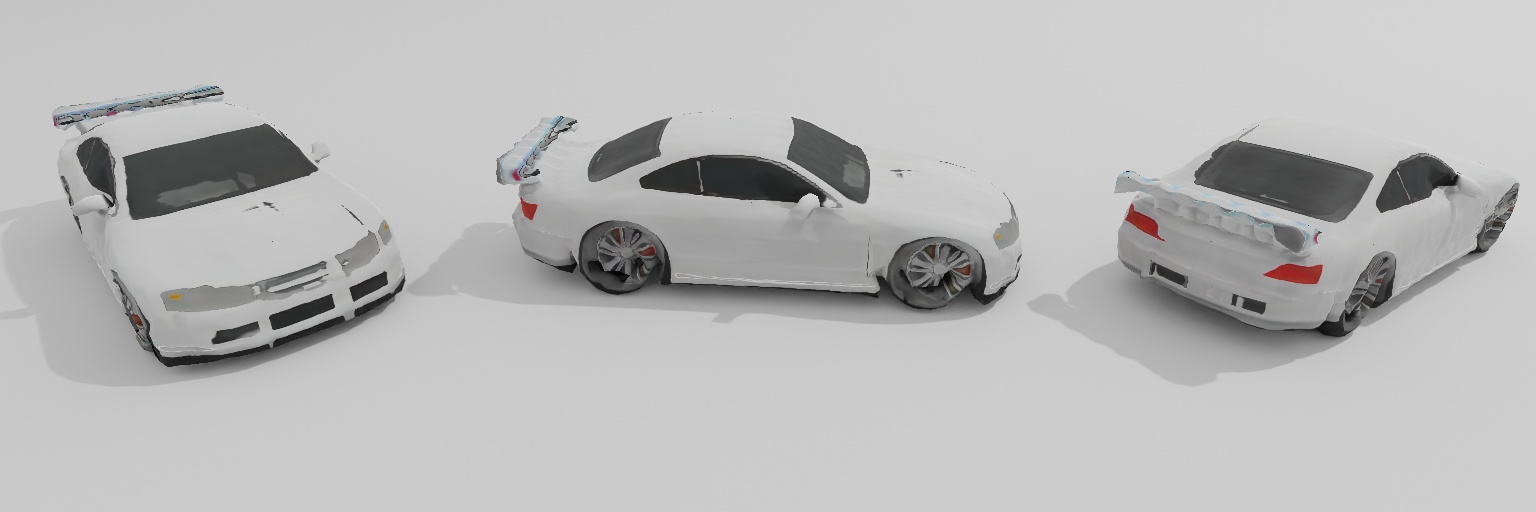}
\includegraphics[width=0.24\textwidth,trim=0 0 0 0,clip]{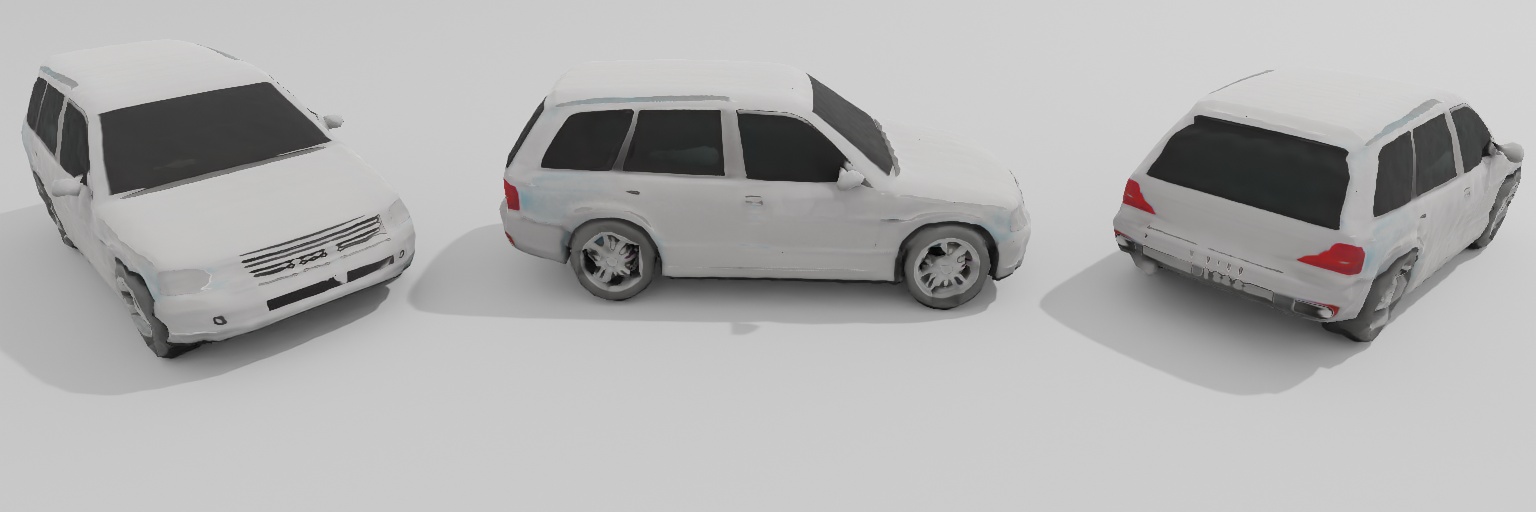}
\includegraphics[width=0.24\textwidth,trim=0 0 0 0,clip]{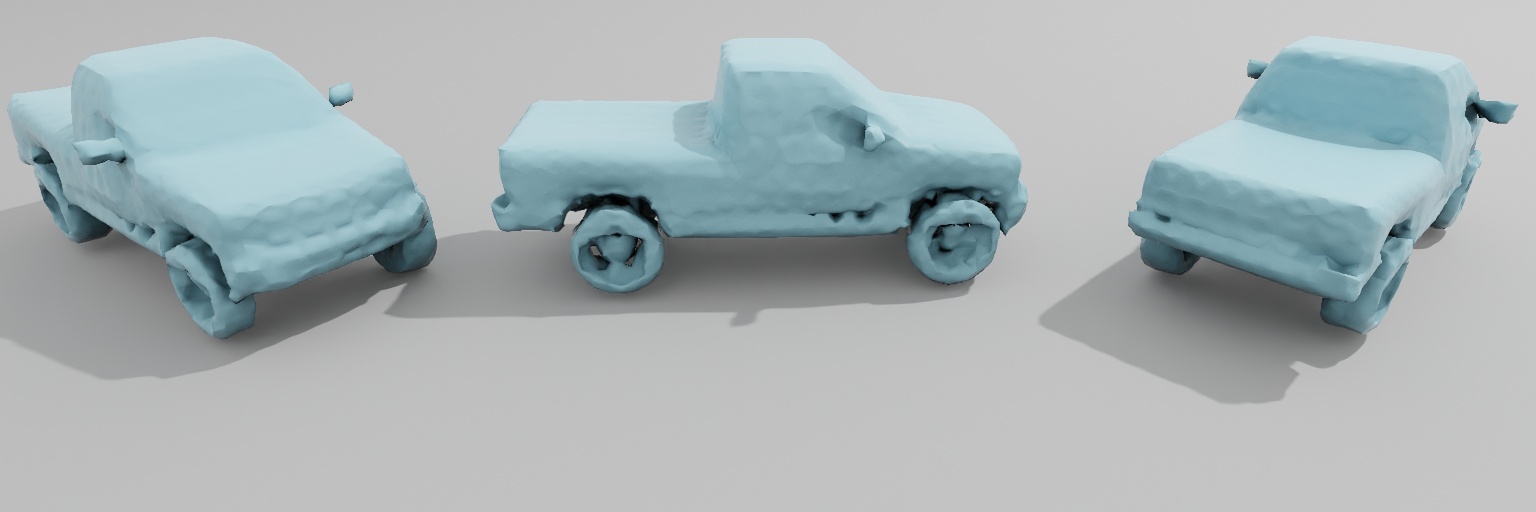}
\includegraphics[width=0.24\textwidth,trim=0 0 0 0,clip]{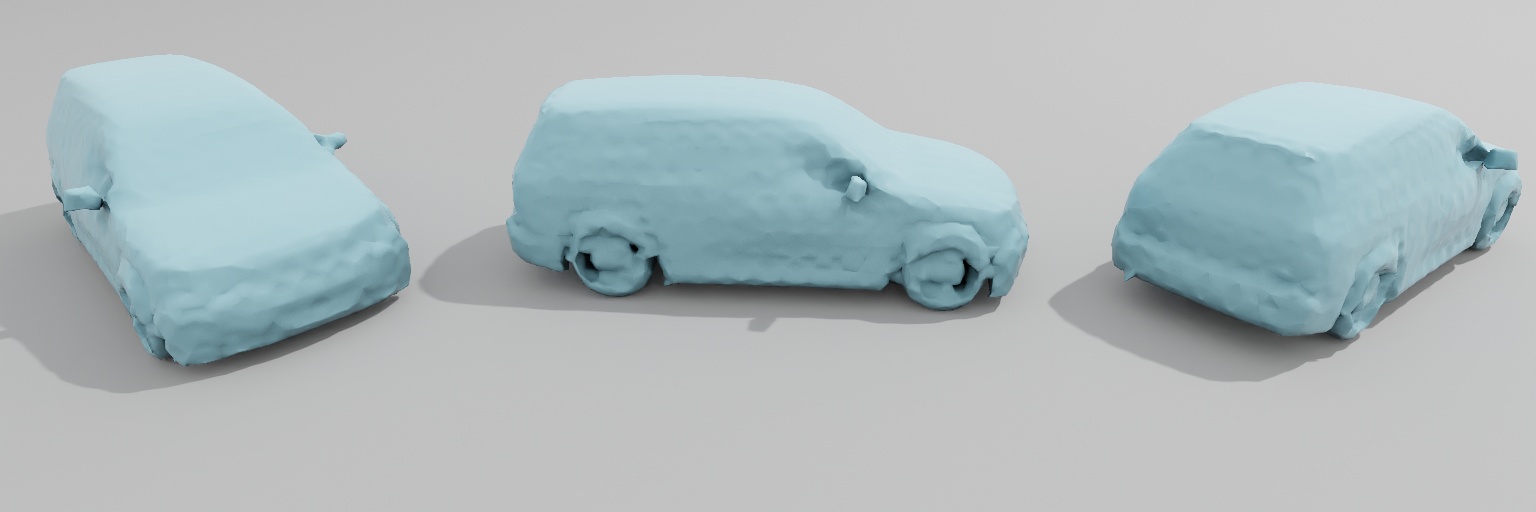}
\includegraphics[width=0.24\textwidth,trim=0 0 0 0,clip]{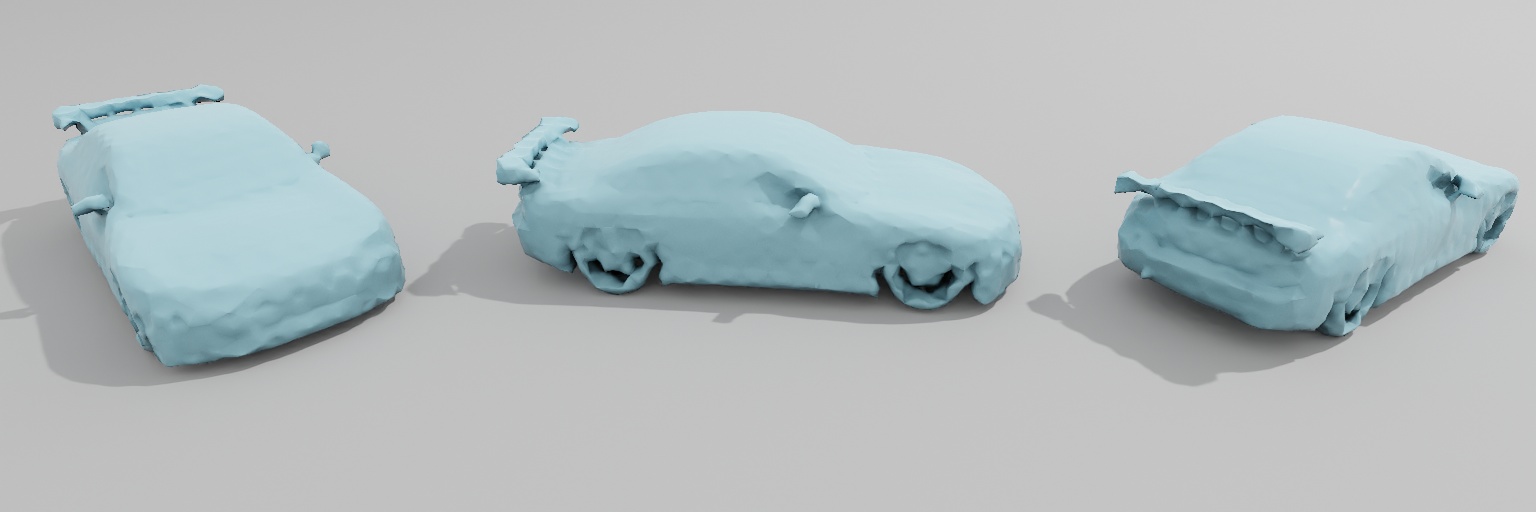}
\includegraphics[width=0.24\textwidth,trim=0 0 0 0,clip]{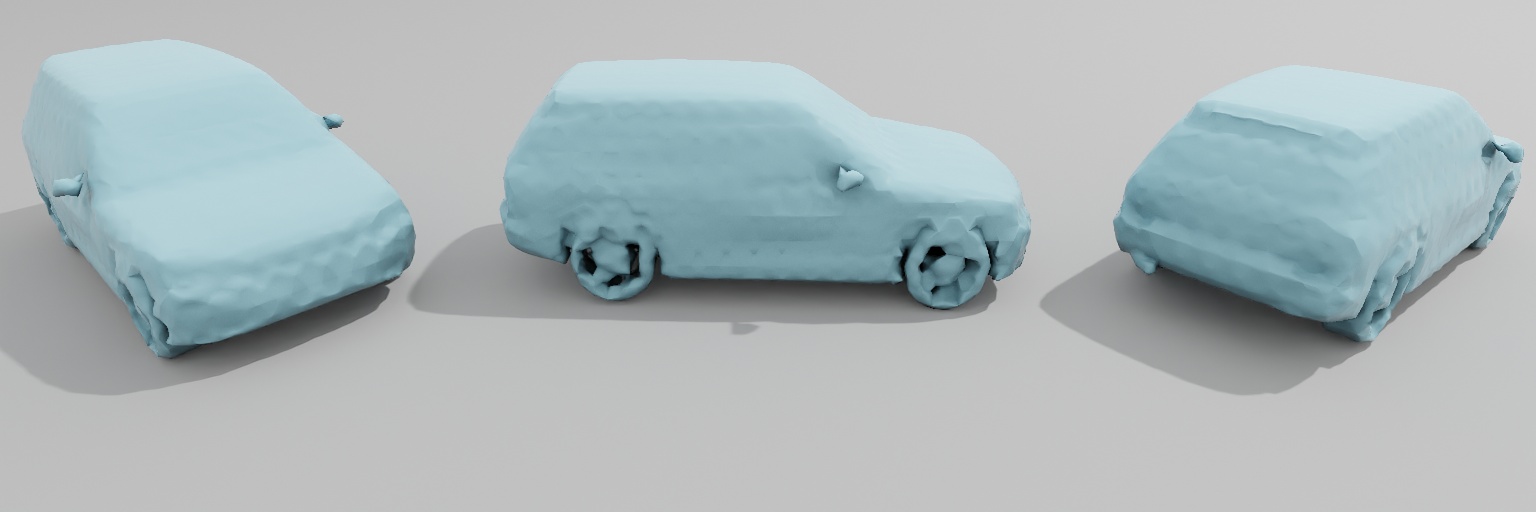}
\caption{ \footnotesize \textbf{Additional qualitative results of {\ourmodel} trained with noisy cameras.} We render generated shapes in Blender.  The visual quality is similar to original {\ourmodel} in the main paper.}
\label{fig:qualitative_results_noisy_camera}
\end{figure*}

\begin{figure*}[t!]
\centering
\includegraphics[width=0.24\textwidth,trim=0 0 0 0,clip]{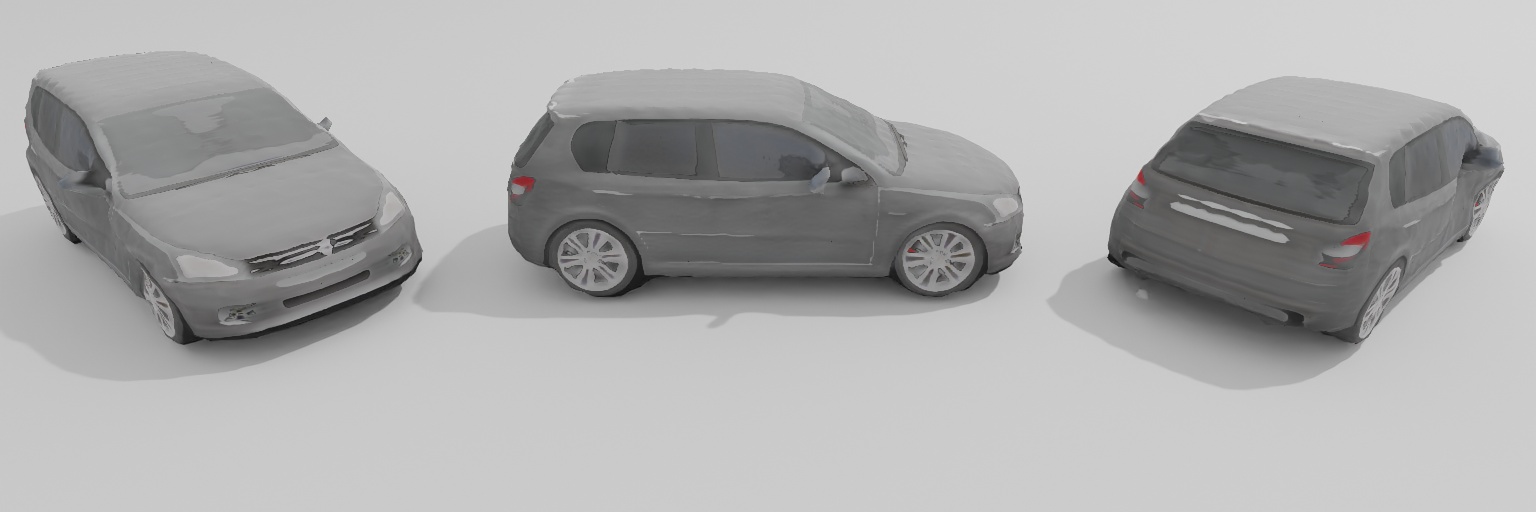}
\includegraphics[width=0.24\textwidth,trim=0 0 0 0,clip]{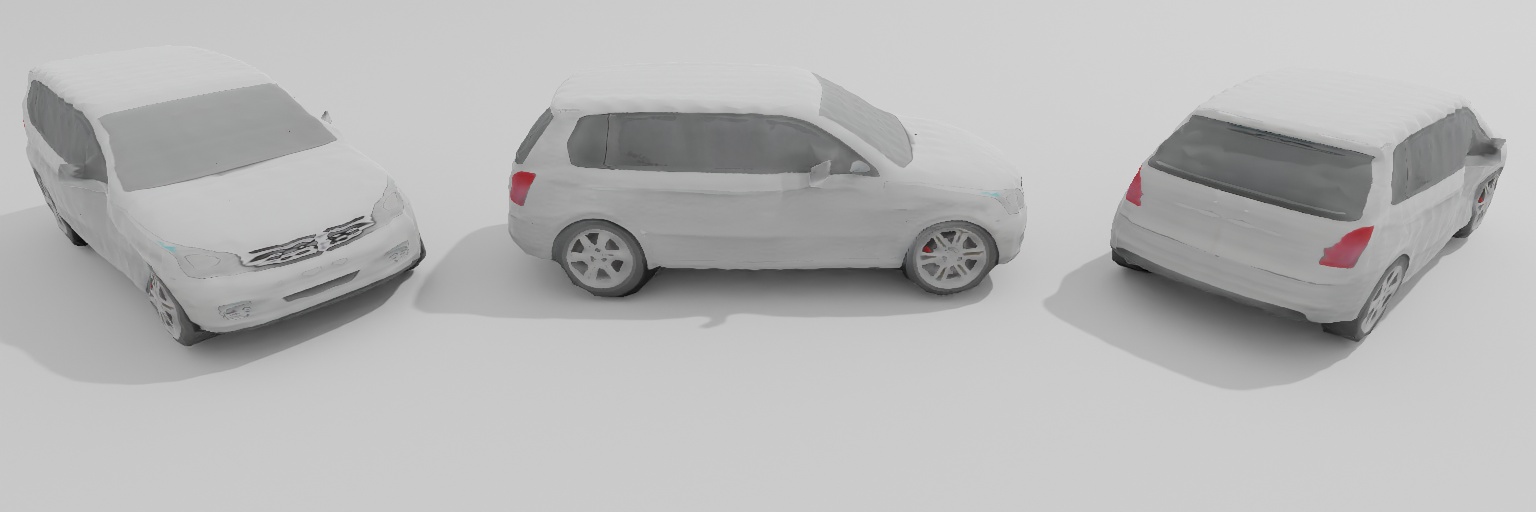}
\includegraphics[width=0.24\textwidth,trim=0 0 0 0,clip]{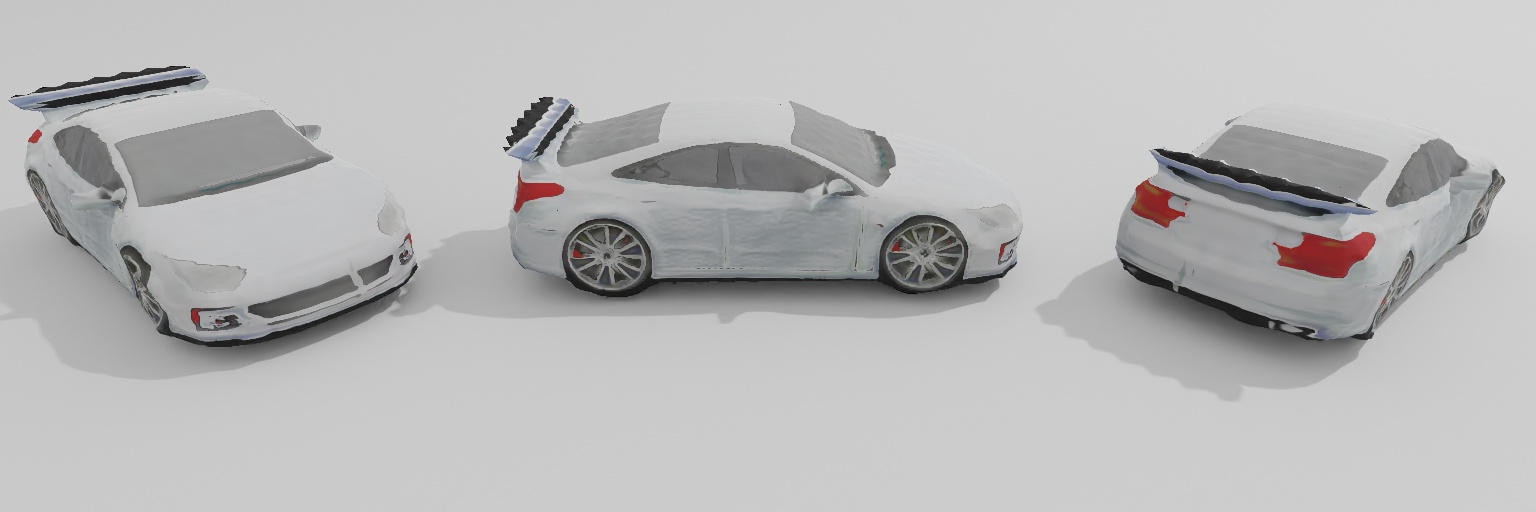}
\includegraphics[width=0.24\textwidth,trim=0 0 0 0,clip]{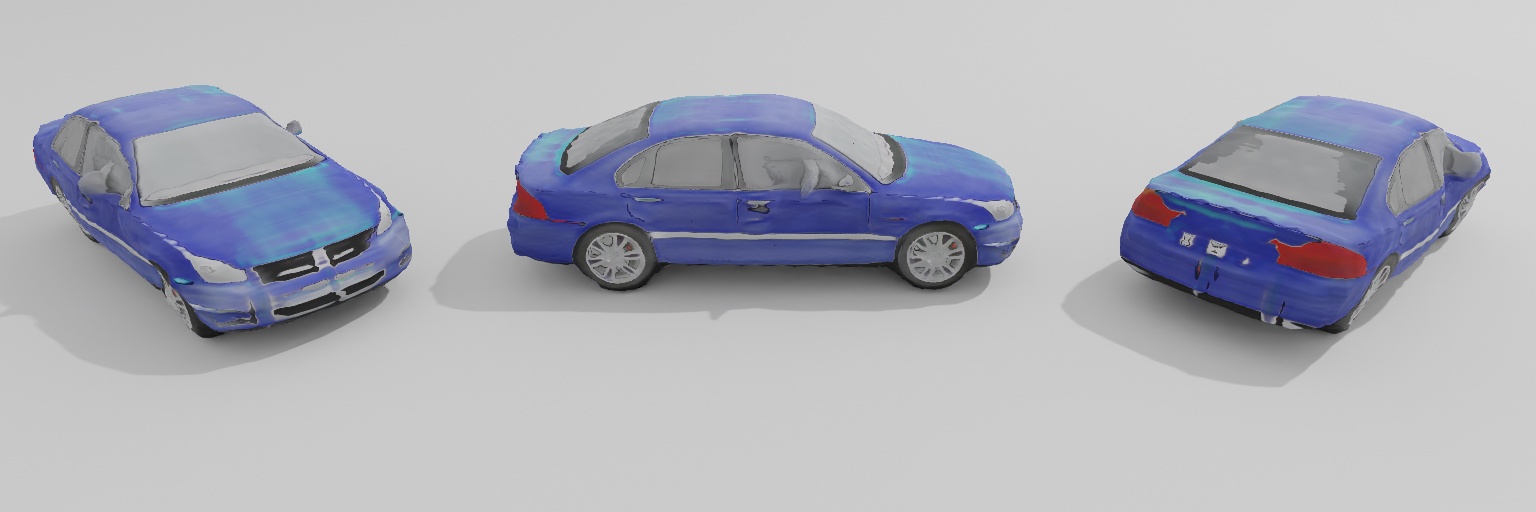}
\includegraphics[width=0.24\textwidth,trim=0 0 0 0,clip]{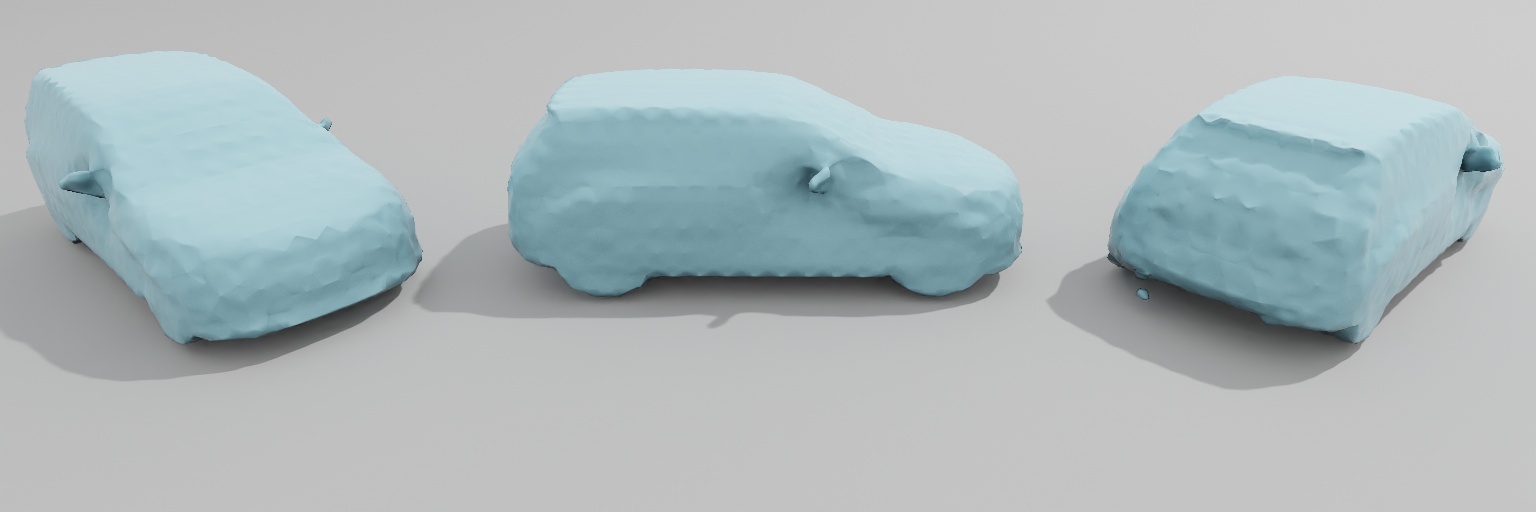}
\includegraphics[width=0.24\textwidth,trim=0 0 0 0,clip]{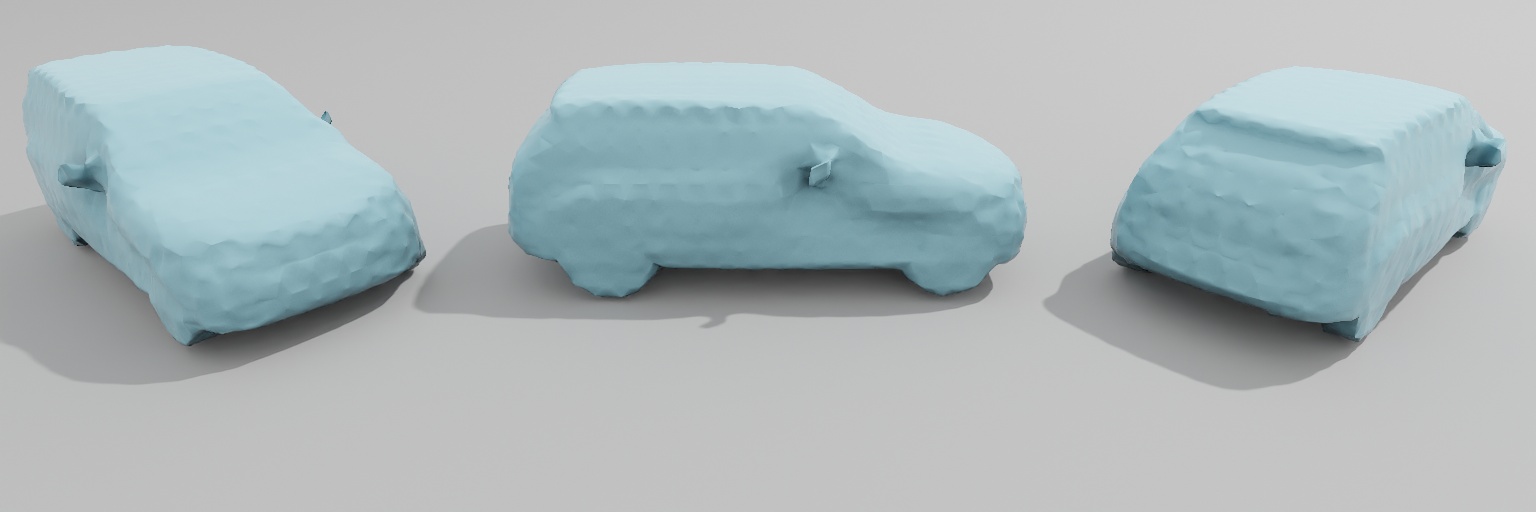}
\includegraphics[width=0.24\textwidth,trim=0 0 0 0,clip]{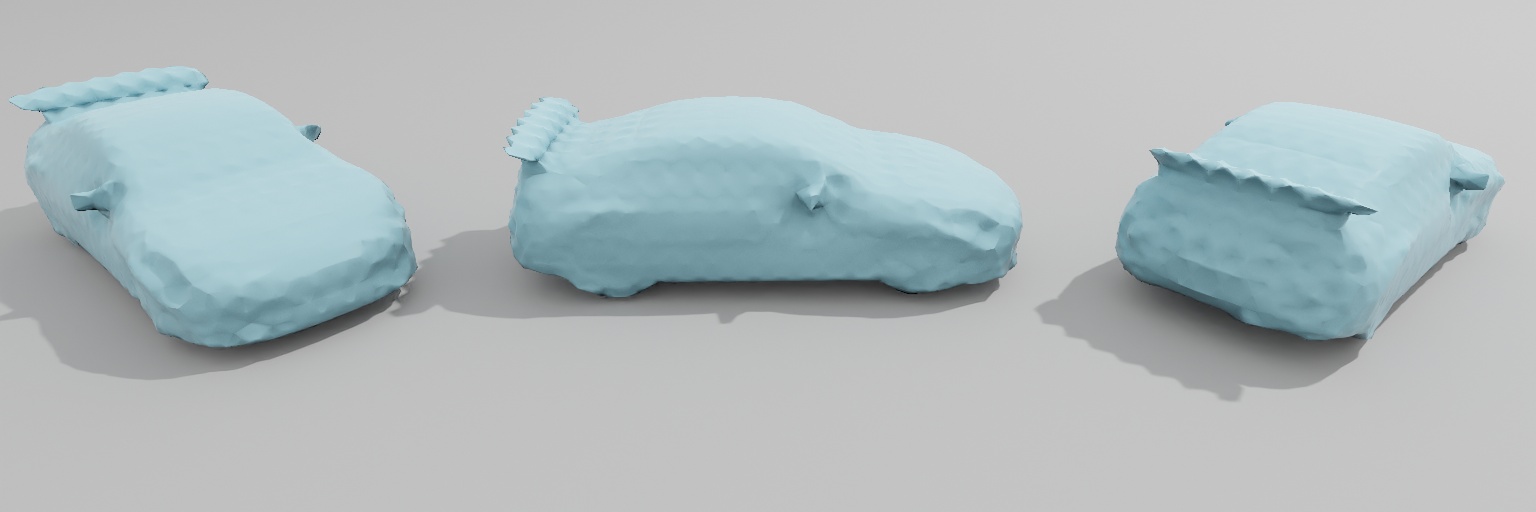}
\includegraphics[width=0.24\textwidth,trim=0 0 0 0,clip]{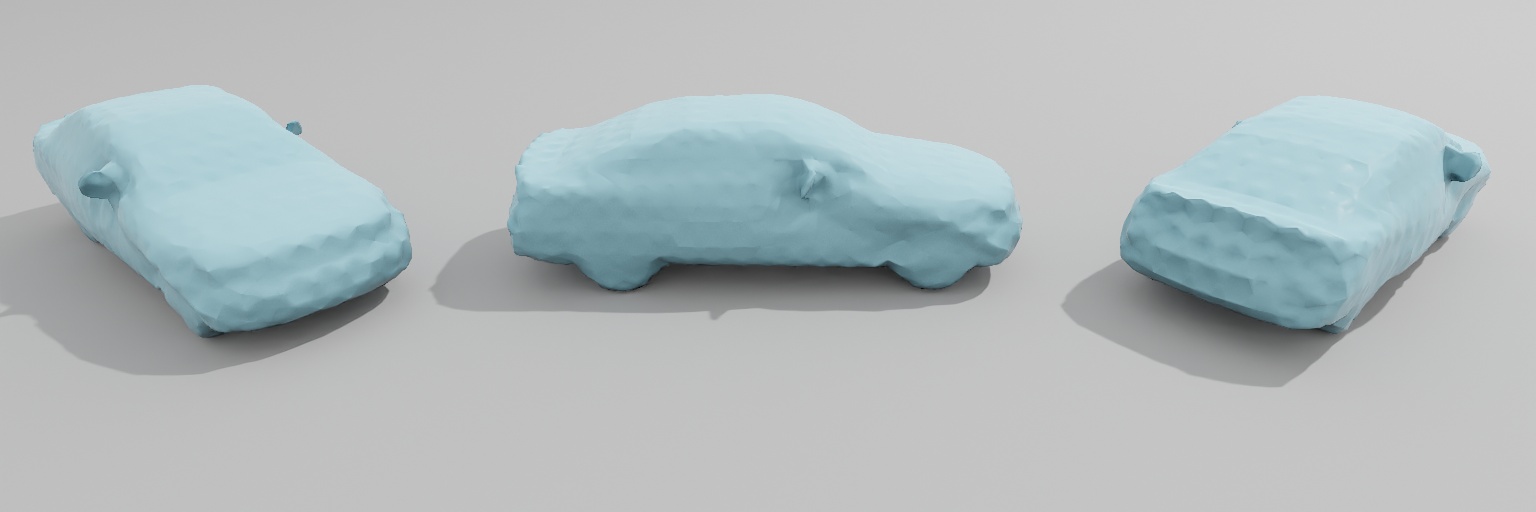}
\caption{ \footnotesize \textbf{Additional qualitative results of {\ourmodel} trained with predicted 2D silhouettes (Mask-Black).} We render generated shapes in Blender.  The visual quality is similar to original {\ourmodel} in the main paper.}
\label{fig:qualitative_results_2d_masks}
\end{figure*}

\begin{figure*}[t!]
\centering
\includegraphics[width=0.24\textwidth,trim=0 0 0 0,clip]{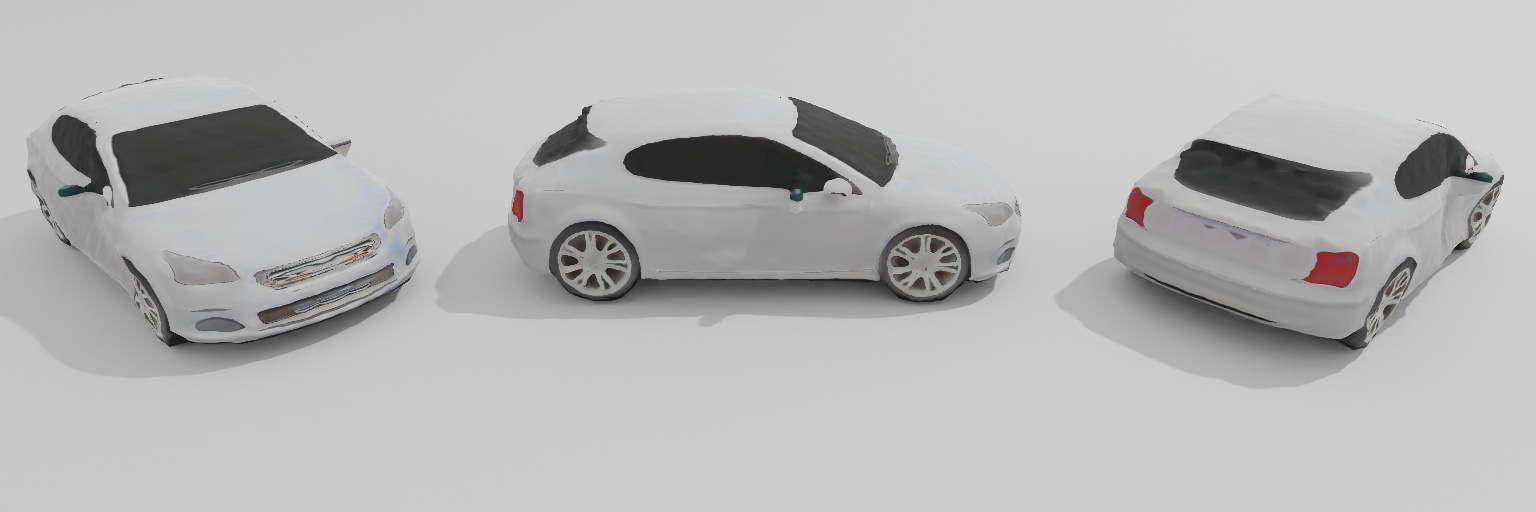}
\includegraphics[width=0.24\textwidth,trim=0 0 0 0,clip]{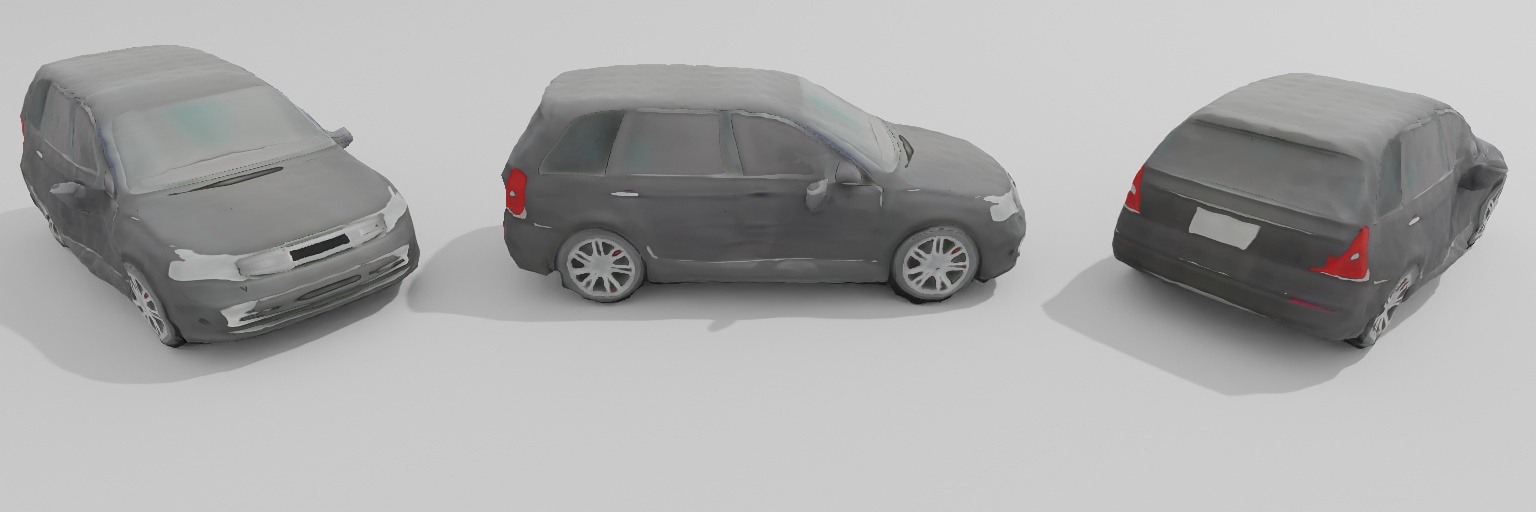}
\includegraphics[width=0.24\textwidth,trim=0 0 0 0,clip]{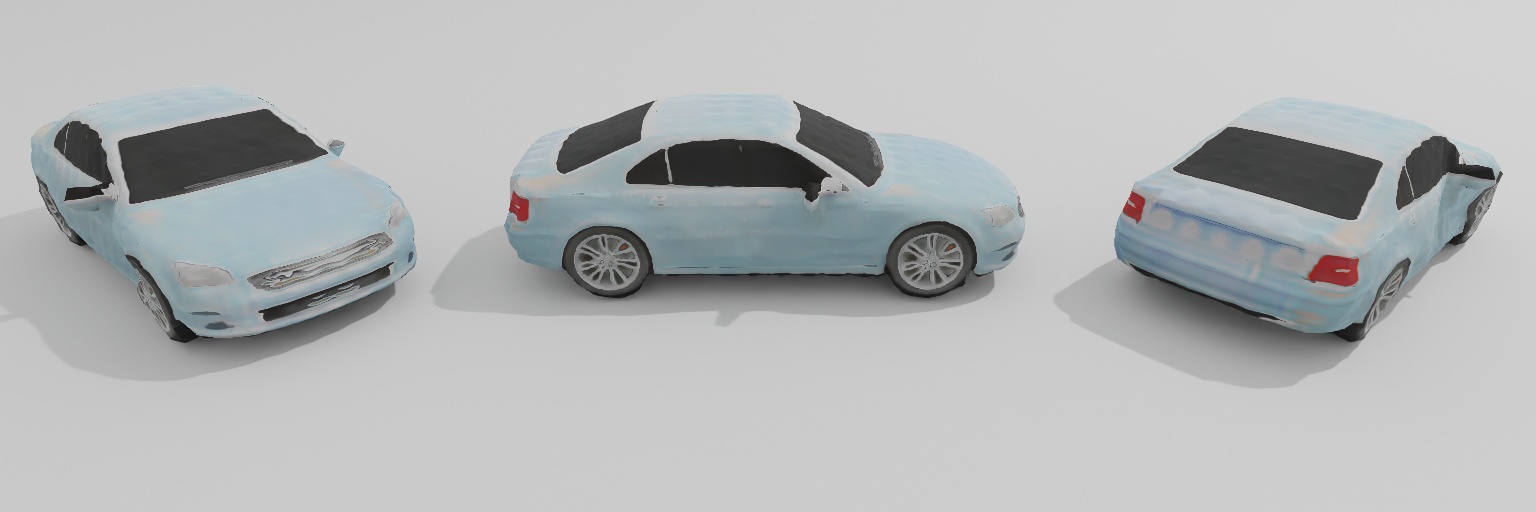}
\includegraphics[width=0.24\textwidth,trim=0 0 0 0,clip]{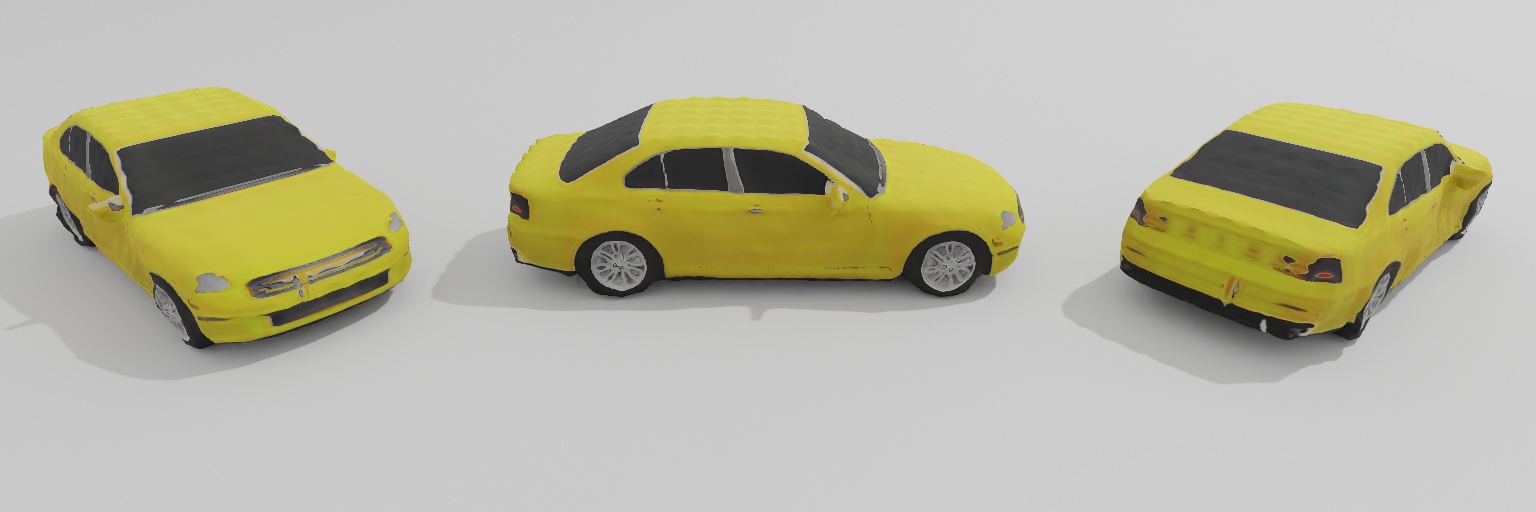}
\includegraphics[width=0.24\textwidth,trim=0 0 0 0,clip]{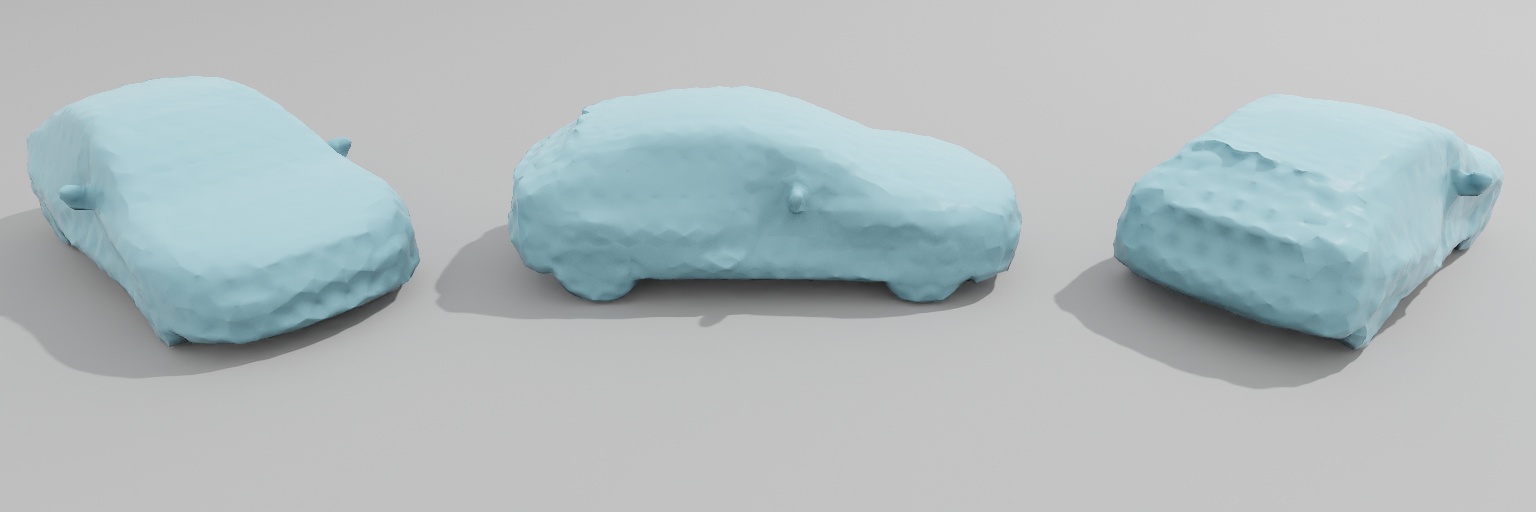}
\includegraphics[width=0.24\textwidth,trim=0 0 0 0,clip]{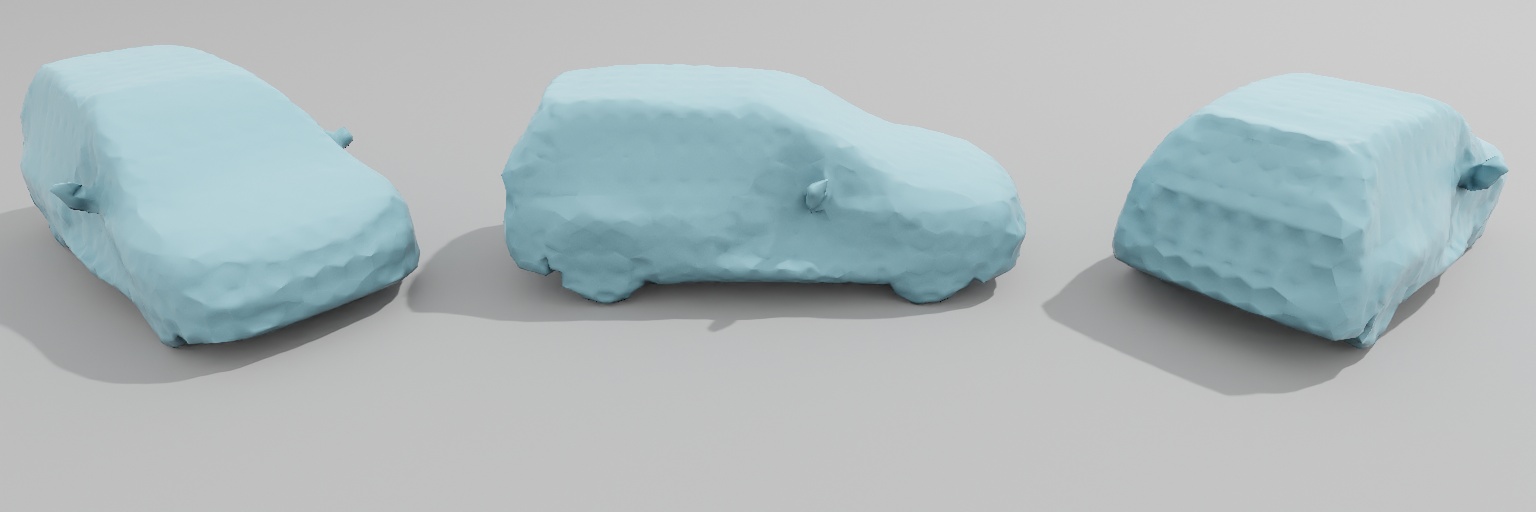}
\includegraphics[width=0.24\textwidth,trim=0 0 0 0,clip]{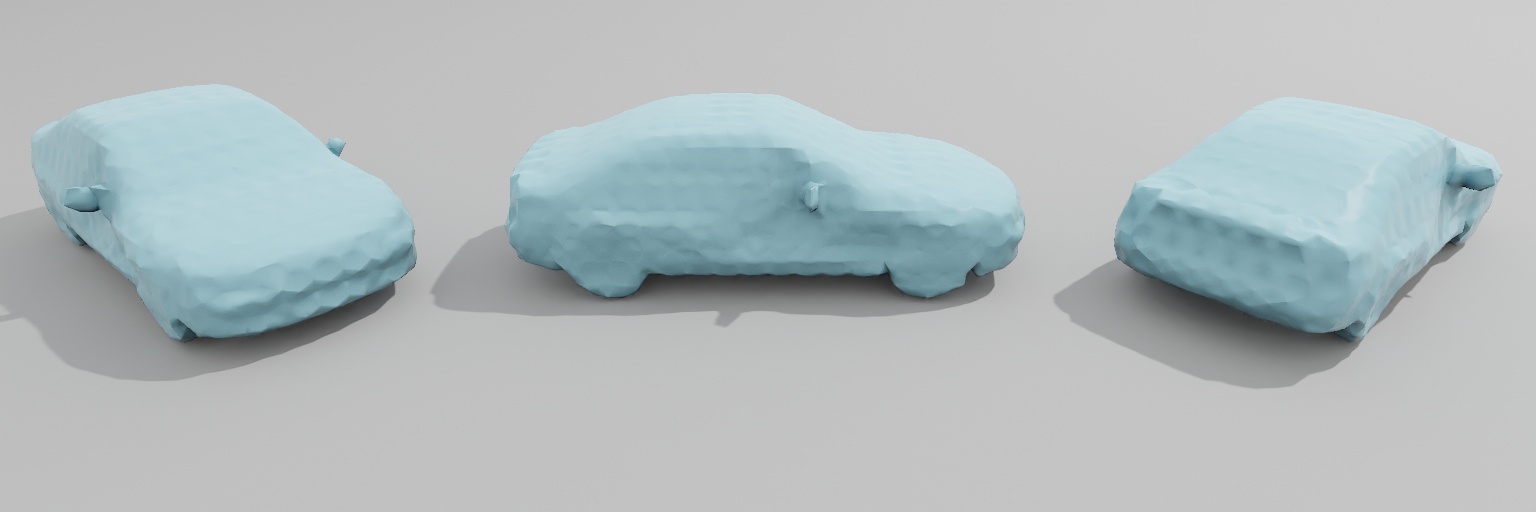}
\includegraphics[width=0.24\textwidth,trim=0 0 0 0,clip]{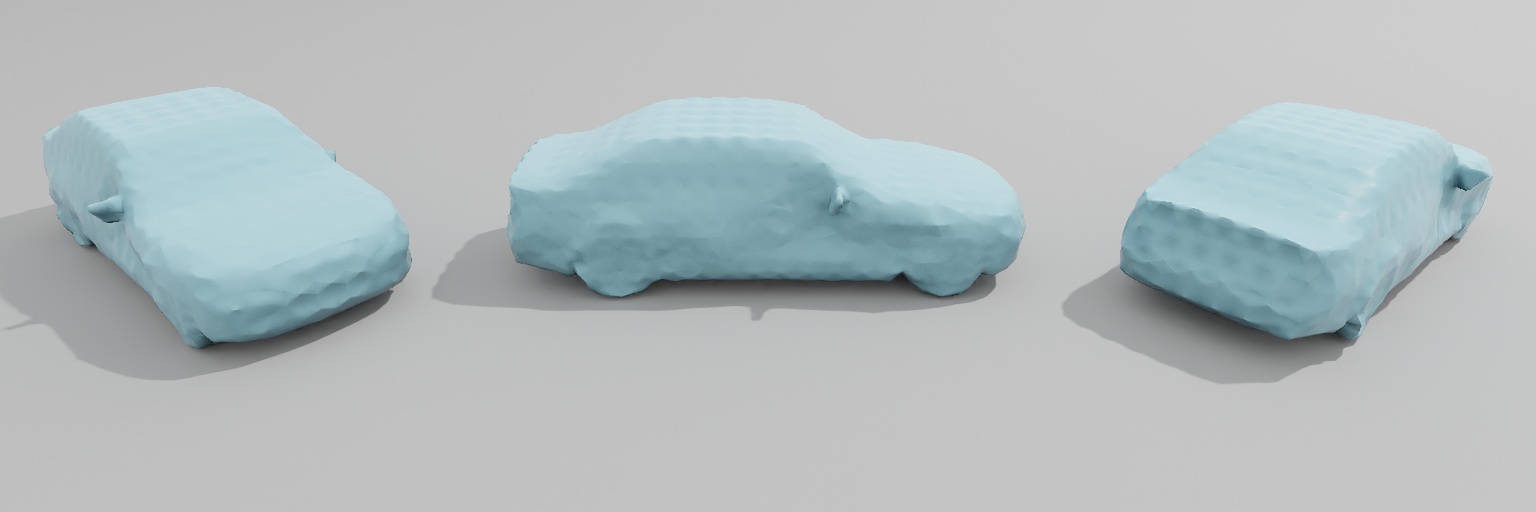}
\caption{ \footnotesize \textbf{Additional qualitative results of {\ourmodel} trained with predicted 2D silhouettes (Mask-Random).} We render generated shapes in Blender.  The visual quality is similar to original {\ourmodel} in the main paper.}
\label{fig:qualitative_results_2d_masks_random}
\end{figure*}

\subsection{Robustness to Noisy Cameras}
\label{sec:noisy_camera}
To demonstrate the robustness of {\ourmodel} to imperfect cameras poses, we add Gaussian noises to the camera poses during training. Specifically, for the rotation angle, we add a noise sampled from a Gaussian distribution with zero mean, and 10 degrees variance. For the elevation angle, we also add a noise sampled from a Gaussian distribution with zero mean, and 2 degrees variance. We use ShapeNet Car dataset~\cite{shapenet} in this experiment. 

The quantitative results are provided in Table~\ref{tbl:add_exp_noisy_camera} and qualitative examples are depicted in Figure ~\ref{fig:qualitative_results_noisy_camera}. Adding camera noise harms the FID metric, whereas we observe only little degradation in visual quality. We hypothesize that the drop in the FID is a consequence of the camera pose distribution mismatch, which occurs as result of rendering the testing dataset, used to calculate the FID score, with a camera pose distribution without added noise. Nevertheless, based on the visual quality of the generated shapes, we conclude that {\ourmodel} is robust to a moderate level of noise in the camera poses.

\subsection{Robustness to Imperfect 2D Silhouettes}
\label{sec:noisy_mask}
To evaluate the robustness of {\ourmodel} when trained with imperfect 2D silhouettes, we replace ground truth 2D masks with the ones obtained from Detectron2\footnote{\url{https://github.com/facebookresearch/detectron2}} using pretrained PointRend checkpoint, mimicking how one could obtain the 2D segmentation masks in the real world. Since our training images are rendered with the black background, we use two approaches to obtain the 2D silhouettes: i) we directly feed the original training image into Detectron2 to obtain the predicted segmentation mask (we refer to this as Mask-Black), and ii) we add a background image, randomly sampled from PASCAL-VOC 2012 dataset (we refer to this as Mask-Random). In this setting, the pretrained Detectron2 model achieved 97.4 and 95.8 IoU for the Mask-Black and Mask-Random versions, respectively. We again use the Shapenet Car dataset~\cite{shapenet} in this experiment. 

\paragraph{Experimental Results}  Quantitative results are summarized in Table~\ref{tbl:add_exp_noisy_camera}, with qualitative examples provided in Figures~\ref{fig:qualitative_results_2d_masks} and~\ref{fig:qualitative_results_2d_masks_random}. Although we observe drop in the FID scores, qualitatively the results are still similar to the original results in the main paper. Our model can generate high quality shapes even when trained with the imperfect masks. Note that, in this scenario, the training data for {\ourmodel} is different from the testing data that is used to compute the FID score, which could be one of the reasons for worse performance.

\begin{figure*}[t!]
\centering
\includegraphics[width=0.24\textwidth,trim=0 0 0 0,clip]{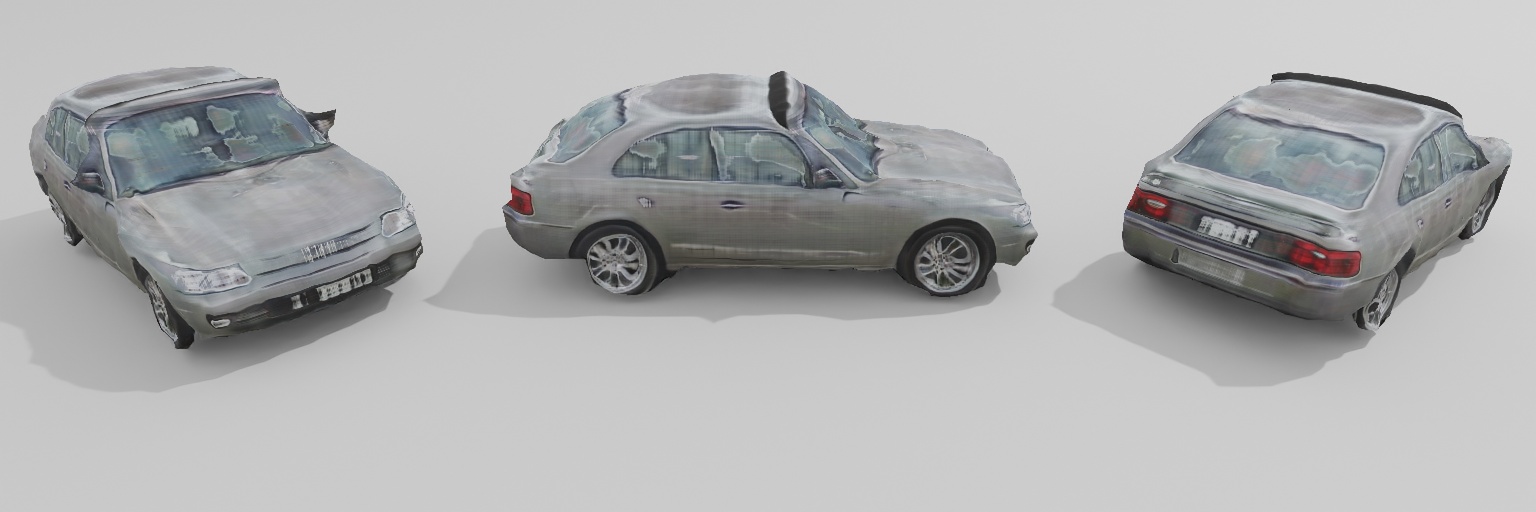}
\includegraphics[width=0.24\textwidth,trim=0 0 0 0,clip]{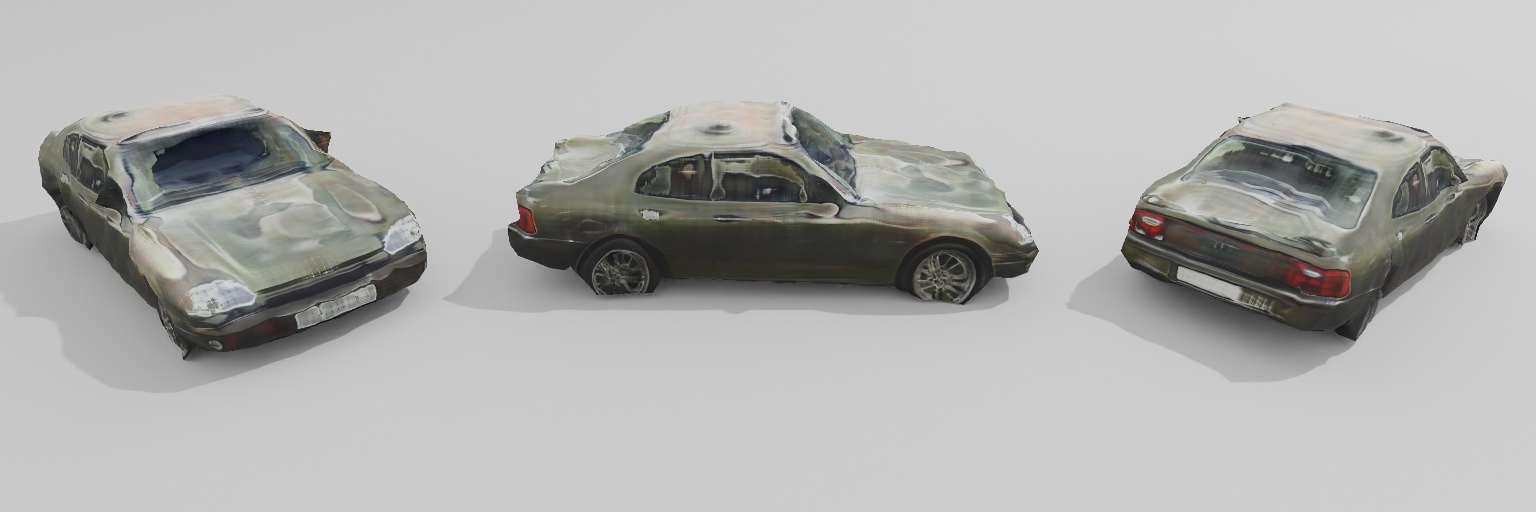}
\includegraphics[width=0.24\textwidth,trim=0 0 0 0,clip]{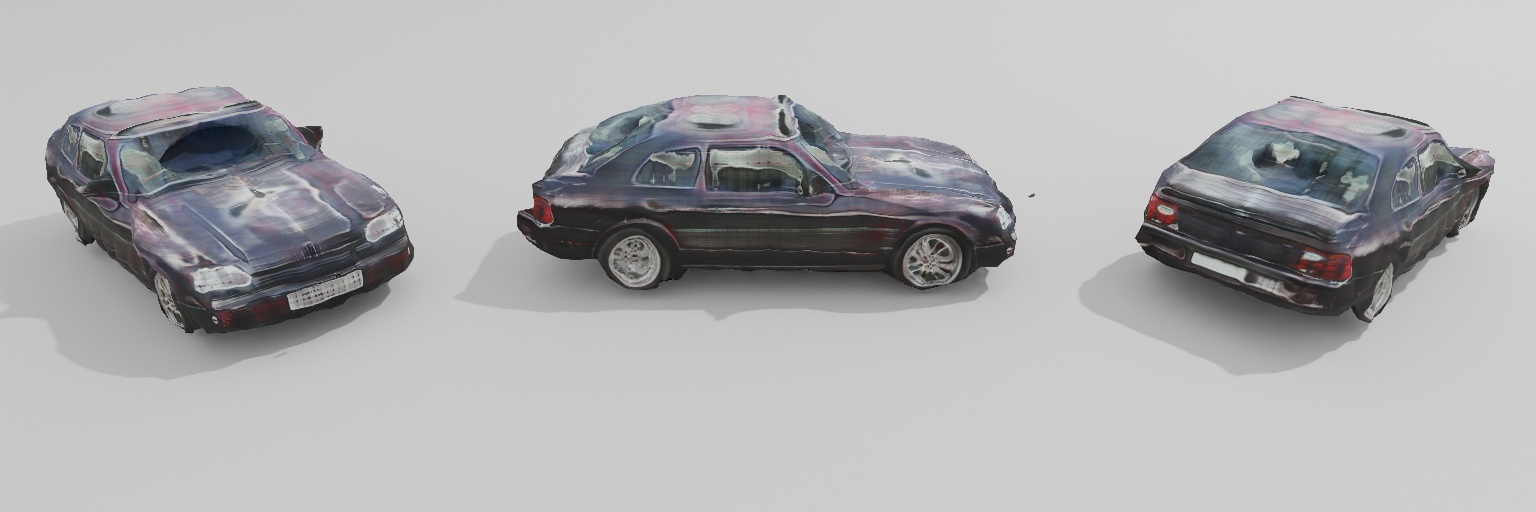}
\includegraphics[width=0.24\textwidth,trim=0 0 0 0,clip]{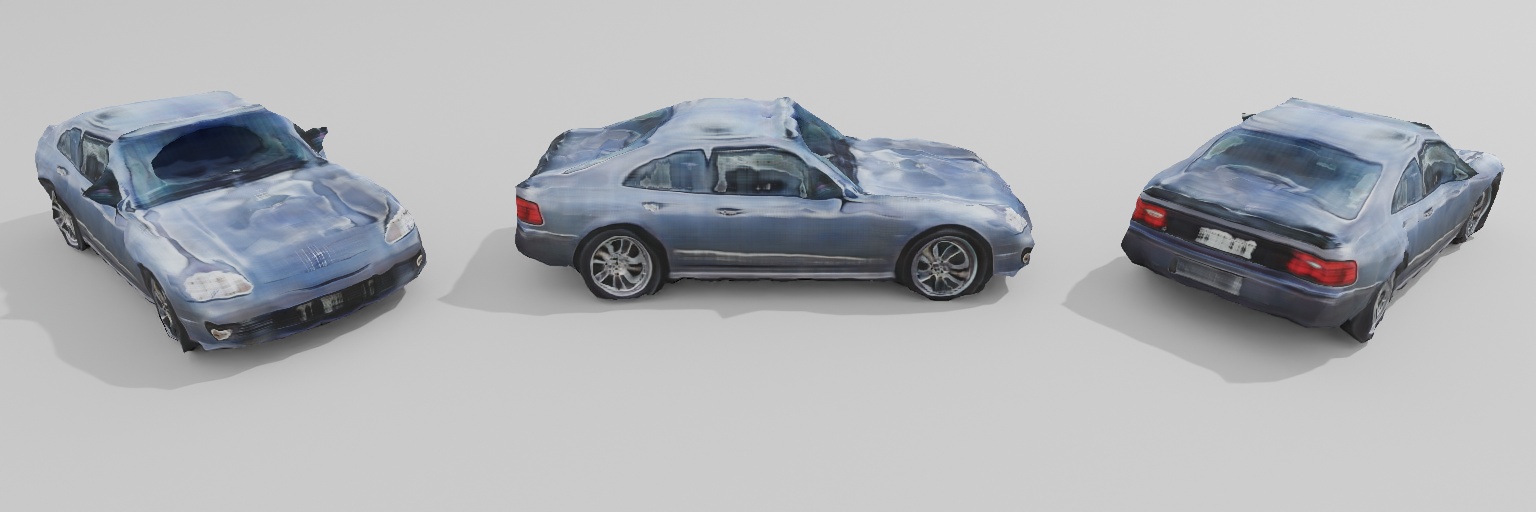}
\includegraphics[width=0.24\textwidth,trim=0 0 0 0,clip]{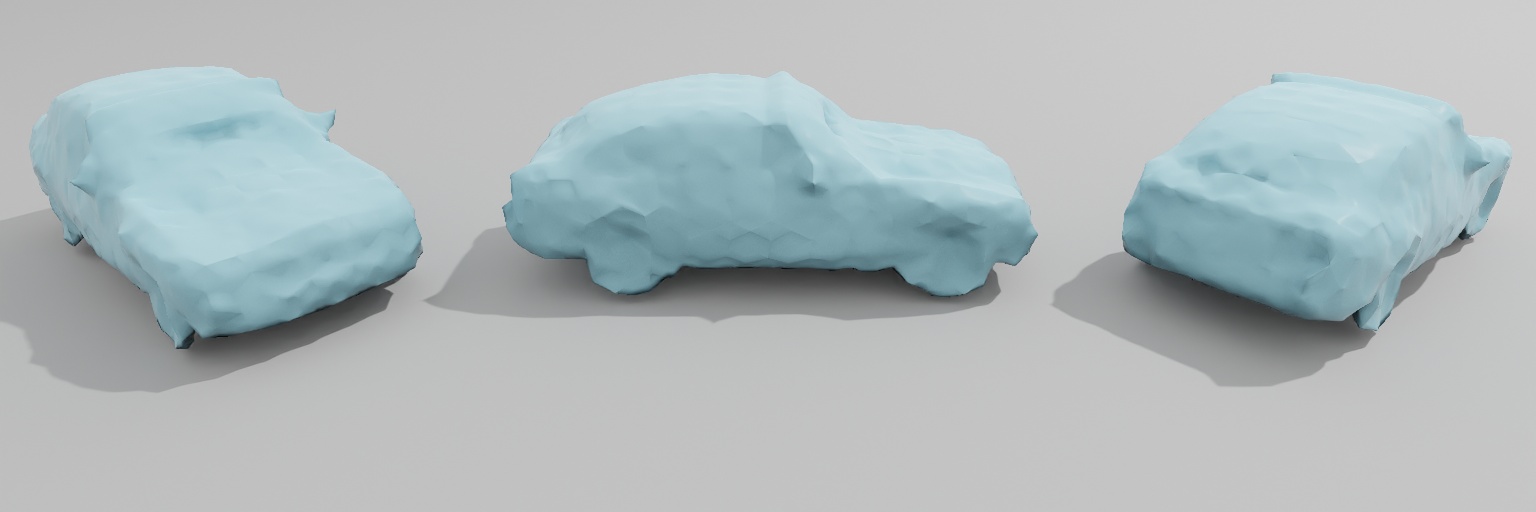}
\includegraphics[width=0.24\textwidth,trim=0 0 0 0,clip]{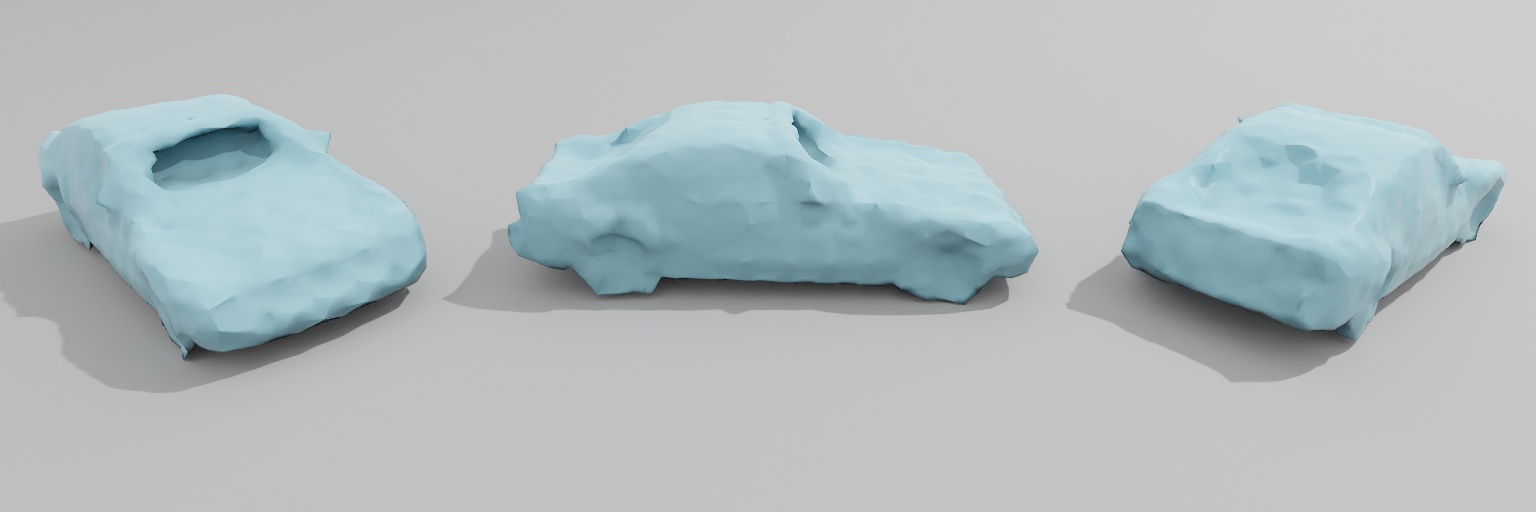}
\includegraphics[width=0.24\textwidth,trim=0 0 0 0,clip]{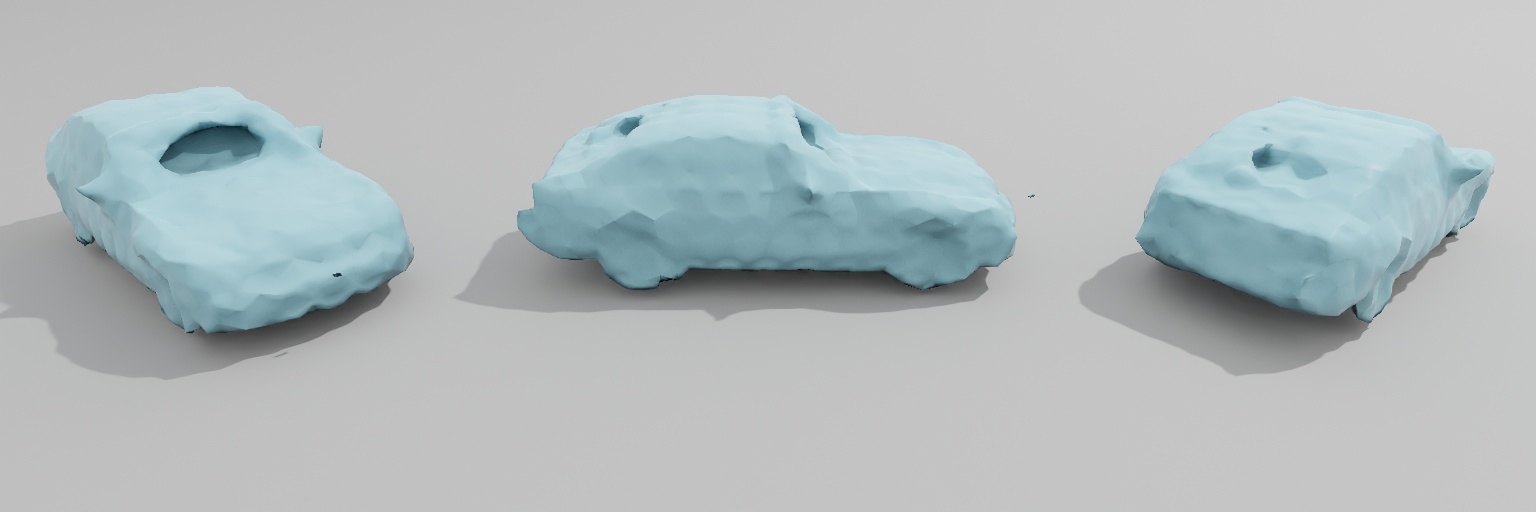}
\includegraphics[width=0.24\textwidth,trim=0 0 0 0,clip]{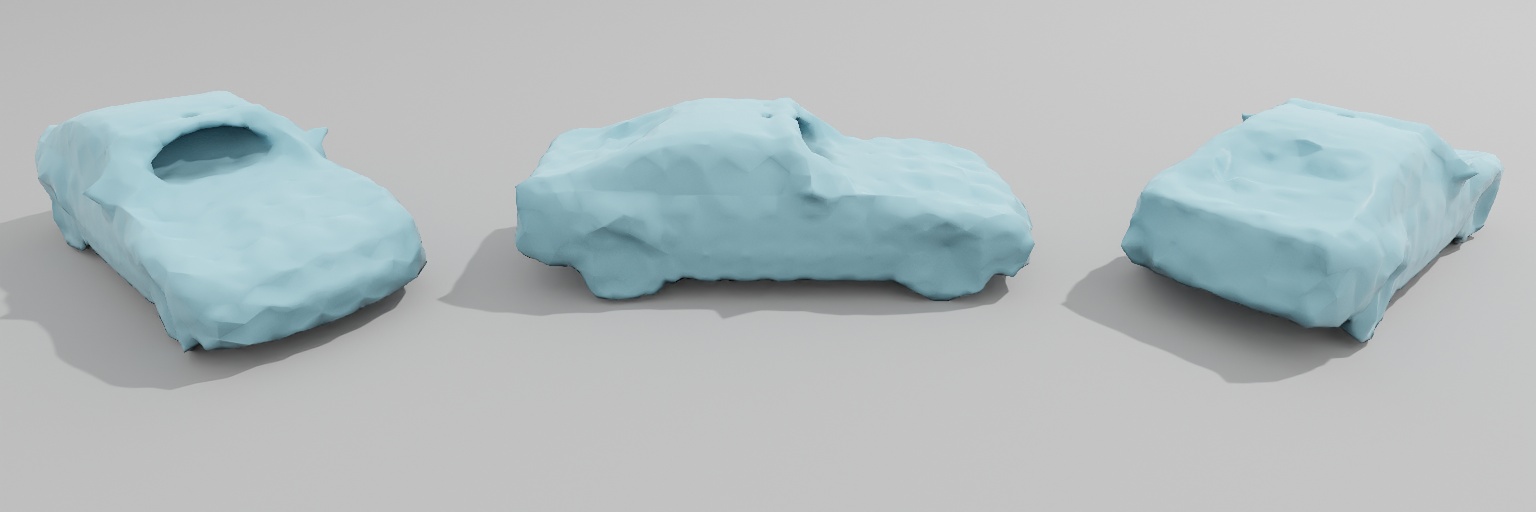}
\caption{ \footnotesize \textbf{Additional qualitative results of {\ourmodel} trained with "real" GANverse3D~\cite{ganverse3d} data.} We render generated shapes in Blender.}
\label{fig:qualitative_results_real}
\end{figure*}

\begin{figure*}[t!]
\centering
\includegraphics[width=\textwidth,trim=0 0 0 0,clip]{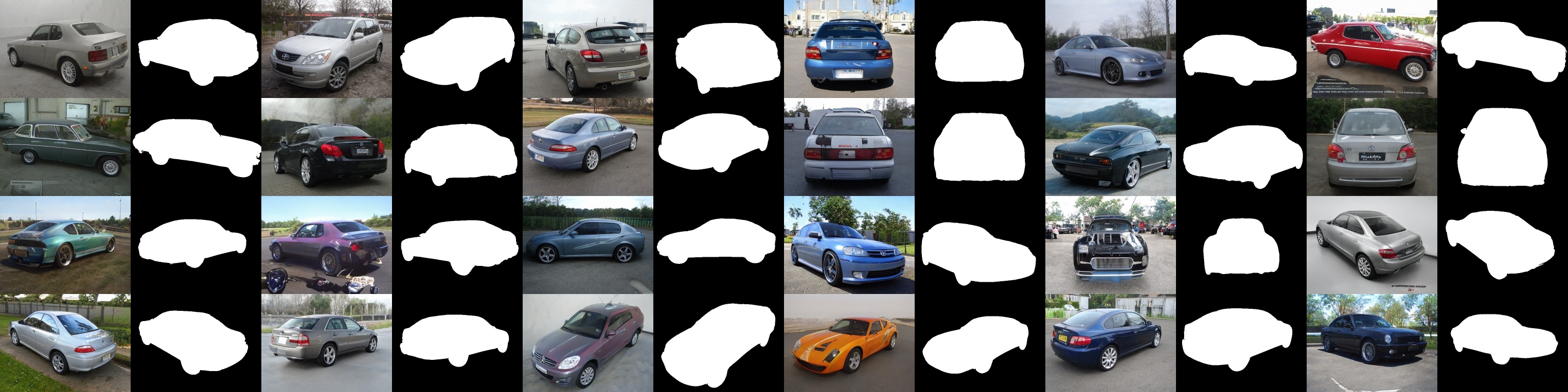}
\caption{\footnotesize We show randomly sampled 2D images and silhouettes from GANverse3D~\cite{ganverse3d} data. Note the realism of the images and the imperfections of the 2D silhouettes.}
\label{fig:samples_stylegan_dataset}
\end{figure*}

\subsection{Experiments on "Real" Image}
\label{sec:real_image}
Since many real-world datasets lack camera poses, we follow GANverse3D~\cite{ganverse3d} and utilize pretrained 2D StyleGAN to generate a realistic car dataset. We train {\ourmodel} on this dataset to demonstrate the potential applications to real-world data. 

\paragraph{Experimental Setting} Following GANverse3D~\cite{ganverse3d}, we manipulate the latent codes of 2D StyleGAN and generate multi-view car images. To obtain the 2D segmentation of each image, we use DatasetGAN~\cite{datasetgan} to predict the 2D silhouette.  We then use SfM~\cite{welinder2010caltech} to obtain the camera initialization for each generated image.  We visualize some examples of this dataset in Fig~\ref{fig:samples_stylegan_dataset} and refer the reader to the original GANverse3D paper for more details. Note that, in this dataset both cameras and 2D silhouettes are imperfect. 

\paragraph{Experimental Results} We provide qualitative examples in Fig.~\ref{fig:qualitative_results_real}. Even when faced with the imperfect inputs during training, {\ourmodel} is still capable of generating reasonable 3D textured meshes, with variation in geometry and texture. 

\subsection{Comparison with EG3D on Human Body}
\label{sec:compare_on_human}
Following the suggestion of the reviewer, we also train EG3D model on the \emph{Human Body} dataset rendered from Renderpeople~\cite{renderpeople} and compare it to the results of {\ourmodel}. 

\begin{figure*}[t!]
\centering
\includegraphics[width=0.32\textwidth,trim=0 0 0 0,clip]{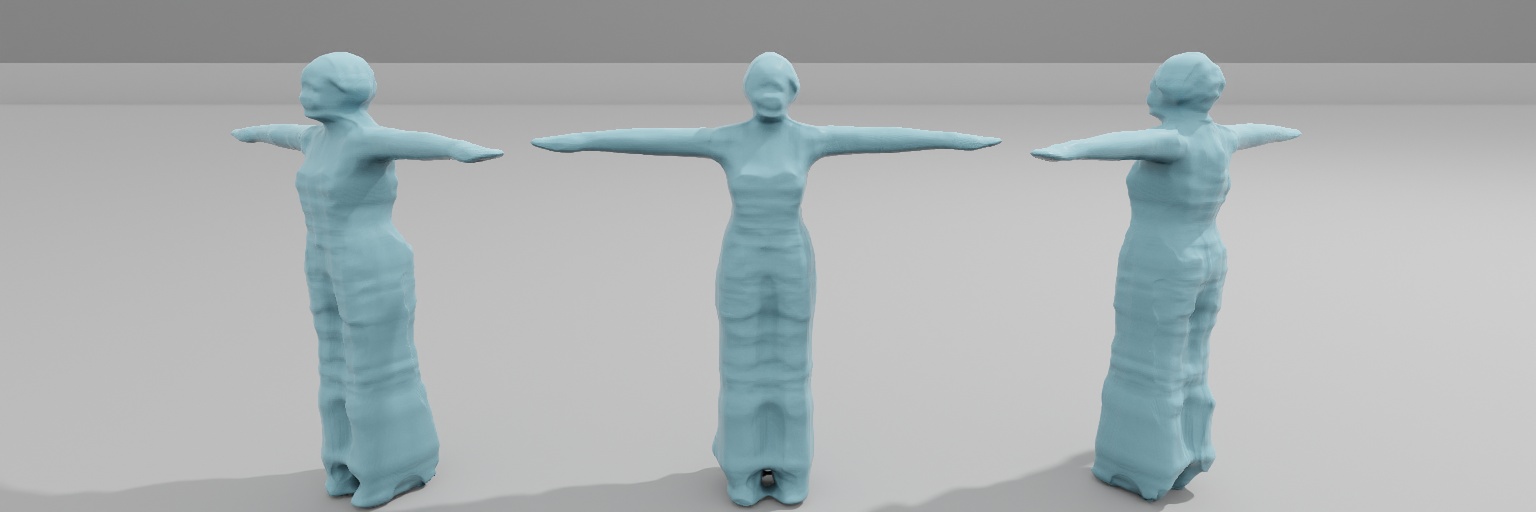}
\includegraphics[width=0.32\textwidth,trim=0 0 0 0,clip]{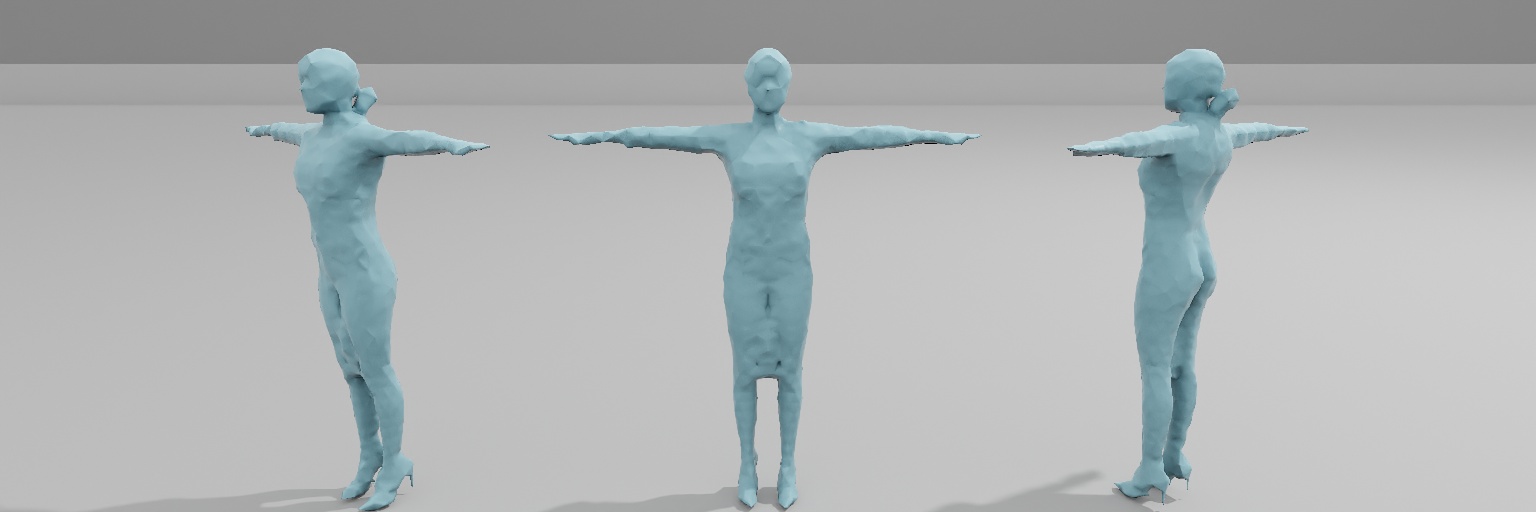}
\includegraphics[width=0.32\textwidth,trim=0 0 0 0,clip]{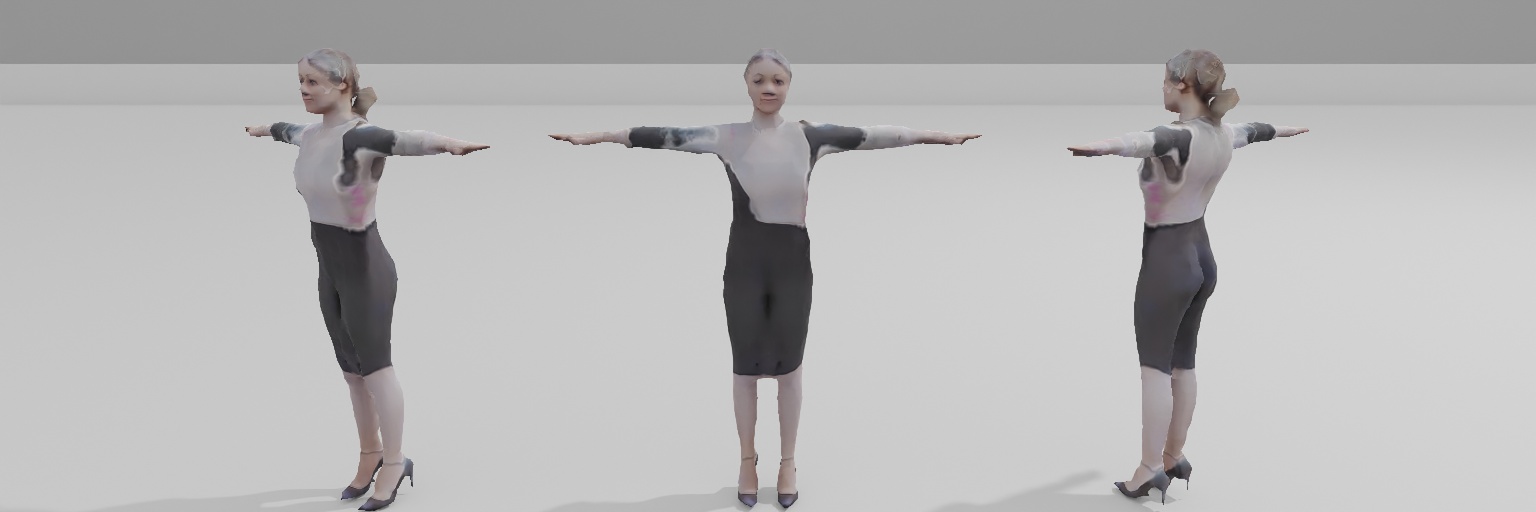}
\includegraphics[width=0.32\textwidth,trim=0 0 0 0,clip]{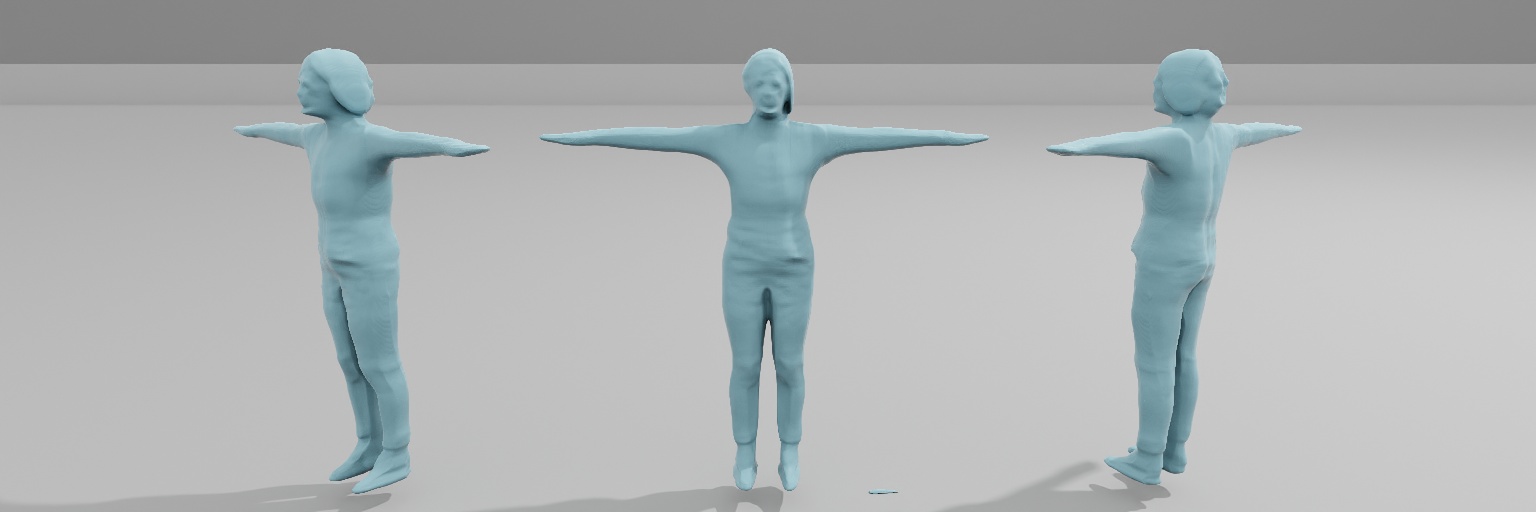}
\includegraphics[width=0.32\textwidth,trim=0 0 0 0,clip]{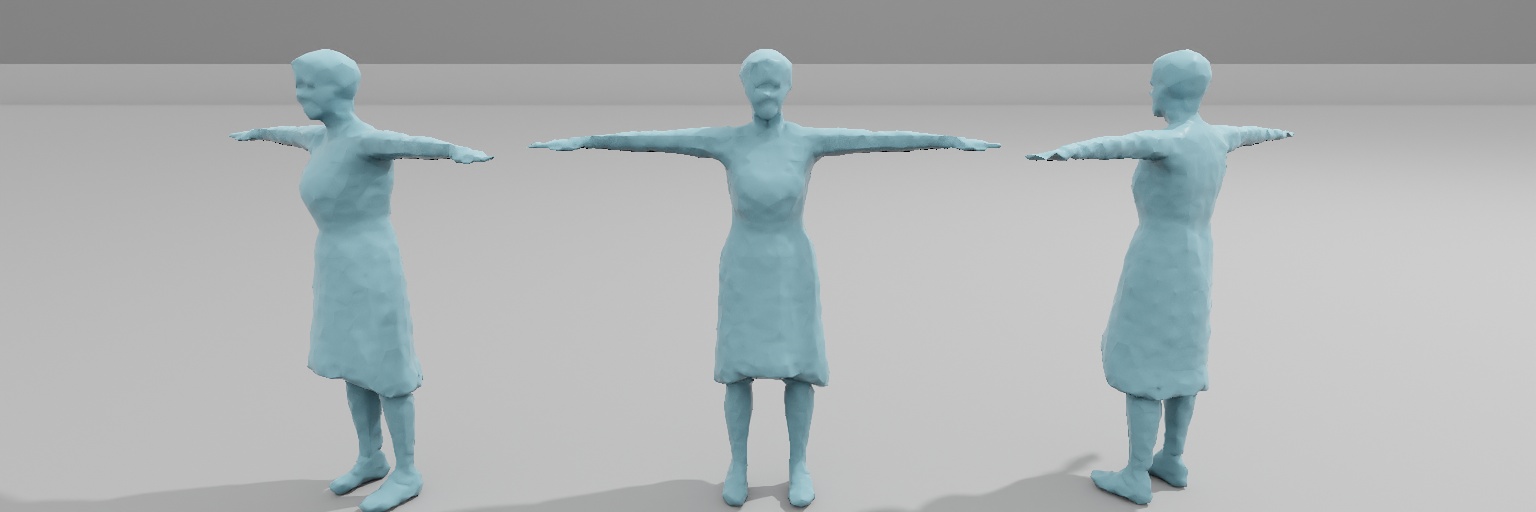}
\includegraphics[width=0.32\textwidth,trim=0 0 0 0,clip]{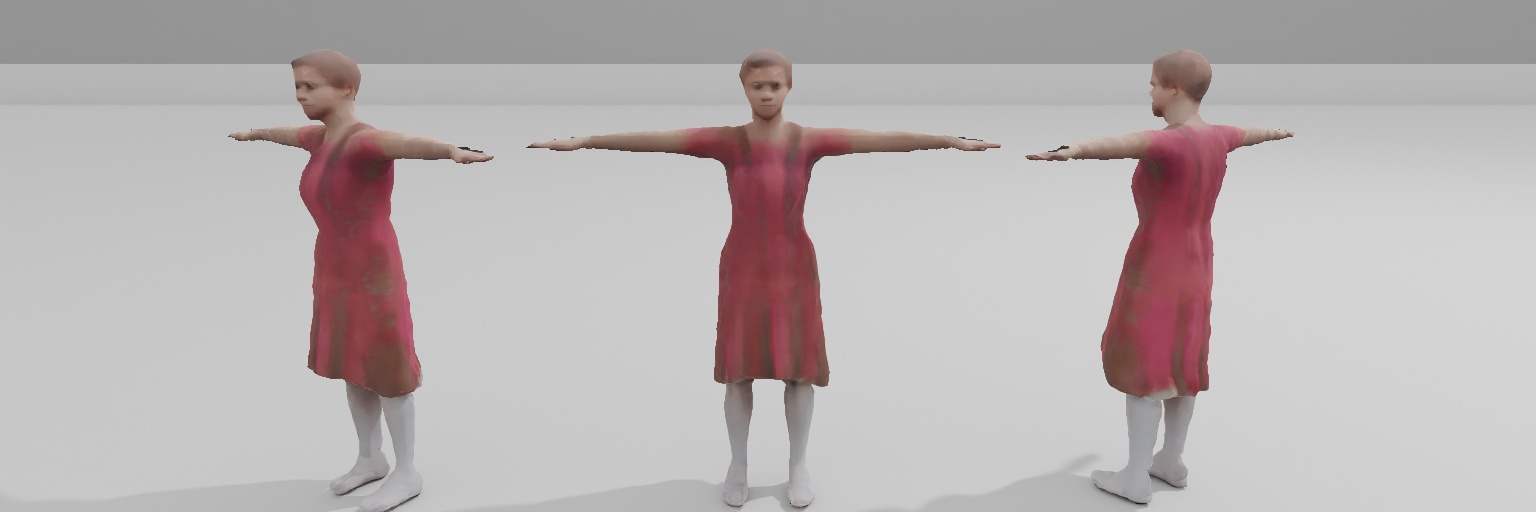}
\includegraphics[width=0.32\textwidth,trim=0 0 0 0,clip]{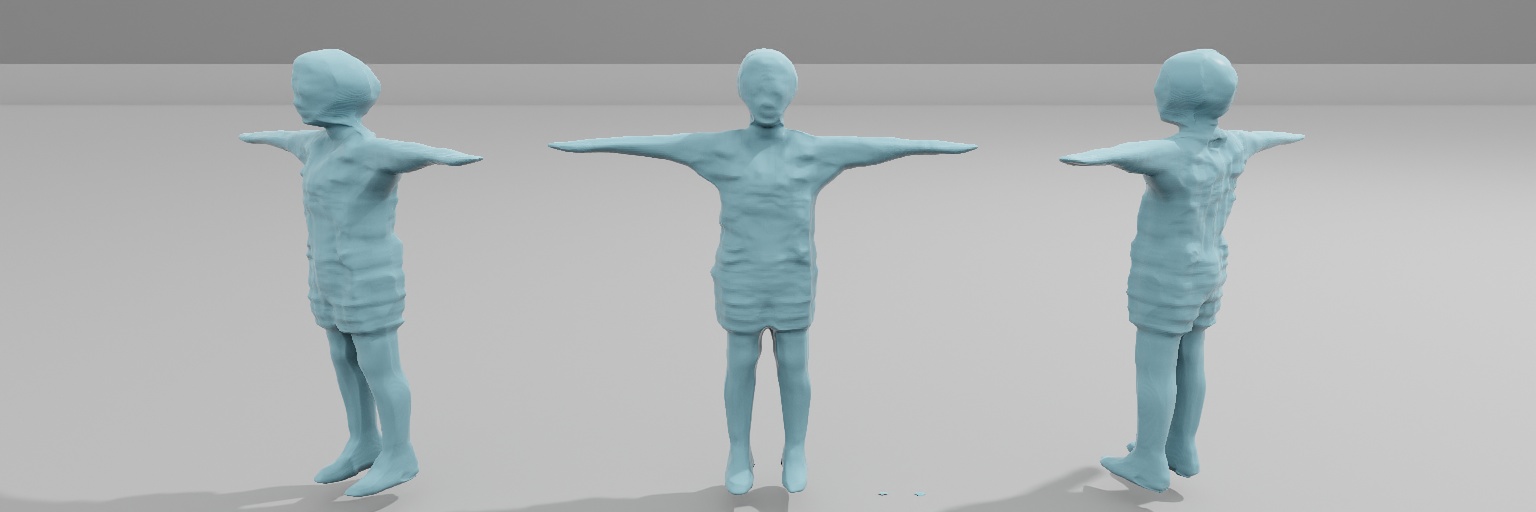}
\includegraphics[width=0.32\textwidth,trim=0 0 0 0,clip]{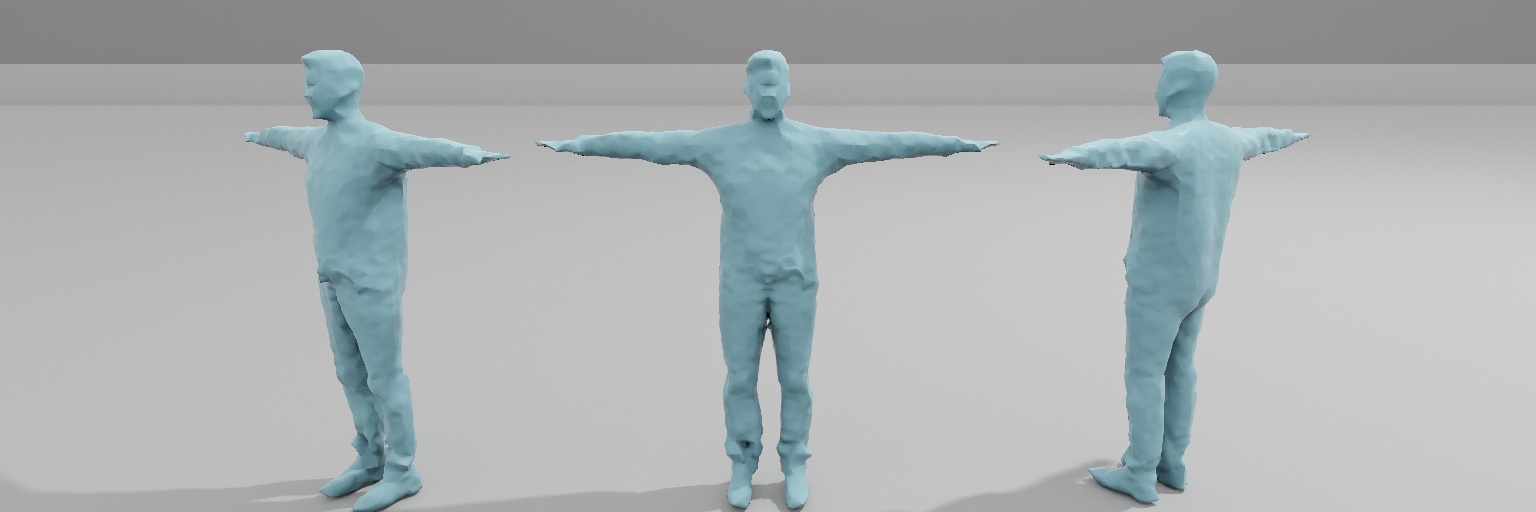}
\includegraphics[width=0.32\textwidth,trim=0 0 0 0,clip]{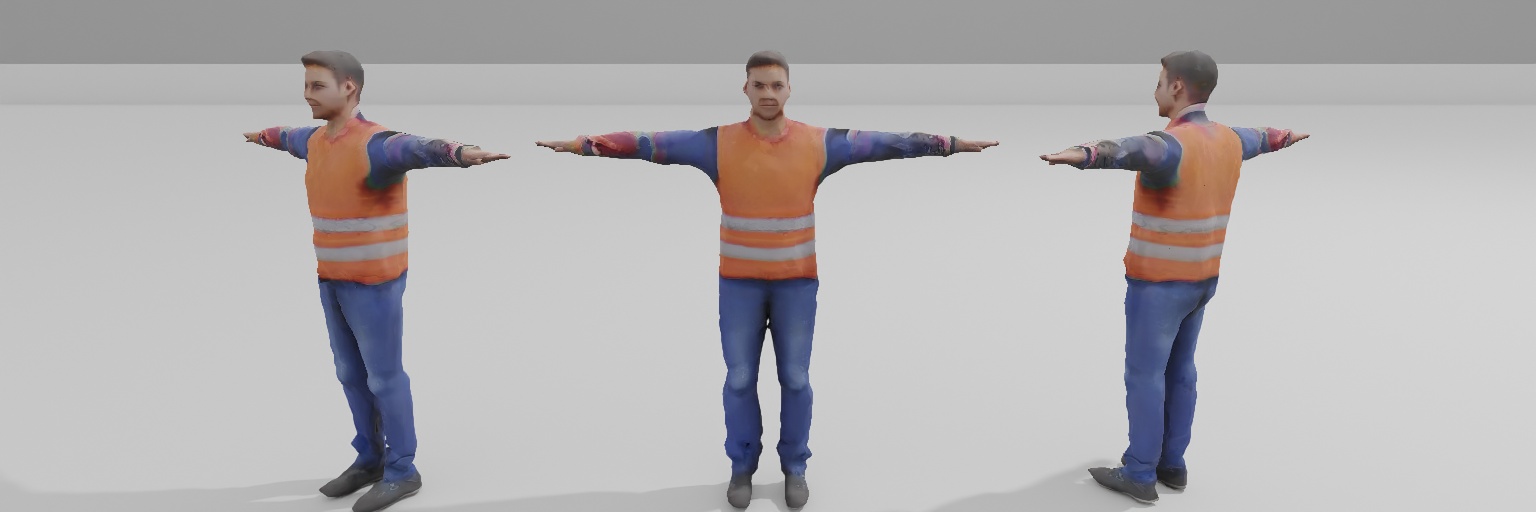}
\includegraphics[width=0.32\textwidth,trim=0 0 0 0,clip]{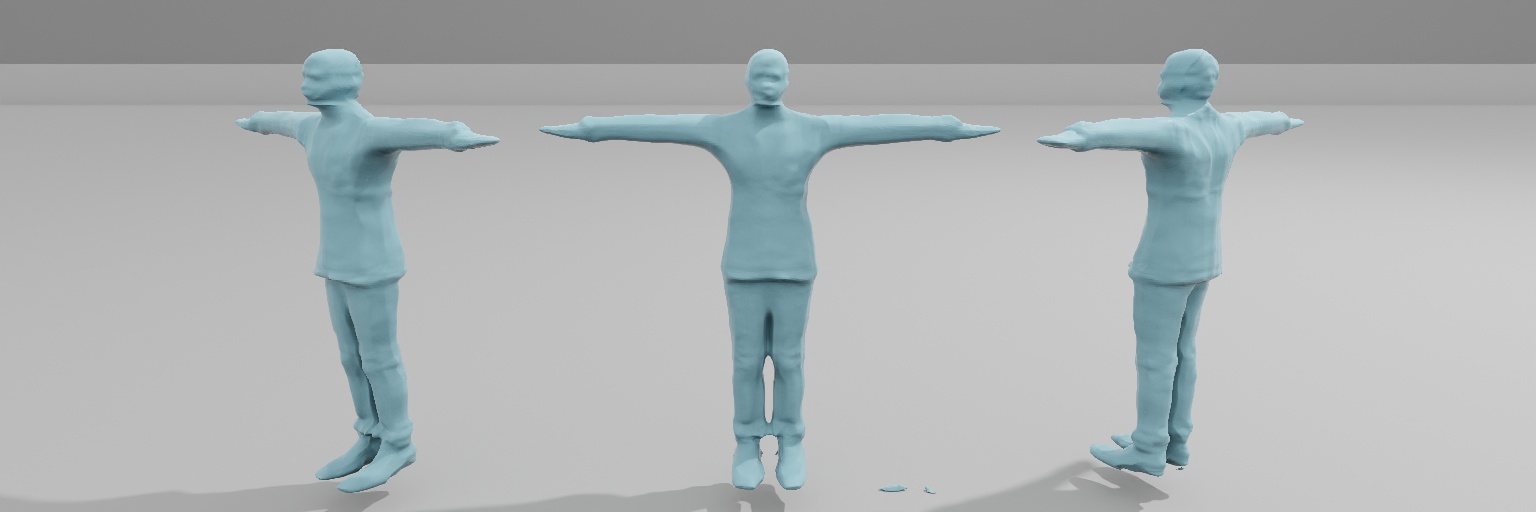}
\includegraphics[width=0.32\textwidth,trim=0 0 0 0,clip]{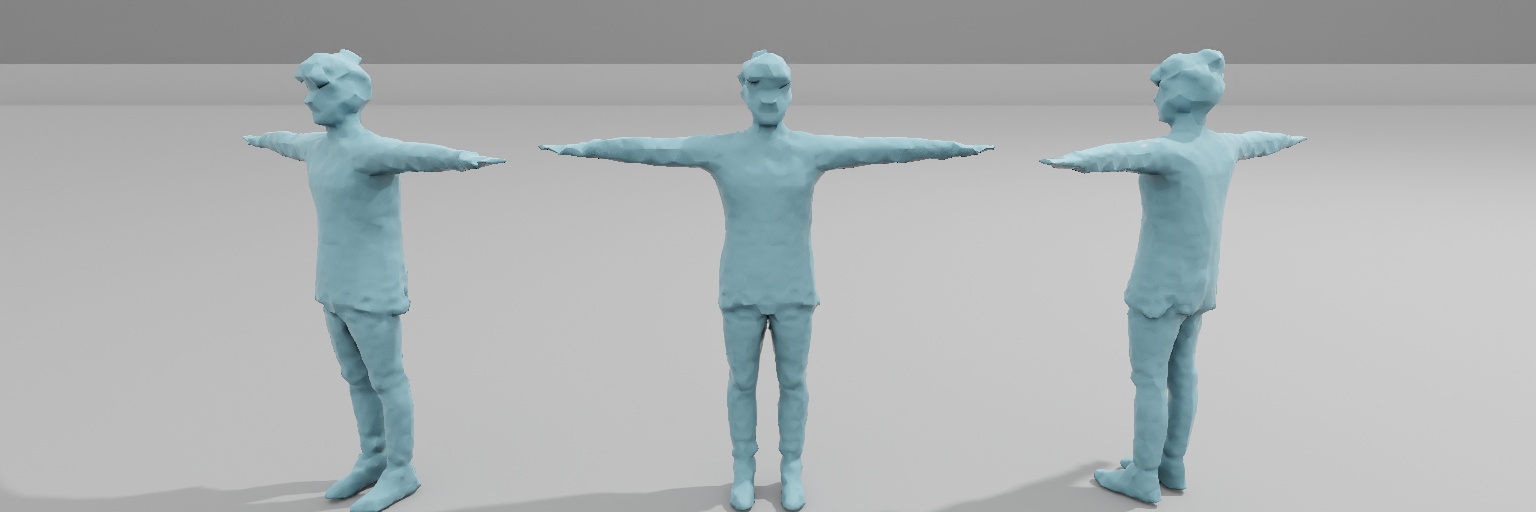}
\includegraphics[width=0.32\textwidth,trim=0 0 0 0,clip]{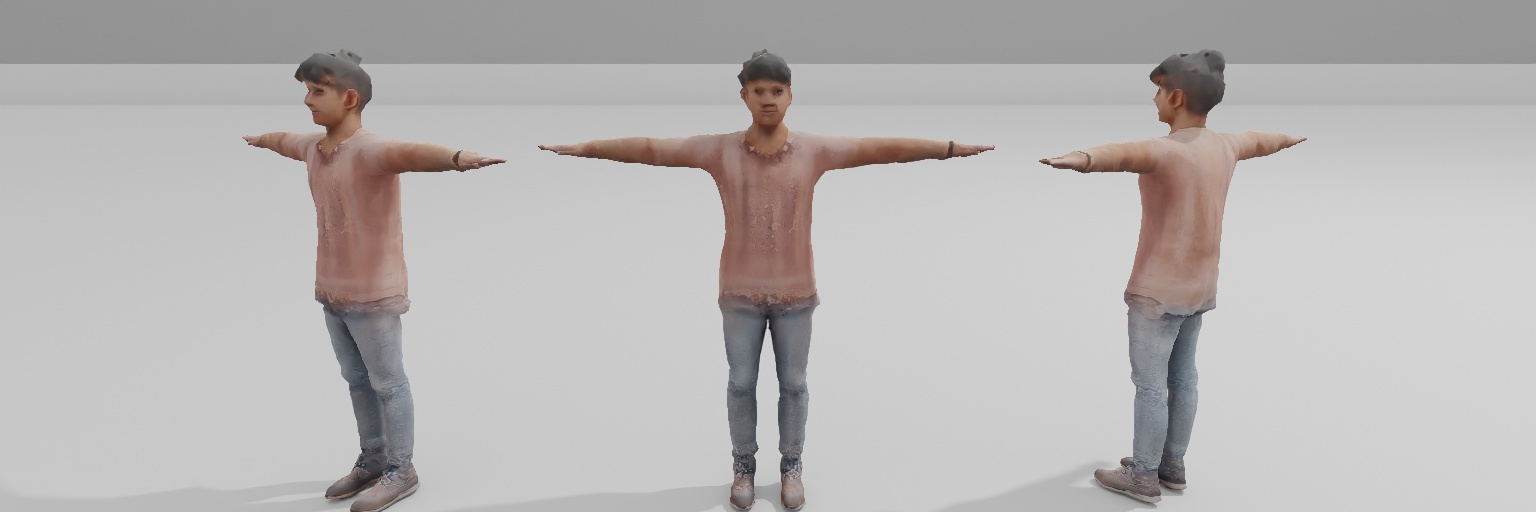}
\begin{minipage}[t] {\textwidth}
\begin{adjustbox}{width=\textwidth}
    \begin{tabular}{>{\centering\arraybackslash} p{6cm}
    >{\centering\arraybackslash} p{6cm}
    >{\centering\arraybackslash} p{6cm}
   }
     EG3D~\cite{eg3d} & Ours & Ours-Tex
    \end{tabular}
    \end{adjustbox}
\end{minipage}
\caption{ \footnotesize \textbf{Additional qualitative comparison on \emph{Human Body} dataset.} We compare our method with EG3D~\cite{eg3d} on the extracted geometry. }
\label{fig:qualitative_results_human}
\end{figure*}

\begin{table}
\centering
{
\begin{tabular}{lcc}
\toprule
Method & \multicolumn{2}{c}{FID ($\downarrow$)}\\
\cmidrule(lr){2-3}
& Ori & 3D \\
\midrule
EG3D~\cite{eg3d}&\textbf{13.77} & 60.42\\
{\ourmodel} & 14.27&\textbf{14.27}\\
\bottomrule
\end{tabular}
{\caption{ Additional quantitative comparison with EG3D~\cite{eg3d} on \emph{Human Body} dataset~\cite{renderpeople}.}
\label{tbl:add_exp_eg3d_human}
}
}
\end{table}
Quantitative results are available in Table~\ref{tbl:add_exp_eg3d_human} and qualitative comparisons in Figure~\ref{fig:qualitative_results_human}. {\ourmodel} achieves comparable performance to EG3D~\cite{eg3d} in terms of generated 2D images (FID-ori), while significantly outperforming it on 3D shape synthesis (FID-3D). This once more demonstrates the effectiveness of our model in learning actual 3D geometry and texture.

\section{Material Generation for View-dependent Lighting Effects}
\label{sec:exp_sg}

In modern computer graphics engines such as Blender~\cite{blender} and Unreal Engine~\cite{karis2013real}, surface properties are represented by material parameters crafted by graphics artists. 
To make the generated assets graphics-compatible, one direct extension of our method is to also generate surface material properties.
In this section, we describe how {\ourmodel} is able to incorporate physics-based rendering models, predicting SVBRDF to represent view-dependent lighting effects such as specular surface reflections. 

As described in main paper Sec.~\ref{sec:material}, two modules need to be adapted to facilitate material generation. Namely, the texture generation and the rendering process. Specifically, we repurpose the texture generator branch to predict the Disney BRDF properties~\cite{burley2012physically,karis2013real} on the surface as a reflectance field. Specifically, the texture generator now outputs a 5-channel reflectance property, 
including surface base color $\basecolor \in \mathbb{R}^3$, roughness $\rough \in \mathbb{R}$ and metallic $\metal \in \mathbb{R}$ parameters. 

Note that different from a texture field, rendering the reflectance field requires one additional shading step after rasterization into the G-buffer. 
Thus, the second adaptation is to replace the texture rasterization with an expressive rendering model capable of rendering the reflectance field. According to the non-emissive rendering equation~\cite{Kajiya86}, the outgoing radiance $\Lo$ at the camera direction $\wo$ is given by:
\begin{align}
    \Lo(\wo) =\int_\sphere \Li(\wi)\brdf(\wi,\wo; \basecolor, \rough, \metal)(\normal\cdot \wi)^+\,\dd \wi, \label{eq:generalrenderequation}
\end{align}
where $\Li$ is the incoming radiance, $\brdf$ is the BRDF, $\normal$ is the normal direction on the surface points, $\normal\cdot \wi$ is the cosine foreshortening term, $\wi$ is incoming light direction sampled on sphere $\sphere$, while $(\normal\cdot \wi)^+ = \max(\normal\cdot \wi, 0)$ constrains the integration over the positive hemisphere.
Standard ray tracing technique adopts Monte Carlo sampling methods to estimate this integral, but this incurs large computation and memory cost. Inspired by~\cite{wang2009sg,physg2020,chen2021dibrpp}, we instead employ a spherical Gaussian (SG) rendering framework \cite{chen2021dibrpp}, which approximates every term in Eq.~\eqref{eq:generalrenderequation} with SGs and allows us to analytically compute the outgoing radiance without sampling any rays, from where we can obtain the RGB color for each pixel in the image. We refer the reader to~\cite{chen2021dibrpp} for more details. 

Similar to the original training pipeline, we randomly sample light from a set of real-world outdoor HDR panoramas (detailed in the following ``Datasets'' paragraph) and render the generated 3D assets into 2D images using cameras sampled from the camera distribution of training set. 
We train the model using the same method as in the main paper by adopting the discriminators to encourage the perceptual realism of the rendered images under arbitrary real-world lighting, along with a second discriminator on the 2D silhouettes to learn the geometry. Note that no supervision from material ground truth is used during training, and the material decomposition emerges in a fully unsupervised manner. 
When equipped with a physics-based rendering models, GET3D successfully predicts reasonable surface material parameters, generating delicate models which can be directly utilized in stand rendering engines like Blender~\cite{blender} and Unreal~\cite{karis2013real}. 

\paragraph{Datasets}
We collect a set of 724 outdoor HDR panoramas from  HDRIHaven\footnote{\url{polyhaven.com/hdris} (License: CC0)}, DoschDesign\footnote{\url{doschdesign.com} (License: \url{doschdesign.com/information.php?p=2})} and HDRMaps\footnote{\url{hdrmaps.com} (License: Royalty-Free)}, which cover a diverse range of real-world lighting distribution for outdoor scenes.
We also apply random flipping and random rotation along azimuth as data augmentation. 
During training, we convert all the environment maps to SG lighting representations, where we adopt 32 SG lobes, optimizing their directions, sharpness and amplitudes such that the approximated lighting is close to the environment map. We optimize 7000 iterations with MSE loss and Adam optimizer. The converged SG lighting can preserve the most contents in the environment map.

As ShapeNet dataset~\cite{shapenet} does not contain consistent material definition, we additionally collect 1494 cars from Turbosquid~\cite{Turbosquid} with materials consistently defined with Disney BRDF. To render the dataset using Blender~\cite{blender}, we follow the camera configuration of ShapeNet Car dataset, and randomly select from the collected set of HDR panoramas as lighting.
In the dataset, the groundtruth roughness for car windows is in the range of $[0.2, 0.4]$ and the metallic is set to $1$; for car paint, the groundtruth roughness is in the range of $[0.3, 0.6]$ and the metallic is set to $0$. 
We disable complex materials such as the transparency and clear coat effects, such that the rendered results can be interpreted by the basic Disney BRDF properties including base color, metallic and roughness.

\paragraph{Evaluation metrics} 
Since we aim to generate 3D assets that can be used in graphics workflow to produce realistic 2D renderings, we quantitatively evaluate the realism of the 2D rendered images under real-world lighting using FID score. 
\paragraph{Comparisons} To the best of our knowledge, up to date no generative model can directly generate complex geometry (meshes) with material information. We therefore only compare different version of our model. In particular, we compare the results to the texture prediction version of {\ourmodel}, where we do not use material and directly predict RGB color for the surface points. We then ablate the effects of using real-world HDR panoramas for lighting, which are typically hard to obtain.  To this end, we manually use two spherical Gaussians for ambient lighting and a random directions to simulate the lighting when rendering the generated shapes during training, and try to learn the materials under this simulated lighting. 

\begin{figure*}[t!]
\centering
\includegraphics[width=\textwidth]{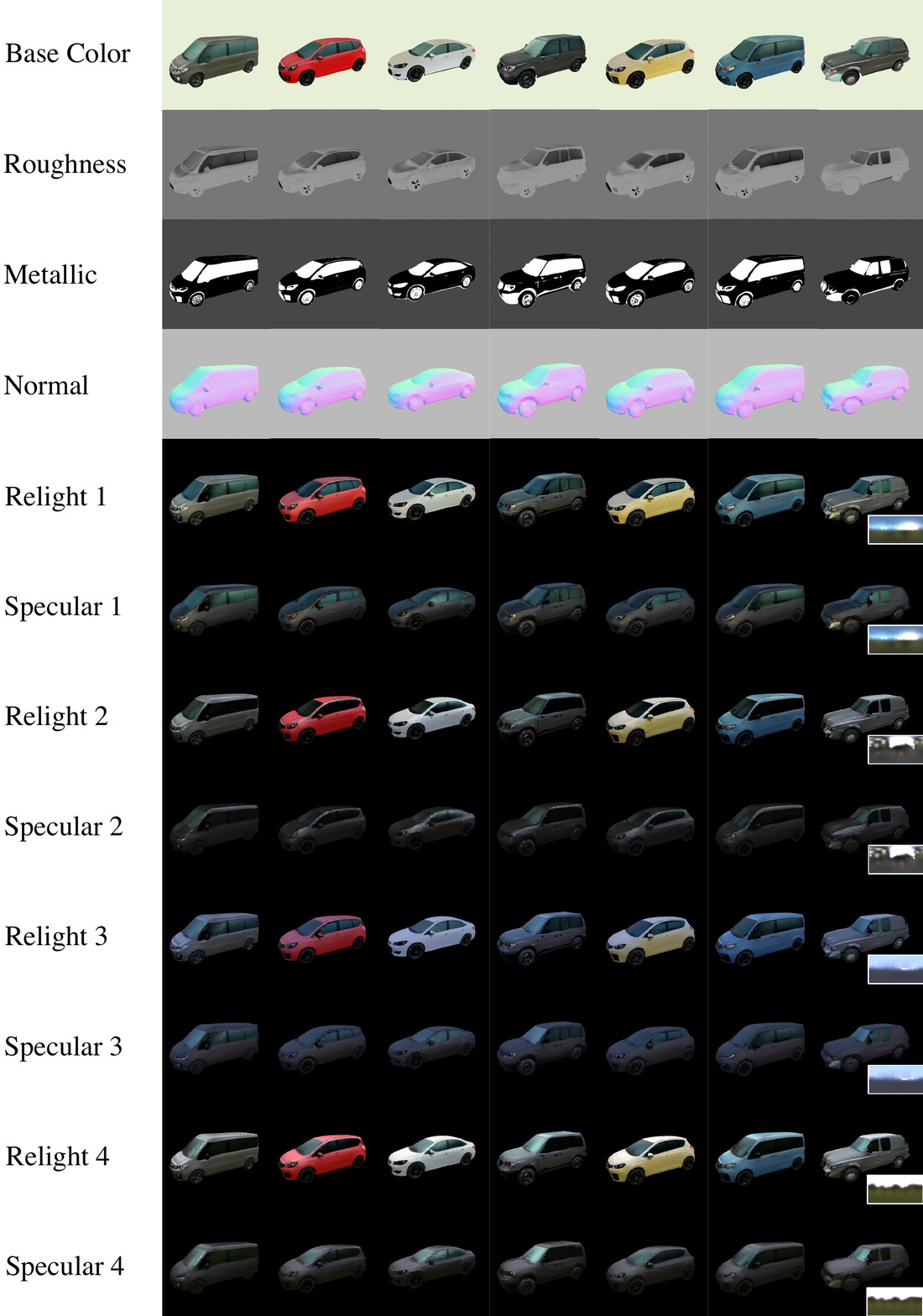}
\caption{\textbf{Material generation and relighting.} 
We visualize seven generated cars' material properties and relight with four different lighting conditions. 
} 
\label{fig:material_supp}
\end{figure*}

\begin{figure*}[t!]
\centering
\includegraphics[width=\textwidth]{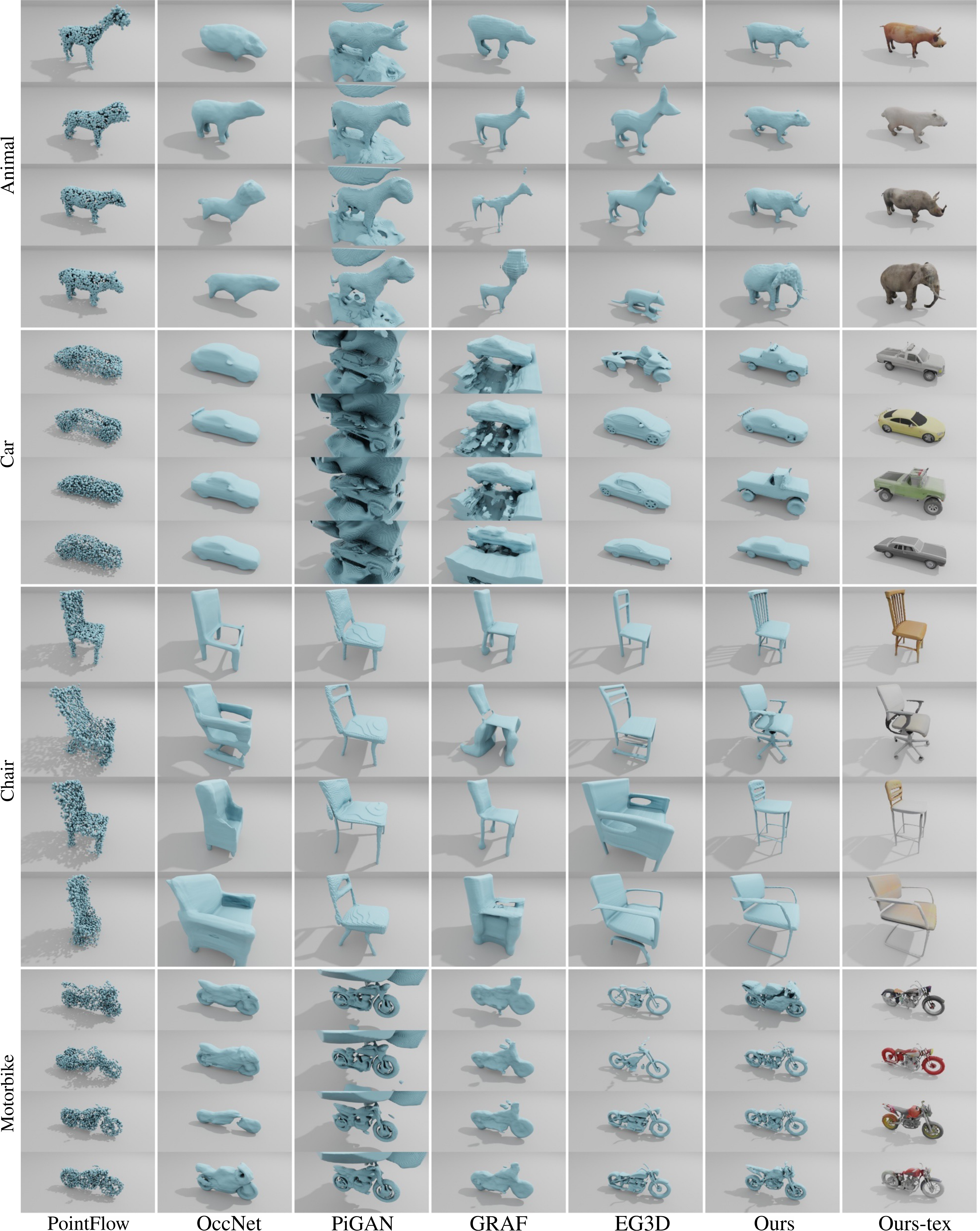}
\caption{\footnotesize \textbf{Generated 3D Geometry.}  Additional qualitative comparison with baseline methods on generated 3D geometry}
\label{fig:additional_gen_geo}
\end{figure*}

\begin{figure*}[t!]
\centering
\includegraphics[width=\textwidth]{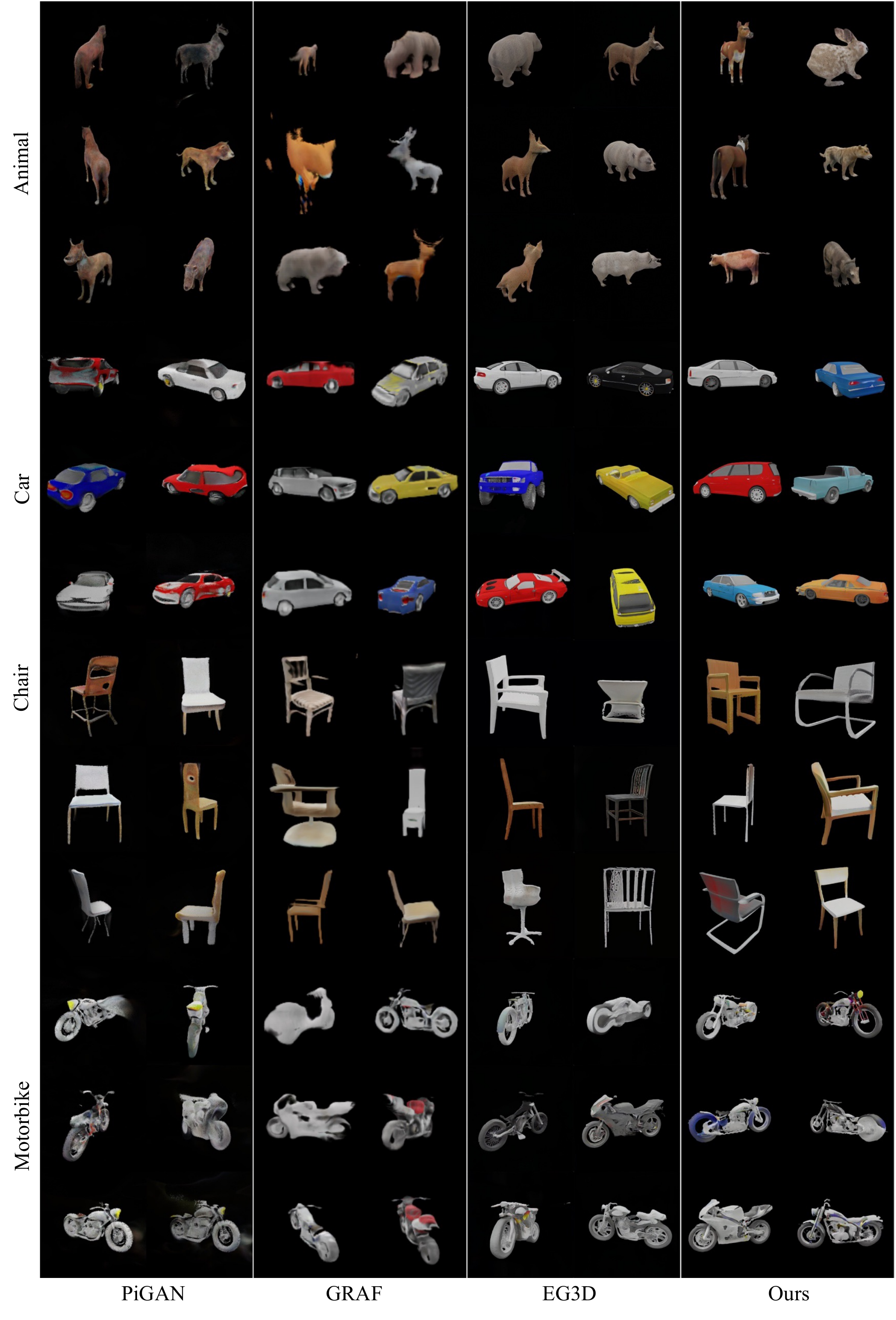}
\caption{\footnotesize \textbf{Generated Image.}  Additional qualitative comparison with baseline methods on generated 2D images.}
\label{fig:additional_gen_imgs}
\end{figure*}

\begin{figure*}[t!]
\centering
\includegraphics[width=.95\textwidth]{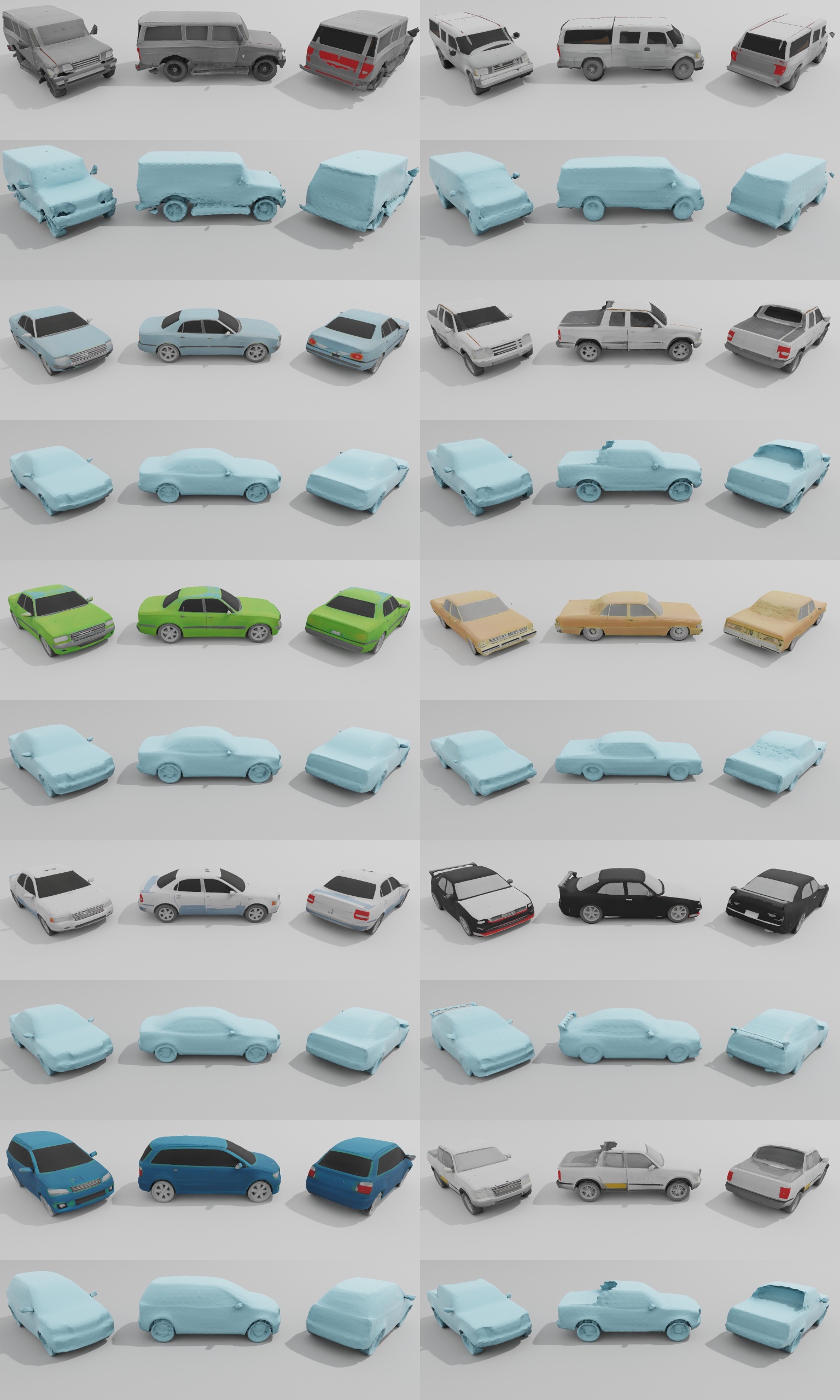}
\caption{Qualitative results on ShapeNet cars.
} 
\label{fig:qual_car}
\vspace{-3mm}
\end{figure*}

\begin{figure*}[t!]
\centering
\includegraphics[width=.95\textwidth]{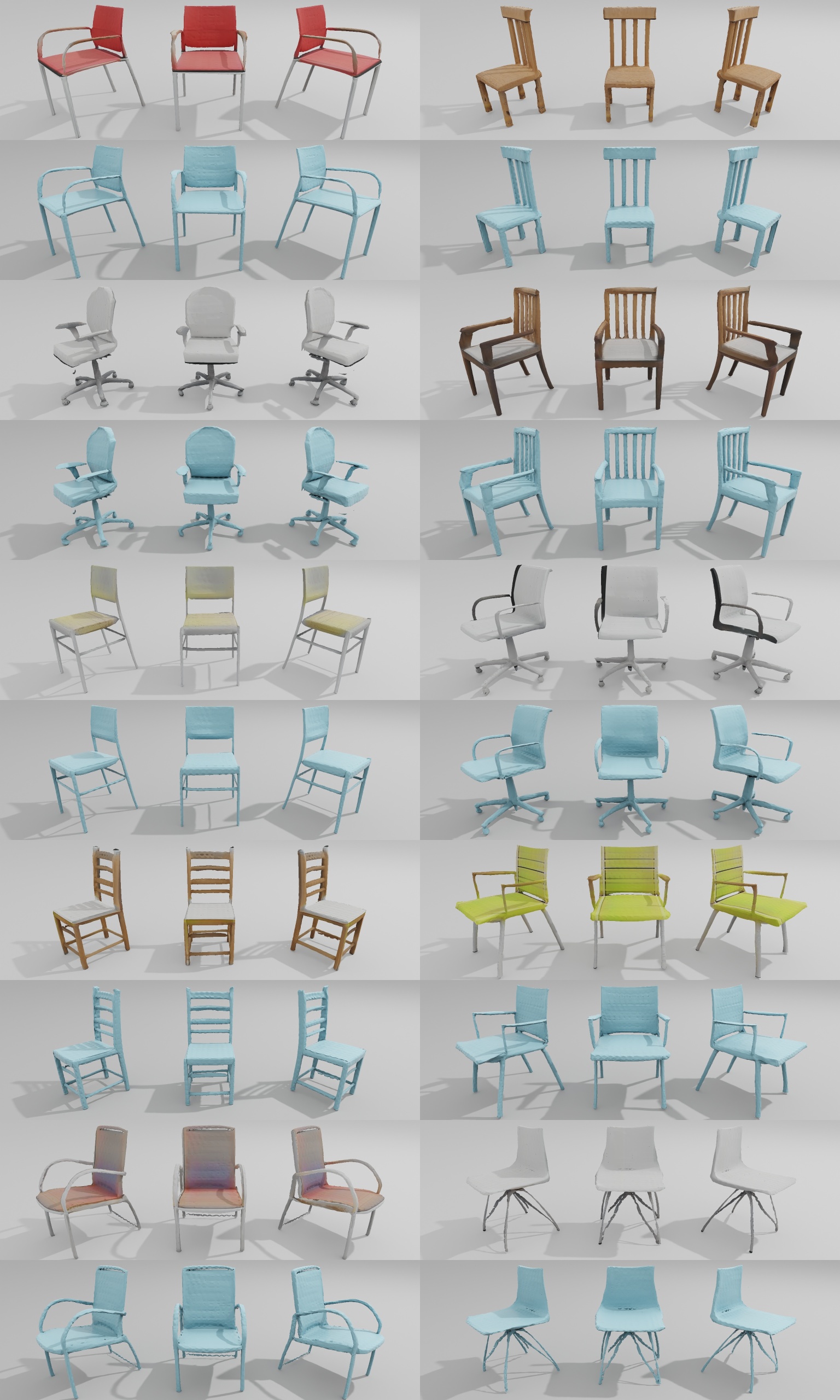}
\caption{Qualitative results on ShapeNet chairs.
} 
\label{fig:qual_chair}
\vspace{-3mm}
\end{figure*}

\begin{figure*}[t!]
\centering
\includegraphics[width=.95\textwidth]{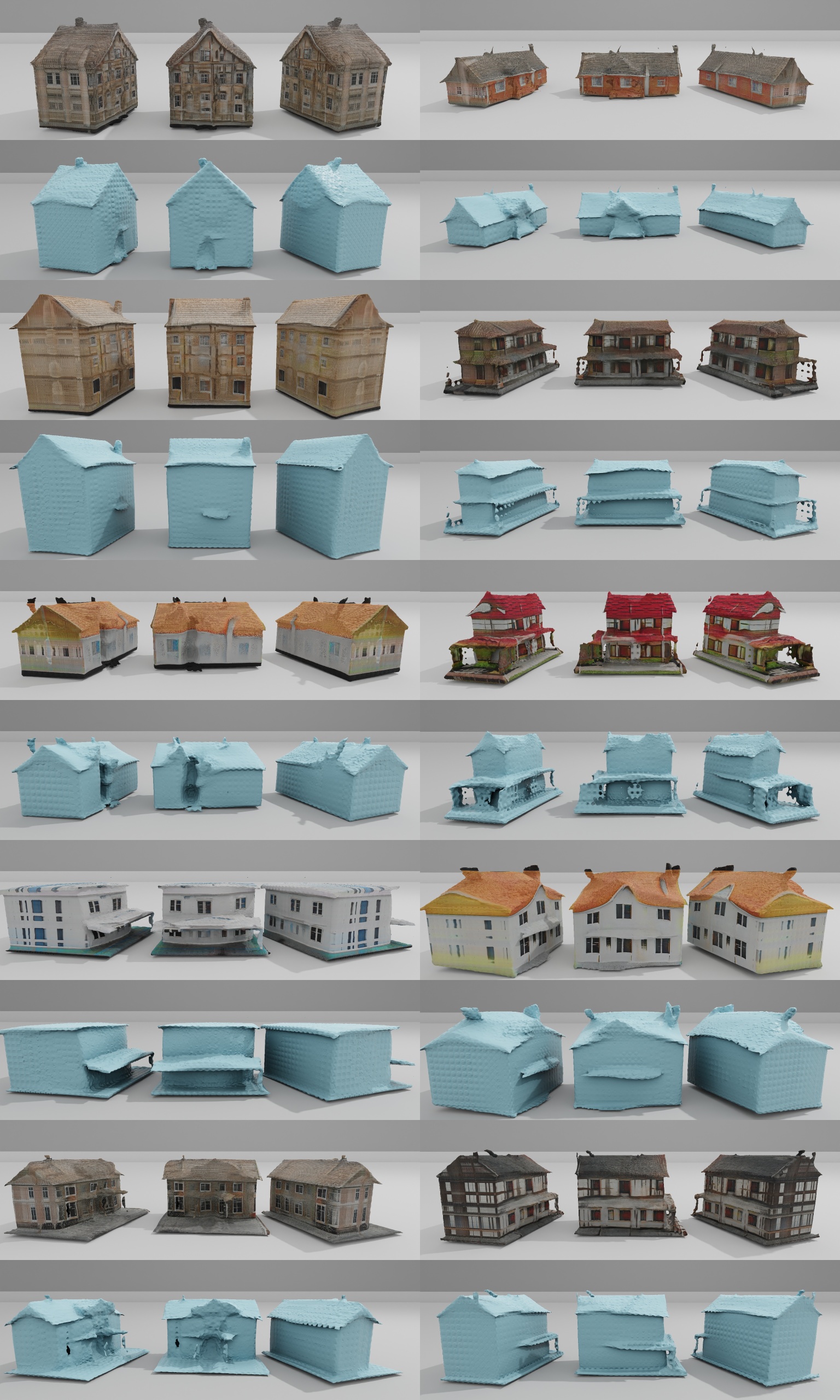}
\caption{Qualitative results on Turbosquid houses.
} 
\label{fig:qual_house}
\vspace{-3mm}
\end{figure*}

\begin{figure*}[t!]
\centering
\includegraphics[width=.95\textwidth]{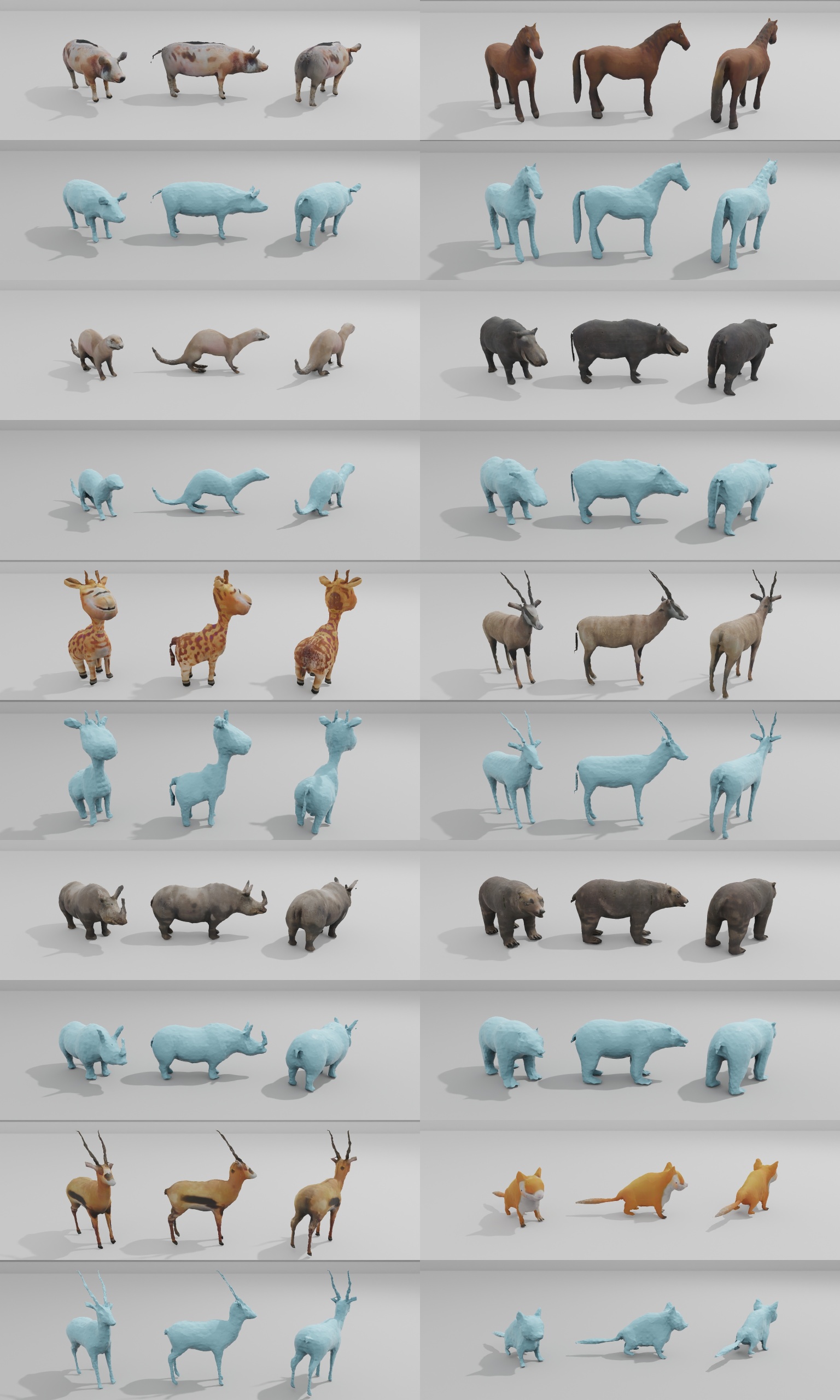}
\caption{Qualitative results on Turbosquid animals.
} 
\label{fig:qual_animal}
\vspace{-3mm}
\end{figure*}

\begin{figure*}[t!]
\centering
\includegraphics[width=.95\textwidth]{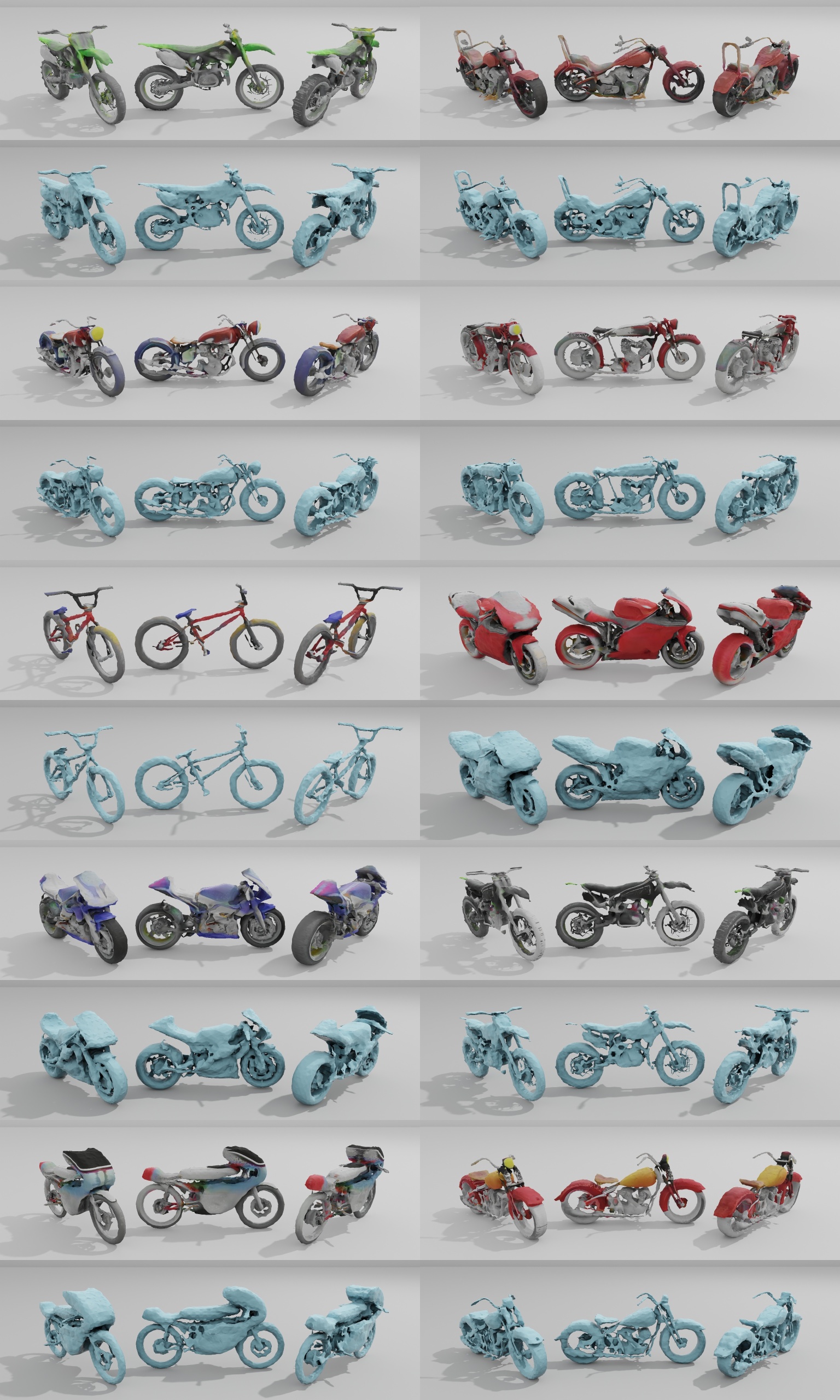}
\caption{Qualitative results on ShapeNet motorbikes.
} 
\label{fig:qual_motorbike}
\vspace{-3mm}
\end{figure*}

\begin{figure*}[t!]
\centering
\includegraphics[width=.95\textwidth]{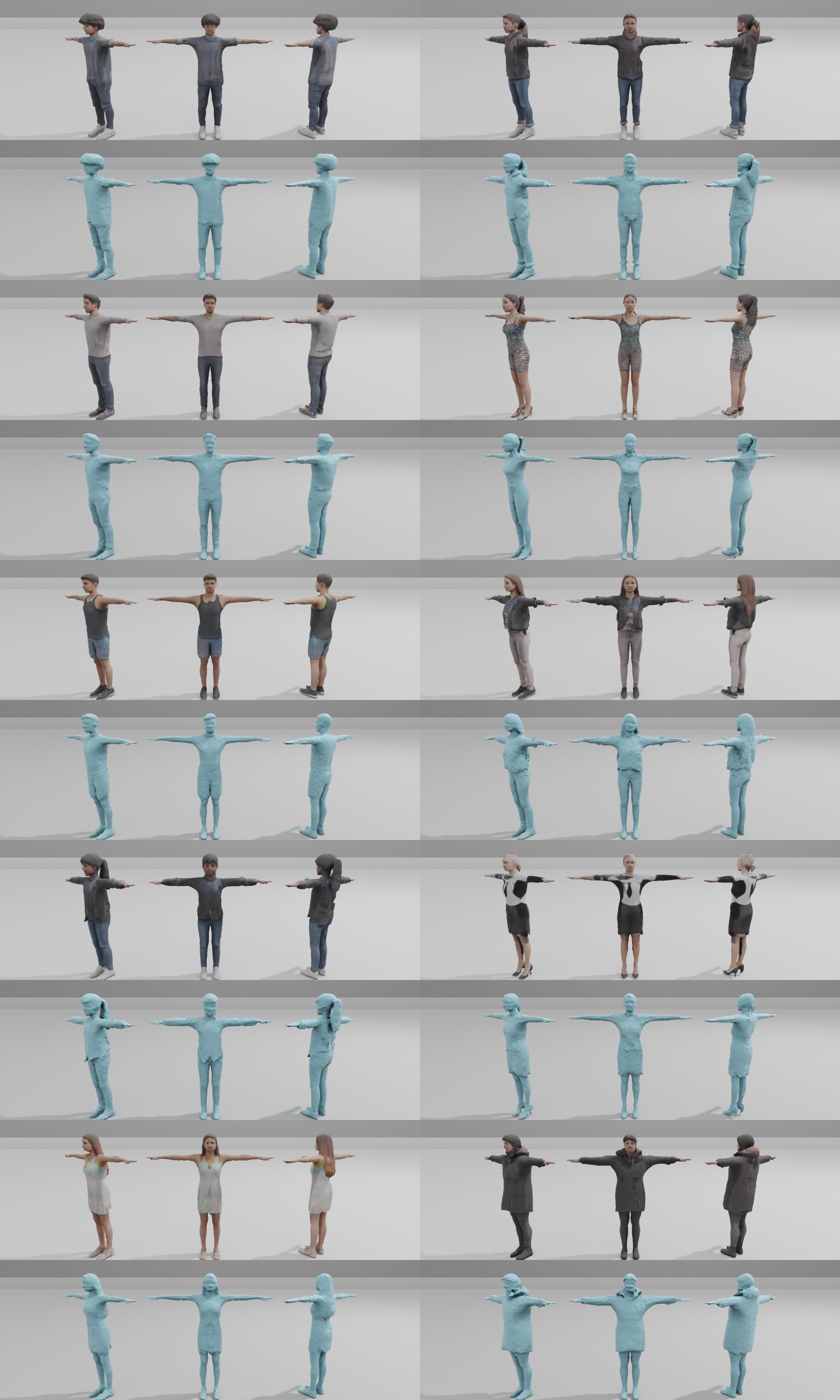}
\caption{Qualitative results on Renderpeople.
} 
\label{fig:qual_people}
\vspace{-3mm}
\end{figure*}

\begin{figure*}[t!]
\centering
\includegraphics[width=.95\textwidth]{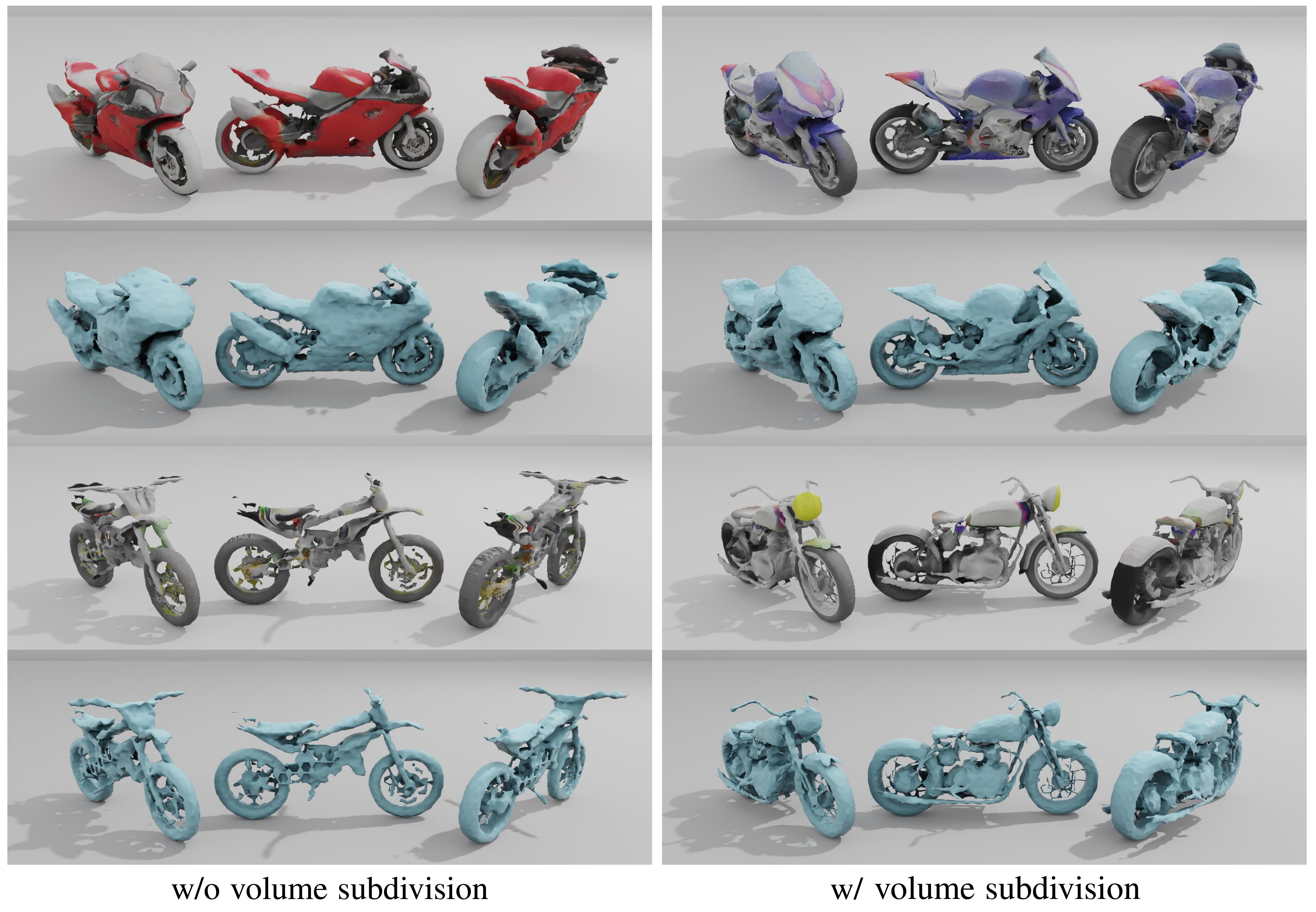}
\caption{We compare results with and without applying volume subdivision on ShapeNet motorbikes. With volume subdivision, our model can generate finer geometric details like handle and steel wire.
} 
\label{fig:ablate_vol_subdivision}
\vspace{-3mm}
\end{figure*}

\paragraph{Results} 
The quantitative FID scores are provided in Table~\ref{tbl:material}. 
With material generation, the FID score improves by more than 2 points when compared to the texture prediction baseline (18.53 vs 20.78). This indicates that the material generation version of {\ourmodel} has better capacity and improved realism compared to the texture only baseline. When using the simulated lighting, instead of real-world HDR panorama, the FID score gets slightly worse but still produces reasonable performance. 
We further provide additional qualitative results in Fig.~\ref{fig:material_supp} visualizing rendering results of generated assets under different real-world lighting conditions. We import our generated assets in Blender and show animated visualization in the accompanied video (\textit{demo.mp4}). 

\begin{table}[ht]
\centering
\begin{tabular}{lc}
\toprule
Method & FID \\
\midrule
Ours (Texture) & $20.78$ \\
Ours + Material (Ambient and directional light) & $22.83$ \\
Ours + Material (Real-world light) & $\mathbf{18.53}$ \\
\bottomrule
\end{tabular}
\caption{Quantitative FID results of material generation.}
\label{tbl:material}
\end{table}

\section{Text-Guided 3D Synthesis}
\label{sec:exp_text2mesh}
\paragraph{Technical details.} 
As briefly described in Sec.~\ref{sec:text_guided_synthesis}, our text-guided 3D synthesis method follows the dual-Generator design from StyleGAN-NADA~\cite{stylegan-nada}, and uses the directional CLIP loss~\cite{stylegan-nada}. In particular, at each optimization iteration, we randomly sample $N=16$ camera views and render $N$ paired images using two generators: the frozen one ($G_f$) and the trainable one ($G_t$).
The directional CLIP loss can then be computed as:
\begin{eqnarray}
L_{clip} = 1 - \frac{1}{N} \sum_{i = 1}^{N} \frac{\Delta I_i \cdot \Delta T }{ | {\Delta I_i} | \cdot | {\Delta T} | } 
\end{eqnarray}
where ${\Delta I_i} = E( R( G_t(w), c_i ) ) - E( R( G_f(w), c_i ) )$  is the translation of the CLIP embeddings ($E$) from the rendering with $G_f$ to the rendering with $G_t$, under camera $c_i$ and $\Delta T$ is the CLIP embedding translation from the class text label to the provided query text. In our implementation, we used two pre-trained CLIP models with different Vision Transformers (‘ViT-32/B’ and ‘ViT-B/16’)~\cite{dosovitskiy2020image} for different level of details, and follow the text augmentation as in the StyleGAN-NADA codebase\footnote{\url{https://github.com/rinongal/StyleGAN-nada} (MIT License)}.

\clearpage

\end{document}